\documentclass[10pt,twocolumn,letterpaper]{article}

\usepackage[pagenumbers]{wacv} 

\usepackage{booktabs}      
\usepackage{multirow}      
\usepackage{amsfonts,amssymb}  
\usepackage{nicefrac}      
\usepackage{adjustbox}     
\usepackage{pifont}
\usepackage{algorithmic}
\usepackage{algorithm}
\usepackage{wrapfig}
\usepackage{colortbl}

\renewcommand{\subfigure}[2][]{\subcaptionbox{#1}{#2}}

\definecolor{wacvblue}{rgb}{0.21,0.49,0.74}
\usepackage[pagebackref,breaklinks,colorlinks,allcolors=wacvblue]{hyperref}

\def\wacvPaperID{2625} 
\def\confName{WACV}
\def\confYear{2027}

\title{Continual Test-Time Adaptation via Entropy Sensitivity-Guidance in Strict Online Setting}

\author{Chandler Timm C. Doloriel$^{1}$, Yunbei Zhang$^{2}$, Muhammad Salman Siddiqui$^{1}$,\\
Tor Kristian Stevik$^{1}$, Fadi Al Machot$^{1}$, Kristian Hovde Liland$^{1}$, Habib Ullah$^{1}$\\
$^{1}$Faculty of Science and Technology (REALTEK), Norwegian University of Life Sciences (NMBU)\\
$^{2}$Tulane University\\
{\tt\small chandler.timm.cagmat.doloriel@nmbu.no}
}

\begin{document}
\maketitle

\begin{abstract}
Test-time adaptation (TTA) promises robustness under distribution shift by updating a pretrained model on unlabeled test data, but strict online TTA with batch size one and no access to source data is especially prone to drift or collapse. We introduce Sensitivity-Guided Erasing Adaptation (SEGA), a method for strict online continual TTA (CTTA) on corruption-style streams. SEGA uses a small number of structured erasures to probe how predictive entropy changes as information is removed, and uses the resulting per-sample sensitivity trajectories to coordinate recovery and sample selection rather than relying on raw entropy or batch statistics. This yields a practical feedback signal for long-horizon batch-size-one adaptation without periodic resets or model reservoirs. In experiments on ImageNet-C, CIFAR10/100-C, and corruption-generated aquaculture streams treated as controlled corruption-style proxies, SEGA yields consistent robustness and stability gains over strong CTTA baselines while reducing backward passes through sensitivity-based gating.
\end{abstract}
\section{Introduction}
\label{sec:intro}

\begin{figure*}[t!]
    \centering
    \small
    \subfigure[SAR~\citep{niu2023sar}]{
        \includegraphics[width=0.3\linewidth]{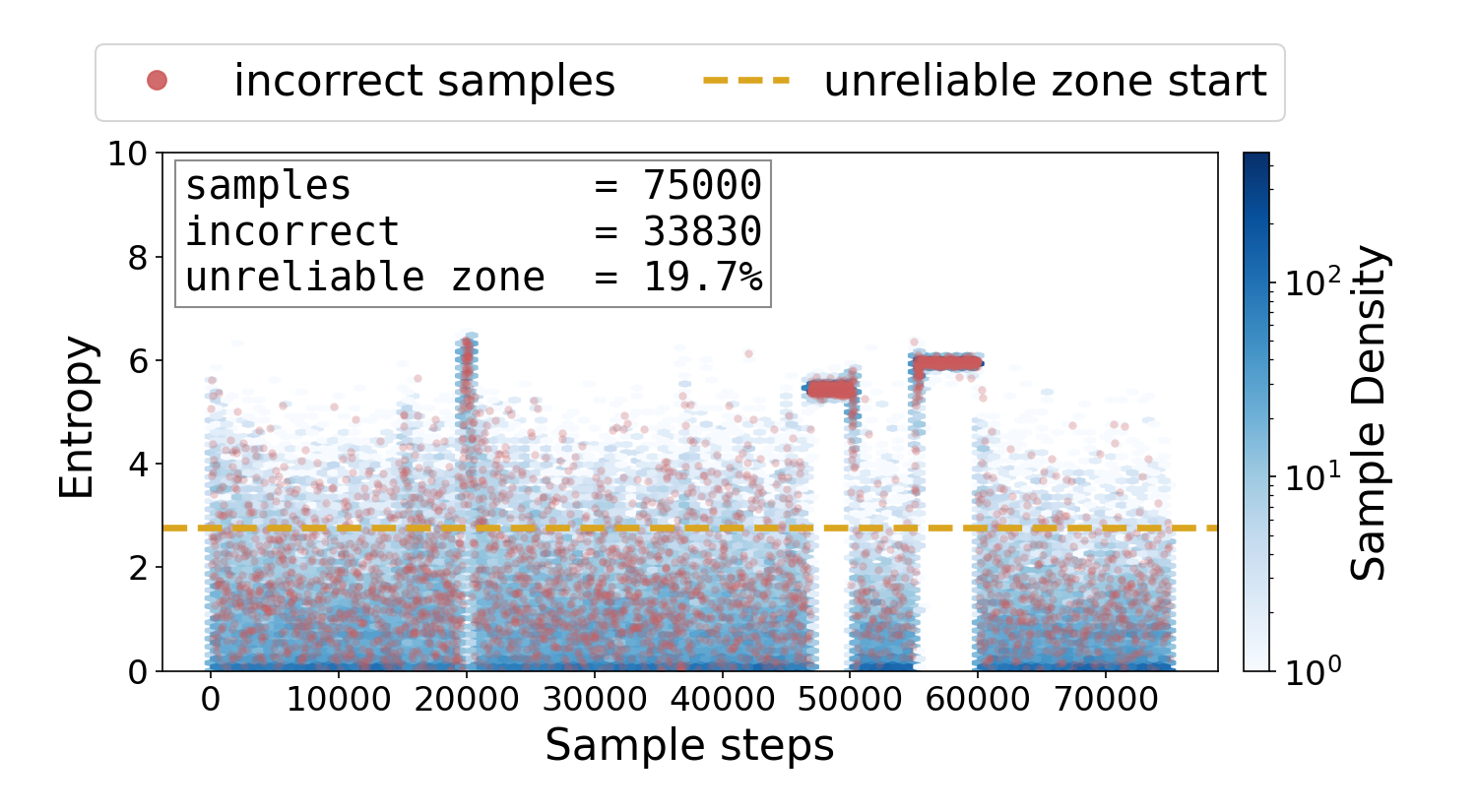}
    }
    \subfigure[RDUMB~\citep{Press2023RDumbAS}]{
        \includegraphics[width=0.3\linewidth]{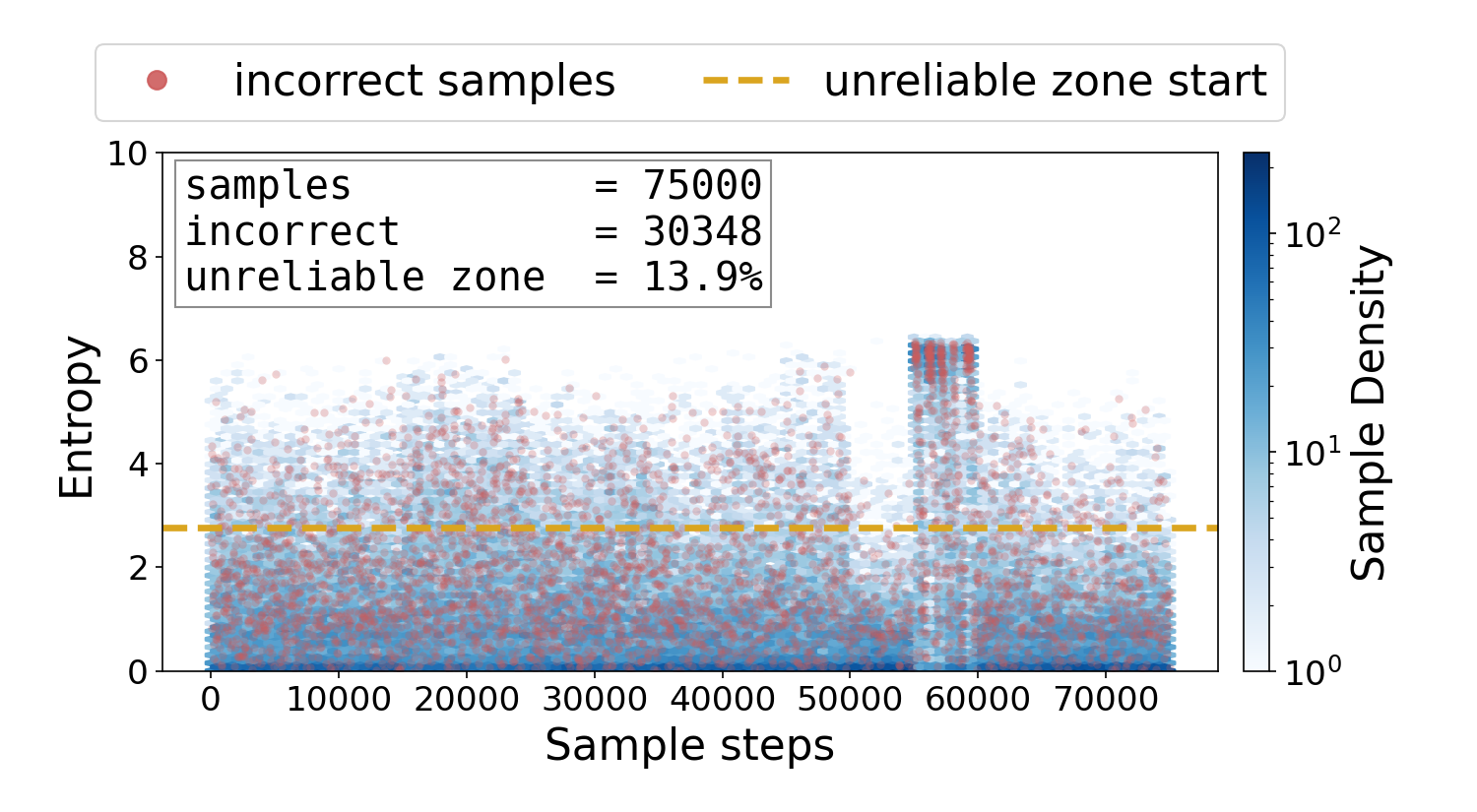}
    }
    \subfigure[SEGA]{
        \includegraphics[width=0.3\linewidth]{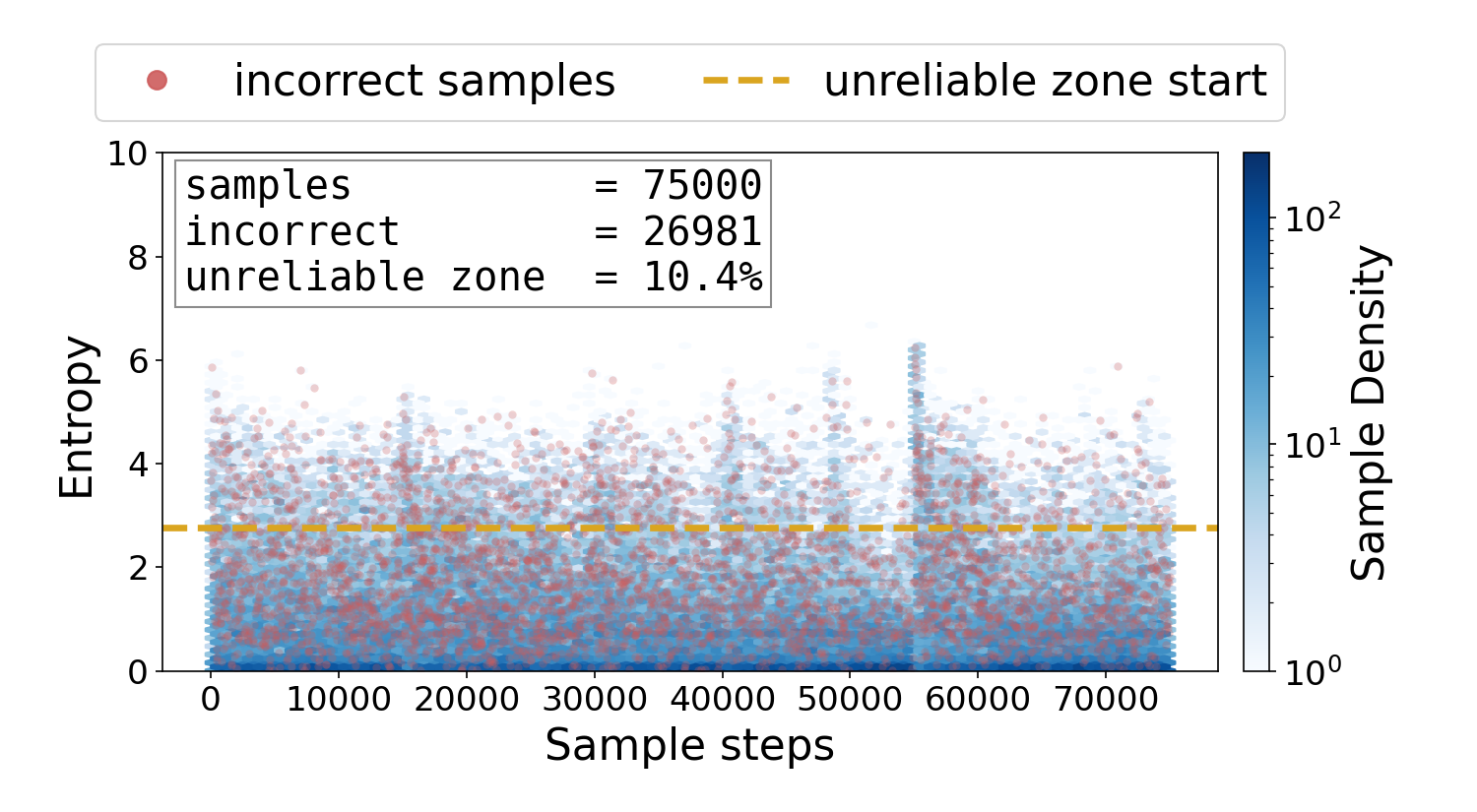}
    }

    \caption{Entropy trajectories along the CTTA stream (batch size 1) on ImageNet-C using ViT-Base for SAR~\citep{niu2023sar}, RDUMB~\citep{Press2023RDumbAS}, and SEGA. The horizontal band at $0.4 \ln C$ marks the ``unreliable'' entropy zone from SAR ($H > 0.4 \ln C$). Under this SAR-defined cutoff applied uniformly to all three methods, \textbf{SEGA} has both the fewest incorrect predictions and the smallest fraction of samples in the unreliable zone (about $10.4\%$ vs. $19.7\%$ for SAR).}
    \label{fig:problem}
\end{figure*}

In strict online continual test-time adaptation (CTTA), a deployed model processes unlabeled test samples one by one with no access to source data or past samples. This regime is especially prone to drift or collapse over long horizons~\citep{niu2023sar,Press2023RDumbAS,Seto2023REALMRE}, yet much of the TTA literature is developed for short streams, moderate batches, or auxiliary stabilization machinery~\citep{DequanWangetal2021,Zhang2021MEMOTT,Wangetal2022cotta,liu2023vida}. When $B{=}1$, batch-dependent normalization becomes unstable or uninformative in practice, and entropy-based adaptation becomes markedly more fragile~\citep{Lim2023TTNAD,ShijiNanoAdapt,NguyenTIPI,niu2023sar}. Methods that rely on teacher averaging, replay, or domain reservoirs also step outside the strict no-replay regime studied here~\citep{Wangetal2022cotta,liu2023vida,Vray2025}. With $B{=}1$, each gradient step is noisy and sample-specific, so without batch averaging or memory buffers to regularize updates, per-sample overfitting can accumulate into model collapse over thousands of steps, where the network becomes overly confident in a small subset of predictions~\citep{Press2023RDumbAS,Duan2025,Seto2023REALMRE}. This failure mode is relevant to streaming monitoring systems such as aquaculture, where frames arrive under changing lighting, weather, and fouling conditions~\citep{BiswasFishDisease2024}, but in this paper we study it through corruption-generated proxy streams that enable controlled long-horizon evaluation of collapse and recovery under $B{=}1$~\citep{hendrycks2019benchmarking,BiswasFishDisease2024}. Our goal is to isolate long-horizon update instability in a controlled proxy regime rather than to reproduce the full complexity of real deployments. Prior single-image and extremely small-batch TTA studies likewise argue that batching assumptions can fail under on-demand or resource-limited inference~\citep{Khurana2021SITASI,ShijiNanoAdapt,NguyenTIPI}. Unlike standard TTA with moderate batches, the model cannot average gradients across a mini-batch before updating. Unlike replay- or memory-based CTTA, it cannot revisit earlier samples or route inputs to stored domain experts. In this regime, adaptation quality is inseparable from collapse control, because every update is immediate, local, and effectively irreversible until a recovery rule intervenes. SEGA targets this collapse-control problem within online adaptation rather than the full universal TTA problem. Figure~\ref{fig:problem} shows that SAR~\citep{niu2023sar} and RDUMB~\citep{Press2023RDumbAS} spend a large fraction of time in the high-entropy ``unreliable'' zone and occasionally collapse, whereas SEGA maintains lower entropy and fewer errors.

Early TTA methods~\citep{sun2020test,DequanWangetal2021,Zhang2021MEMOTT,Gao2022BackTT} show that unlabeled updates can improve robustness but are mostly studied with short horizons and moderate batch sizes. CTTA methods add continual machinery such as stochastic restoration, prompts, masked reconstruction, and domain memories~\citep{Wangetal2022cotta,Chakrabarty2023SATASA,liu2023vida,liu2024continual,Duan2025,zhang2024dpcore,wang2025paid}, while stability-focused methods such as SAR, RDumb, and ReservoirTTA~\citep{niu2023sar,Press2023RDumbAS,Vray2025} use reliable-entropy filtering, periodic resets, and domain reservoirs. These approaches rely on raw entropy and batch- or memory-based statistics rather than per-sample predictive-entropy sensitivity.

We propose \emph{Sensitivity-Guided Erasing Adaptation} (SEGA), which uses multi-view erasing to compute a scalar operational sensitivity from the predictive-entropy trajectory under progressive masking, then uses this signal for two decisions: a \emph{trend recovery rule} that resets the model when sustained upward drift in sensitivity indicates instability, and a \emph{quantile gate} that skips the lowest-sensitivity samples. The same sensitivity signal therefore determines both when the model should recover and which samples should drive adaptation. We evaluate SEGA under strict CTTA with $B{=}1$ on ImageNet-C, CIFAR10/100-C, and corruption-generated aquaculture streams~\citep{BiswasFishDisease2024,hendrycks2019benchmarking}. Our contributions are: (1) we \textbf{frame strict online CTTA through per-sample sensitivity} for corruption-style batch-size-1 streams rather than batch statistics or raw entropy, (2) we \textbf{introduce SEGA}, a sensitivity-guided feedback mechanism for recovery and update selection in single-backbone online adaptation, and (3) we \textbf{provide empirical evidence across multiple corruption benchmarks and backbone architectures} that this design improves robustness and stability over prior CTTA baselines in this regime.

\section{Related Work}
\label{sec:relatedwork}

\paragraph{Early Test-Time Adaptation.} Early TTA showed that unlabeled test inputs can improve robustness at inference time. TTT~\citep{sun2020test} reformulates each test sample as a self-supervised problem, Tent~\citep{DequanWangetal2021} adapts by entropy minimization, MEMO~\citep{Zhang2021MEMOTT} couples adaptation with augmentations, and diffusion-based approaches~\citep{Gao2022BackTT} adapt the input instead of the model. These methods established source-free adaptation but are mostly evaluated on short benchmarks with batches or augmentations that do not expose the failure modes of strict online, batch-size-1 streams. Some later work does study single-image or extremely small-batch adaptation as a practical constraint~\citep{Khurana2021SITASI,ShijiNanoAdapt,NguyenTIPI}, but not as a long-horizon continual collapse problem.

\paragraph{Continual Test-Time Adaptation.} CTTA extends TTA to non-stationary streams. CoTTA~\citep{Wangetal2022cotta} reduces drift with teacher averaging and stochastic restoration, SATA~\citep{Chakrabarty2023SATASA} anchors updates to source prototypes, ViDA~\citep{liu2023vida} learns domain-specific and shared adaptation modules, Continual-MAE~\citep{liu2024continual} uses masked reconstruction, and DPCore, PAID, and LCoTTA~\citep{zhang2024dpcore,wang2025paid,Duan2025} control updates through prompt coresets, weight-structure priors, or gradient subspaces. Most still depend on batch- or memory-based statistics and auxiliary modules rather than treating each sample as the main control unit. These methods are orthogonal to the strict online setting studied here, which explicitly forbids access to past samples. Our setting is also narrower than broader universal TTA formulations that jointly vary additional deployment factors such as class-prior shift and wider online setting coverage~\citep{Marsden2023UniversalTA}.

\begin{table}[t!]
    \centering
    \caption{Collapse/drift policies of CTTA methods designed for test-time stability.}
    \label{tab:collapse_policy}
    \tiny
    \setlength\tabcolsep{8pt}
    \begin{adjustbox}{width=\linewidth,center=\linewidth}
    \begin{tabular}{l|cc}
        \toprule
        Method & Recovery & Filtering \\
        \midrule
        RES.TTA~\citep{Vray2025}                & Domain Reservoir & Style Routing \\
        SAR~\citep{niu2023sar}                  & Raw Entropy      & Raw Entropy \\
        RDUMB~\citep{Press2023RDumbAS}          & Periodic         & Raw Entropy \\
        M2A~\citep{Doloriel2026Family}          & None             & None \\
        SEGA                                    & Sensitivity      & Sensitivity \\
        \bottomrule
    \end{tabular}
    \end{adjustbox}
\end{table}

\paragraph{Test-Time Stability.} Normalization-based work shows the utility and fragility of batch statistics under shift~\citep{Ioffe2015BatchNA,Nado2020EvaluatingPB,Schneider2020ImprovingRA,Wu2021RethinkingI,Khurana2021SITASI,Lim2023TTNAD}. SAR~\citep{niu2023sar} replaces BatchNorm with GroupNorm and uses reliable entropy filtering and recovery heuristics. RDumb~\citep{Press2023RDumbAS} shows that periodic resets can outperform sophisticated CTTA schemes. ReservoirTTA~\citep{Vray2025} clusters style features and maintains domain-specialized model reservoirs (Table~\ref{tab:collapse_policy} summarizes these policies). M2A~\citep{Doloriel2026Family} uses progressive masking as a structured stability probe. Recent work also argues that raw entropy alone can be an unreliable control signal under shift and especially at small batch sizes~\citep{NguyenTIPI,Lee2024EntropyIN,Seto2023REALMRE}. Risk monitoring is adjacent to this problem, but it focuses on when to raise an alarm rather than how to recover the model~\citep{Schirmer2025MonitoringRI}. These works show that long-horizon stability depends on deciding when to update and recover, but they rely on entropy or style clustering with periodic or detector-driven resets rather than a direct per-sample sensitivity signal on a single backbone.

Across these strands, existing methods differ mainly in whether they use periodic global-entropy resets, style-based routing into domain reservoirs, or no explicit collapse policy. SEGA occupies a more specific point in this design space: a single-backbone method for strict online CTTA at $B{=}1$ on corruption-style streams, where it computes a per-sample sensitivity score $s_b$ from the erasing channel, enforces a stability budget through a trend rule, and selects update samples from an information-stability band. It should therefore be read as a targeted solution for strict online corruption-style CTTA rather than as a universal replacement for broader CTTA systems. Relative to SAR, the key change is replacing entropy-threshold filtering with a sensitivity signal that also drives recovery. Relative to RDumb and ReservoirTTA, SEGA avoids explicit replay, routing, and domain-specialized model storage. Relative to M2A, it does not use masking only as an adaptation loss or probe, but turns the resulting sensitivity into a shared control signal for both reset and update selection. The novelty is thus not masking alone, but reusing the resulting sensitivity as a common control variable for temporal recovery and sample-level updating.

\section{Sensitivity-Guided Erasing Adaptation}
\label{sec:methodology}

\begin{figure*}[!ht]
    \centering
    \small
    \includegraphics[width=0.9\linewidth]{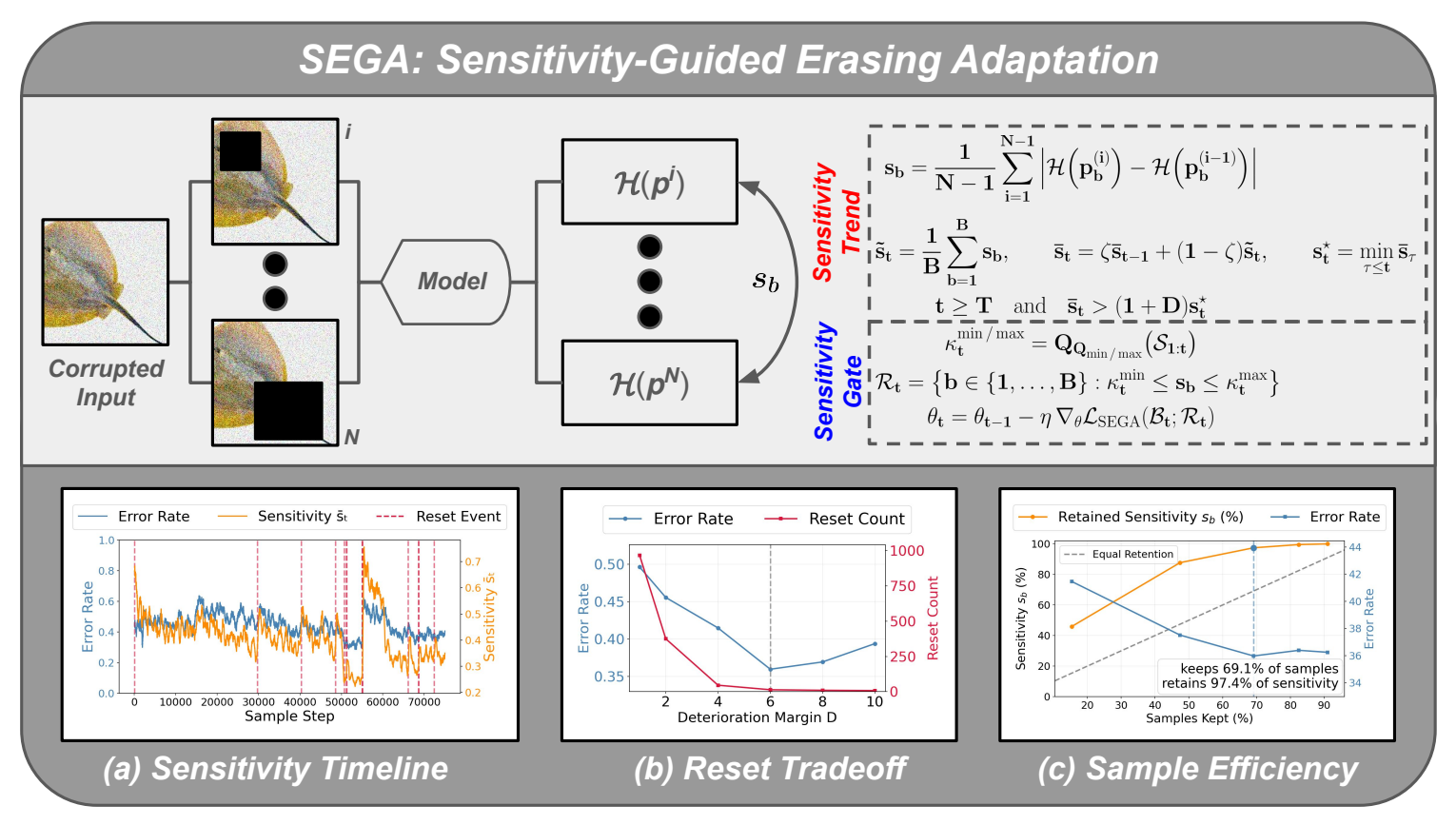}
    \caption{Overview of SEGA's workflow for strict online CTTA (batch size 1) on ImageNet-C with a ViT-Base backbone. The upper part visualizes the SEGA workflow: erased views are generated from the input and sensitivity is computed for the sensitivity-trend control and sensitivity-gated sample selection. The lower part shows the empirical validation of SEGA: (a) \textbf{Sensitivity Timeline:} overlays smoothed error and sensitivity with reset events, showing resets landing near sensitivity spikes and rising error so recovery starts before long unstable plateaus. (b) \textbf{Reset Tradeoff:} summarizes how the deterioration margin $D$ in the sensitivity-trend recovery trades off reset frequency against error. (c) \textbf{Sample Efficiency:} summarizes how the sensitivity-gated sample selection filters around 30\% of the samples while retaining most of the sensitivity signals.}
    \label{fig:method}
\end{figure*}

\paragraph{Preliminaries.} At test time, labels are unavailable, so SEGA measures how predictive uncertainty increases as spatial information is removed using only the model's outputs. Let $\mathcal{M}_i$ for $i\in\{0,\dots,N{-}1\}$ denote erasing operators that generate views $X^{(i)}=\mathcal{M}_i(X)$. SEGA is built on the premise that when predictions rely on robust cues, entropy changes mildly under progressive erasing, whereas brittle or spurious predictions yield sharp entropy changes. We treat this as an empirically supported heuristic rather than a formally proven property. As the model drifts toward collapse and becomes overly confident on a narrow prediction set, erasing disrupts these overfit predictions more sharply, producing elevated sensitivity. SEGA operationalizes this as a scalar sensitivity score $s_b$ (Eq.~\eqref{eq:sensitivity-score}) and uses it as a useful operational proxy for collapse risk rather than as a direct uncertainty estimate.

SEGA combines a multi-view erasing-based loss with two feedback control rules: a sensitivity-trend rule that triggers resets when a running sensitivity EMA drifts above its best value, and a quantile gate that drops the lowest-sensitivity samples. The trend rule handles sustained stream-level instability, whereas the gate handles per-sample update selectivity, with both decisions driven by the same sensitivity signal. SEGA is designed for immediate per-sample control under source-free, no-replay updates, not for domain accumulation, replay, or multi-model routing. SEGA uses a randomly placed block-wise spatial erasing family (studied in the supplementary material) and the resulting per-sample sensitivity trajectory for both decisions.

Let $f_{\theta}$ denote the classifier, where only a subset of parameters is updated at test time. Given a target batch $\mathcal{B}=\{x_b\}_{b=1}^{B}$, the model produces logits $z_b^{(0)} = f_{\theta}(x_b)$ and predictive distributions $p_b^{(0)} = \sigma\!\left(z_b^{(0)}\right)$, where $\sigma(\cdot)$ denotes the softmax operator and $p_b^{(0)} \in [0,1]^C$ is the class-probability vector for sample $x_b$ over $C$ classes.

\subsection{Sensitivity-trend control}
SEGA replaces periodic or entropy-based reset rules~\citep{Press2023RDumbAS,Vray2025} with the \emph{trend} of predictive-entropy sensitivity under erasing as its recovery signal. For each sample $x_b$ and erasing level $i$, we define $\rho_i = iA$ with $\rho_i \leq 1$, $x_b^{(i)} = x_b \odot \bigl(1-M_b^{(i)}\bigr)$, $z_b^{(i)} = f_{\theta}\bigl(x_b^{(i)}\bigr)$, and $p_b^{(i)} = \sigma\!\bigl(z_b^{(i)}\bigr)$ for $i \in \{1,\dots,N-1\}$, where $A>0$ is the erasing step size, $N$ is the number of erasing levels, and $M_b^{(i)} \in \{0,1\}^{H \times W}$ is a spatial mask with removed area approximately $\rho_i$, broadcast across channels. In the default ImageNet-C tuning, $(A,N)=(0.1,3)$, reused on CIFAR-C and the aquaculture streams.

With
\begin{equation}
\begin{split}
\mathcal{H}(p) &= -\sum_{c=1}^{C} p_c \log p_c, \\
s_b &= \frac{1}{N-1}\sum_{i=1}^{N-1} \left| \mathcal{H}\!\left(p_b^{(i)}\right) - \mathcal{H}\!\left(p_b^{(i-1)}\right) \right|,
\end{split}
\label{eq:sensitivity-score}
\end{equation}
$s_b$ is SEGA's operational sensitivity score: a model-dependent finite-difference of the predictive-entropy trajectory along the erasing path. Large $s_b$ occurs when modest erasing causes substantial entropy changes, indicating poorly captured domain shift. Small $s_b$ corresponds to stable predictions under perturbation. We use sustained periods of elevated $s_b$ as a proxy for increased collapse risk: as the model becomes overconfident on a narrow prediction set, erasing disrupts these predictions more sharply, so rising $s_b$ correlates with phases where recovery improves downstream error. The sensitivity timeline in Fig.~\ref{fig:method}(a) shows this correlation qualitatively, though we do not claim a formal causal link.

SEGA monitors how sensitivity evolves over the stream and intervenes only when the trajectory indicates sustained instability. Defining the batch-mean sensitivity, its EMA, and the best value observed so far as
\begin{equation}
\begin{split}
\tilde{s}_t &= \frac{1}{B}\sum_{b=1}^{B} s_b, \\
\bar{s}_t &= \zeta \bar{s}_{t-1} + (1-\zeta)\tilde{s}_t, s_t^{\star} = \min_{\tau \leq t} \bar{s}_{\tau},\\
\end{split}
\label{eq:slope-guard-stats}
\end{equation}
where $\zeta=0.9$, $\tilde{s}_t$ is the batch-average sensitivity, $\bar{s}_t$ is its EMA, and $s_t^{\star}$ is the best (lowest) running value. The trend rule enforces a stability budget by requiring $\bar{s}_t$ to stay within a deterioration margin of $s_t^{\star}$ after enough observations. Recovery is triggered whenever
\begin{equation}
t \geq T
\quad \text{and} \quad
\bar{s}_t > (1+D)s_t^{\star},
\label{eq:slope-guard-trigger}
\end{equation}
at which point the model is restored to its source-state parameters. This requires retaining the source weights (not source data) and discards adaptations accumulated since the last reset, making each recovery a fresh start rather than a partial rollback. Here $D>0$ is a relative deterioration margin and $T \in \mathbb{N}$ is a minimum activation horizon that prevents the EMA from reacting to transient spikes. We use default values $(T,D)=(50,6.0)$. Fig.~\ref{fig:method}(b) shows how $D$ trades off frequent shallow resets against delayed recovery.

\subsection{Sensitivity-gated sample selection}
SEGA uses $s_b$ to gate which images update the model, replacing SAR-style confidence thresholds~\citep{niu2023sar} and small-batch BN statistics. Low-sensitivity samples produce no useful adaptation signal because the model is already confident and stable on them. Mid-to-high-sensitivity samples carry useful gradient signal for addressing current domain shift. The trend rule handles the separate case where sensitivity remains elevated across many consecutive steps, indicating systemic drift rather than per-sample difficulty. The gate focuses updates on samples whose sensitivity lies \emph{inside} a chosen quantile band, discarding a lowest-sensitivity tail. Let $Q_{\min}$ and $Q_{\max}$ with $0\leq Q_{\min}\leq Q_{\max}\leq 1$ specify a sensitivity-quantile band, and let $U \in \mathbb{N}$ be a warm-up horizon for a reliable empirical estimate before activating the gate. After accumulating sensitivities during the first $U$ test steps, the retained set at time $t$ is defined by
\begin{equation}
\begin{split}
\kappa_t^{\min/\max} &= Q_{Q_{\min/\max}}\bigl(\mathcal{S}_{1:t}\bigr), \\
\mathcal{R}_t &= \left\{ b \in \{1,\dots,B\} : \kappa_t^{\min} \leq s_b \leq \kappa_t^{\max} \right\},
\end{split}
\label{eq:slope-gate}
\end{equation}
where $\mathcal{S}_{1:t}$ is the collection of observed sensitivities up to time $t$ and $Q_{q}(\cdot)$ denotes the empirical $q$-quantile. Given this retained set, SEGA performs one gradient update $\theta_t = \theta_{t-1} - \eta \, \nabla_{\theta} \mathcal{L}_{\mathrm{SEGA}}(\mathcal{B}_t,\mathcal{R}_t)$ per batch whenever $\mathcal{R}_t \neq \emptyset$. The optimizer uses learning rate $\eta=10^{-3}$ and weight decay $\omega=0$. For $B=1$, the gate is binary: with $(Q_{\min},Q_{\max})=(0.2,1.0)$ we skip roughly the lowest-sensitivity $20\%$ of samples and update on all others. Fig.~\ref{fig:method}(c) shows that this narrow skipped tail discards many low-sensitivity samples while retaining almost all total sensitivity, whereas moving the lower quantile too high sacrifices informative samples and increases error.

\subsection{Adaptation objective}
Given the retained set $\mathcal{R}_t$, SEGA minimizes a multi-view consistency and entropy objective, with
\begin{equation}
\begin{split}
\mathcal{L}_{\mathrm{SEGA}} &= \mathcal{L}_{\mathrm{cl}} + \lambda \, \mathcal{L}_{\mathrm{el}}, \\
\mathcal{L}_{\mathrm{cl}} &= \frac{1}{|\mathcal{R}_t|}\sum_{b \in \mathcal{R}_t}\sum_{i=1}^{N-1}\bigl[\mathrm{CE}\!\left(p_b^{(0)}, z_b^{(i)}\right) \\
&\quad + \sum_{j=1}^{i-1} \mathrm{CE}\!\left(p_b^{(j)}, z_b^{(i)}\right)\bigr], \\
\mathcal{L}_{\mathrm{el}} &= \frac{1}{N|\mathcal{R}_t|}\sum_{b \in \mathcal{R}_t}\sum_{i=0}^{N-1} \mathcal{H}\!\left(p_b^{(i)}\right),
\end{split}
\label{eq:sega-loss}
\end{equation}
where $\mathrm{CE}(q,z) = -\sum_{c=1}^{C} q_c \log \sigma(z)_c$ with detached soft targets. $\mathcal{L}_{\mathrm{cl}}$ enforces pairwise predictive consistency across erasing levels, while $\mathcal{L}_{\mathrm{el}}$ drives confident predictions without additional labels or teacher networks. In our default setting, $\lambda=1$. Algorithm~\ref{alg:sega_algorithm} summarizes the full procedure.

\begin{algorithm}[t]
\small
\caption{SEGA: Sensitivity-Guided Erasing Adaptation for strict online CTTA.}
\label{alg:sega_algorithm}
\begin{algorithmic}[1]
\REQUIRE Source model $f_{\theta_0}$, stream $\{x_t\}$, $B{=}1$
\REQUIRE Erasing step $A$, levels $N$, lr $\eta$, decay $\omega$
\REQUIRE EMA $\zeta$, recovery $(T, D)$, gate $(Q_{\min}, Q_{\max}, U)$
\STATE Init $\theta \leftarrow \theta_0$, $\mathcal{S}_{1:0} \leftarrow \emptyset$, $\bar{s}_0$, $s_0^{\star}$
\FOR{$t = 1, 2, \dots$}
  \STATE Receive batch $\mathcal{B}_t = \{x_b\}_{b=1}^{B}$
  \FOR{each $x_b$ in $\mathcal{B}_t$}
    \STATE $z_b^{(0)} = f_{\theta}(x_b)$, $p_b^{(0)} = \sigma(z_b^{(0)})$
    \FOR{$i = 1, \dots, N{-}1$}
      \STATE Sample mask $M_b^{(i)}$ with area $\rho_i = iA$
      \STATE $x_b^{(i)} = x_b \odot (1{-}M_b^{(i)})$
      \STATE $p_b^{(i)} = \sigma(f_{\theta}(x_b^{(i)}))$
    \ENDFOR
    \STATE $s_b \leftarrow \tfrac{1}{N-1}\sum_{i=1}^{N-1} \bigl\lvert \mathcal{H}(p_b^{(i)}){-}\mathcal{H}(p_b^{(i-1)}) \bigr\rvert$
    \STATE Append $s_b$ to $\mathcal{S}_{1:t}$
  \ENDFOR
  \STATE \textbf{Recovery:}
  \STATE $\tilde{s}_t = \tfrac{1}{B}\sum_{b=1}^{B} s_b$
  \STATE $\bar{s}_t = \zeta \bar{s}_{t-1} + (1{-}\zeta)\tilde{s}_t$
  \STATE $s_t^{\star} = \min_{\tau \leq t} \bar{s}_{\tau}$
  \IF{$t \geq T$ \textbf{and} $\bar{s}_t > (1{+}D)s_t^{\star}$}
    \STATE Reset $\theta \leftarrow \theta_0$
  \ENDIF
  \STATE \textbf{Gate:}
  \IF{$t > U$}
    \STATE $\kappa_t^{\min}, \kappa_t^{\max} \leftarrow$ quantiles of $\mathcal{S}_{1:t}$
    \STATE $\mathcal{R}_t = \{b : \kappa_t^{\min} \leq s_b \leq \kappa_t^{\max}\}$
  \ELSE
    \STATE $\mathcal{R}_t = \{1, \dots, B\}$
  \ENDIF
  \IF{$\mathcal{R}_t = \emptyset$}
    \STATE Skip update
  \ENDIF
  \STATE \textbf{Update:}
  \IF{$\mathcal{R}_t \neq \emptyset$}
    \STATE $\mathcal{L} = \mathcal{L}_{\mathrm{cl}} + \lambda \mathcal{L}_{\mathrm{el}}$ over $b \in \mathcal{R}_t$
    \STATE $\theta \leftarrow \theta - \eta \nabla_{\theta} \mathcal{L}$ \COMMENT{decay $\omega$}
  \ENDIF
\ENDFOR
\end{algorithmic}
\end{algorithm}

\section{Experiments}
\label{sec:experiments}

\subsection{Dataset}
\label{sec:dataset}
We evaluate on CIFAR100/10-C and ImageNet-C~\citep{hendrycks2019benchmarking}, derived from CIFAR~\citep{krizhevsky2009cifar} and ImageNet~\citep{deng2009imagenet}. Each contains 15 corruption types at 5 severity levels. Following prior work~\citep{Wangetal2022cotta,liu2023vida,liu2024continual,Han2025RankedEM}, we use severity level~5 and report online classification error after adaptation, with 10{,}000 test samples per corruption domain in CIFAR-10-C and 5{,}000 in ImageNet-C. To probe the same long-horizon failure mode in a domain-specific setting, we also build corruption-generated aquaculture proxy streams from FreshFish~\citep{BiswasFishDisease2024} by applying the same 15 synthetic corruptions at severity level~5 to the clean test split. We fine-tune a ViT-B/16 backbone on the clean dataset, obtaining 99.57\% accuracy on FreshFish. This controlled proxy allows us to study batch-size-1 collapse and recovery on aquaculture imagery, although it does not fully capture real monitoring shifts such as gradual lighting changes or biofouling. It is included to test the paper's main question of whether collapse can be detected and controlled over long horizons in the same strict online regime used on the standard corruption benchmarks.\footnote{These corruptions map to plausible aquaculture monitoring artifacts: defocus and motion blur from camera shake, Gaussian and shot noise from sensor noise, brightness and contrast shifts from variable lighting, JPEG compression from bandwidth-limited transmission, and pixelation from distant subjects.} Corruption abbreviations used in figures and tables: GN=Gaussian noise, SN=Shot noise, IN=Impulse noise, DB=Defocus blur, GB=Glass blur, MB=Motion blur, ZB=Zoom blur, S=Snow, Fr=Frost, F=Fog, B=Brightness, C=Contrast, ET=Elastic transform, P=Pixelate, JC=JPEG compression.

\subsection{Implementation Details}
Our main experiments use a ViT-B/16 backbone~\citep{dosovitskiy2021vit}. Baselines are instantiated from the public implementation of Marsden \etal~\citep{Marsden2023UniversalTA}\footnote{https://github.com/mariodoebler/test-time-adaptation.git}, with RES.TTA~\citep{Vray2025ResTTA} run on the SAR base model~\citep{niu2023sar}. We sweep Adam and SGD with learning rates $10^{-3}$, $10^{-4}$, $10^{-5}$ on ImageNet-C and report the best configuration. For SEGA: one gradient step per batch, $A{=}0.1$, $N{=}3$, $\lambda{=}1$, $(Q_{\min},Q_{\max},U){=}(0.2,1.0,2048)$, $(T,D){=}(50,6.0)$, $B{=}1$. These settings were tuned on ImageNet-C (Fig.~\ref{fig:ablation}) and frozen for all other datasets because the paper evaluates one strict online regime rather than re-optimizing for multiple deployment settings. All runs use a single AMD Instinct MI200 GPU.

\subsection{Results}
\label{sec:results}

\begin{table*}[t!]
    \centering
    \small
    \caption{Classification error rate (\%) under CTTA (batch size 1) on ImageNet-C and CIFAR10-C using ViT-Base. Mean is the average across 15 corruption domains, each with 5000 test samples. Gain is the relative improvement over the source model. Entries report mean of three runs (seeds 1, 2, 3). Fine-grained result is available in the supp. material.}
    \label{tab:imagenetc_cifar10c_ctta_avg}
    \setlength\tabcolsep{6pt}
    \resizebox{0.9\linewidth}{!}{
    \begin{tabular}{l|ccc|cccc|cccc|cccc|cc}
        \toprule
        Time & \multicolumn{15}{c|}{$t\xrightarrow{\;}$}& \\ \hline
        Method &
        \rotatebox[origin=c]{0}{GN} & \rotatebox[origin=c]{0}{SN} & \rotatebox[origin=c]{0}{IN} & \rotatebox[origin=c]{0}{DB} & \rotatebox[origin=c]{0}{GB} & \rotatebox[origin=c]{0}{MB} & \rotatebox[origin=c]{0}{ZB} & \rotatebox[origin=c]{0}{S} & \rotatebox[origin=c]{0}{Fr} & \rotatebox[origin=c]{0}{F}  & \rotatebox[origin=c]{0}{B} & \rotatebox[origin=c]{0}{C} & \rotatebox[origin=c]{0}{ET} & \rotatebox[origin=c]{0}{P} & \rotatebox[origin=c]{0}{JC}
        & Mean$\downarrow$ & Gain$\uparrow$\\\hline

        & \multicolumn{15}{c}{\textbf{ImageNet-C}} &  & \\\hline
        SOURCE (2021)~\citep{dosovitskiy2021vit}   & 53.0 & 51.8 & 52.1 & 68.5 & 78.8 & 58.5 & 63.3 & 49.9 & 54.2 & 57.7 & 26.4 & 91.4 & 57.5 & 38.0 & 36.2 & 55.8 & 0.0\\
        ROTTA (2023)~\citep{Yuan2023RobustTA}      & 49.9 & 48.2 & 48.0 & 67.9 & 70.5 & 55.1 & 60.0 & 45.5 & 48.7 & 53.4 & 24.9 & 88.4 & 54.0 & 37.2 & 34.6 & 52.4 & +3.4\\
        RPL (2021)~\citep{Rusak2021IfYD}           & 46.6 & 40.6 & 40.6 & 55.7 & 55.9 & 44.6 & 51.1 & 41.0 & 42.5 & 40.9 & 23.5 & 69.0 & 64.6 & 54.9 & 54.3 & 48.4 & +7.4\\
        SANTA (2023)~\citep{Chakrabarty2023SATASA} & 46.0 & 43.2 & 44.6 & 58.7 & 64.4 & 47.7 & 54.1 & 40.0 & 46.6 & 41.9 & 23.3 & 89.5 & 49.1 & 34.7 & 33.5 & 47.8 & +8.0\\
        TENT (2021)~\citep{DequanWangetal2021}     & 47.4 & 41.8 & 41.7 & 57.5 & 58.4 & 45.8 & 51.5 & 40.4 & 41.9 & 41.8 & 22.5 & \bf 54.0 & 50.2 & 31.9 & 30.9 & 43.8 & +12.0\\
        LCOTTA (2025)~\citep{Duan2025}             & 41.5 & 37.6 & 39.3 & 54.9 & 54.4 & 46.4 & 51.8 & 41.1 & 40.5 & 42.2 & 24.2 & 58.7 & 47.7 & 33.1 & 32.1 & 43.0 & +12.8\\
        RES.TTA (2025)~\citep{Vray2025ResTTA} & 41.9 & 39.5 & 39.5 & 50.0 & 47.5 & 45.0 & 49.8 & 39.0 & 39.4 & 37.3 & 21.9 & 88.7 & 38.5 & 29.3 & 29.9 & 42.5 & +13.3\\
        
        SAR (2023)~\citep{niu2023sar}              & 41.4 & 38.3 & 39.4 & 54.8 & 45.3 & 42.9 & 43.2 & 39.3 & 41.5 & 82.0 & 28.7 & 99.6 & 37.9 & \bf 27.7 & 28.8 & 46.1 & +9.7\\
        RDUMB (2023)~\citep{Press2023RDumbAS}      & 44.0 & 42.4 & 42.4 & 49.5 & 48.2 & 43.6 & 46.2 & 35.1 & 37.9 & 37.0 & 21.6 & 69.4 & 37.4 & 27.9 & 29.5 & 40.8 & +15.0\\
        M2A (2026)~\citep{Doloriel2026Family}      & 41.2 & \bf 35.8 & \bf 36.2 & 49.6 & 44.9 & \bf 38.7 & 41.1 & 34.1 & \bf 33.7 & \bf 32.7 & 21.2 & 64.6 & 57.4 & 51.6 & 52.5 & 42.4 & +13.4\\
        SEGA                                       & \bf 40.5 & 37.3 & 38.3 & \bf 47.7 & \bf 42.1 & 38.9 & \bf 39.4 & \bf 31.9 & 39.7 & 36.3 & \bf 21.1 & 55.6 & \bf 31.6 & \bf 27.7 & \bf 28.4 & \bf 37.1 & +18.7\\
        \midrule

        & \multicolumn{15}{c}{\textbf{CIFAR10-C}} &  & \\\hline
        SOURCE (2021)~\citep{dosovitskiy2021vit}   & 60.1 & 53.2 & 38.3 & 19.9 & 35.5 & 22.6 & 18.6 & 12.1 & 12.7 & 22.8 & 5.3 & 49.7 & 23.6 & 24.7 & 23.1 & 28.2 & 0.0\\
        ROTTA (2023)~\citep{Yuan2023RobustTA}      & 60.9 & 56.2 & 38.4 & 18.4 & 34.2 & 17.5 & 13.1 & 10.8 & 12.1 & 14.0 & \bf 4.2 & 21.1 & 14.8 & 27.5 & 19.8 & 24.2 & +4.0\\
        TENT (2021)~\citep{DequanWangetal2021}     & 59.4 & 59.0 & 30.6 & 16.7 & 35.2 & 16.8 & 12.7 & 10.7 & 12.1 & 15.6 & 4.8 & 26.7 & 16.9 & 28.6 & 20.6 & 24.4 & +3.8\\
        RPL (2021)~\citep{Rusak2021IfYD}           & 52.3 & 38.7 & 27.1 & 18.6 & 33.4 & 19.3 & 13.7 & 10.8 & 10.4 & 16.9 & 4.6 & 27.4 & 18.1 & 29.7 & 21.3 & 22.8 & +5.4\\
        LCOTTA (2025)~\citep{Duan2025}             & 57.0 & 50.2 & 37.4 & 20.1 & 35.1 & 22.8 & 18.8 & 11.8 & 12.5 & 22.9 & 5.3 & 50.5 & 23.5 & 25.0 & 22.9 & 27.7 & +0.5\\
        RES.TTA (2025)~\citep{Vray2025ResTTA} & 63.2 & 59.7 & 38.5 & 13.5 & 42.7 & 16.5 & 14.0 & 13.5 & 15.2 & 13.3 & 4.5 & 15.4 & 23.5 & 27.3 & 28.4 & 25.9 & +2.3\\
        SAR (2023)~\citep{niu2023sar}              & 55.0 & 49.4 & 38.0 & 19.5 & 35.1 & 22.3 & 18.3 & 12.0 & 12.7 & 22.4 & 5.3 & 41.7 & 23.1 & 24.2 & 23.0 & 26.9 & +1.3\\
        RDUMB (2023)~\citep{Press2023RDumbAS}      & 27.7 & 24.0 & 15.1 & 11.5 & 28.0 & 13.1 & 10.8 & 9.3 & 9.2 & 13.0 & 5.0 & 16.8 & 17.1 & 12.7 & 19.4 & 15.5 & +12.7\\
        M2A (2026)~\citep{Doloriel2026Family}      & 23.4 & \bf 13.9 & 11.7 & 9.6 & 22.9 & 9.7 & \bf 6.3 & \bf 7.4 & \bf 5.9 & \bf 8.1 & 4.3 & \bf 6.5 & \bf 12.4 & 10.0 & 15.6 & 11.1 & +17.1\\
        SEGA                                       & \bf 15.8 & 14.0 & \bf 9.8 & \bf 9.3 & \bf 22.8 & \bf 9.4 & 7.7 & 7.5 & 7.5 & 8.8 & 4.4 & 8.4 & 13.2 & \bf 9.9 & \bf 13.8 & \bf 10.8 & +17.4\\
        
        \bottomrule
    \end{tabular}%
    }

\end{table*}

\begin{figure}[t!]
    \centering
    \footnotesize
    \subfigure[CIFAR100-C]{
        \includegraphics[width=0.45\linewidth]{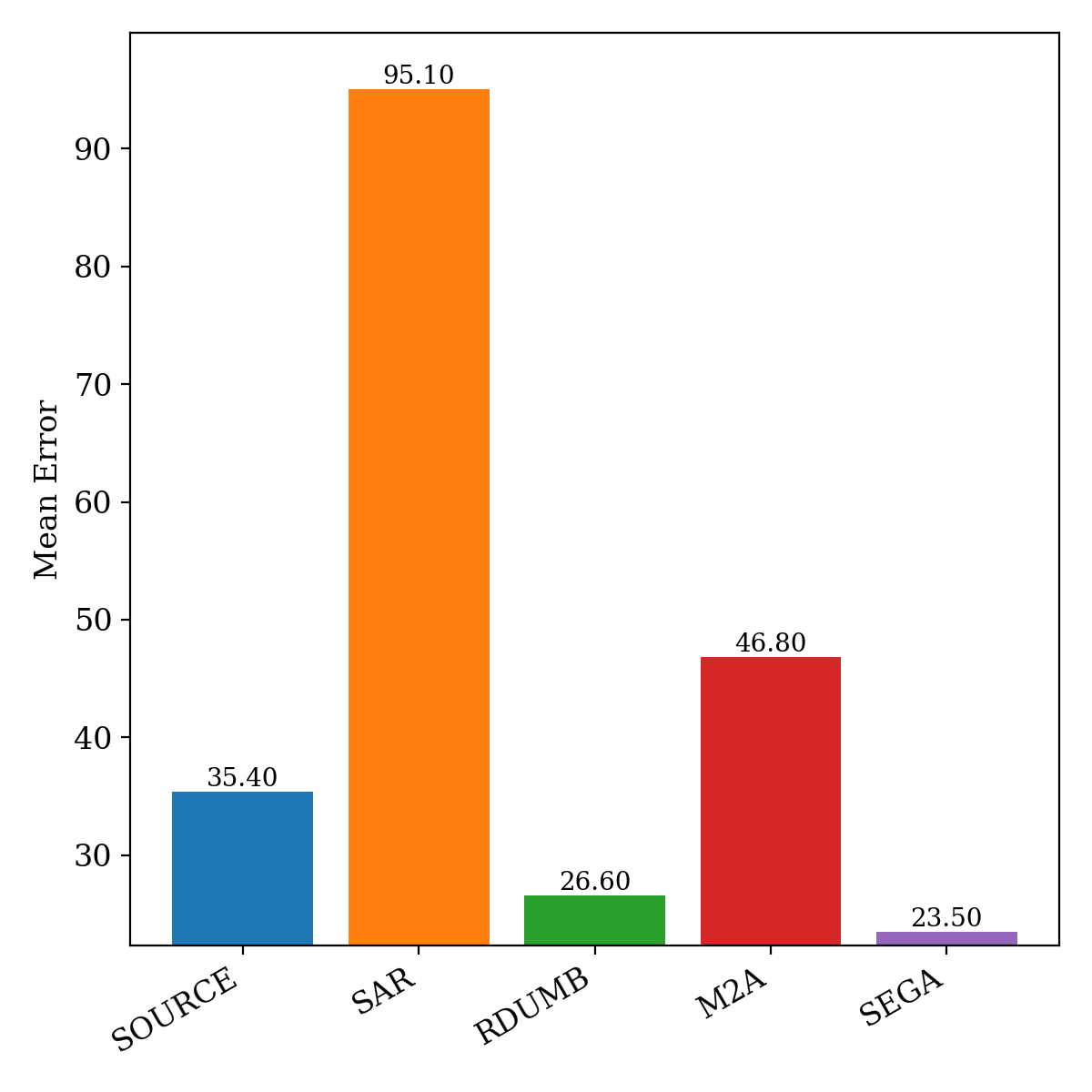}
    }
    \subfigure[FreshFish-C]{
        \includegraphics[width=0.45\linewidth]{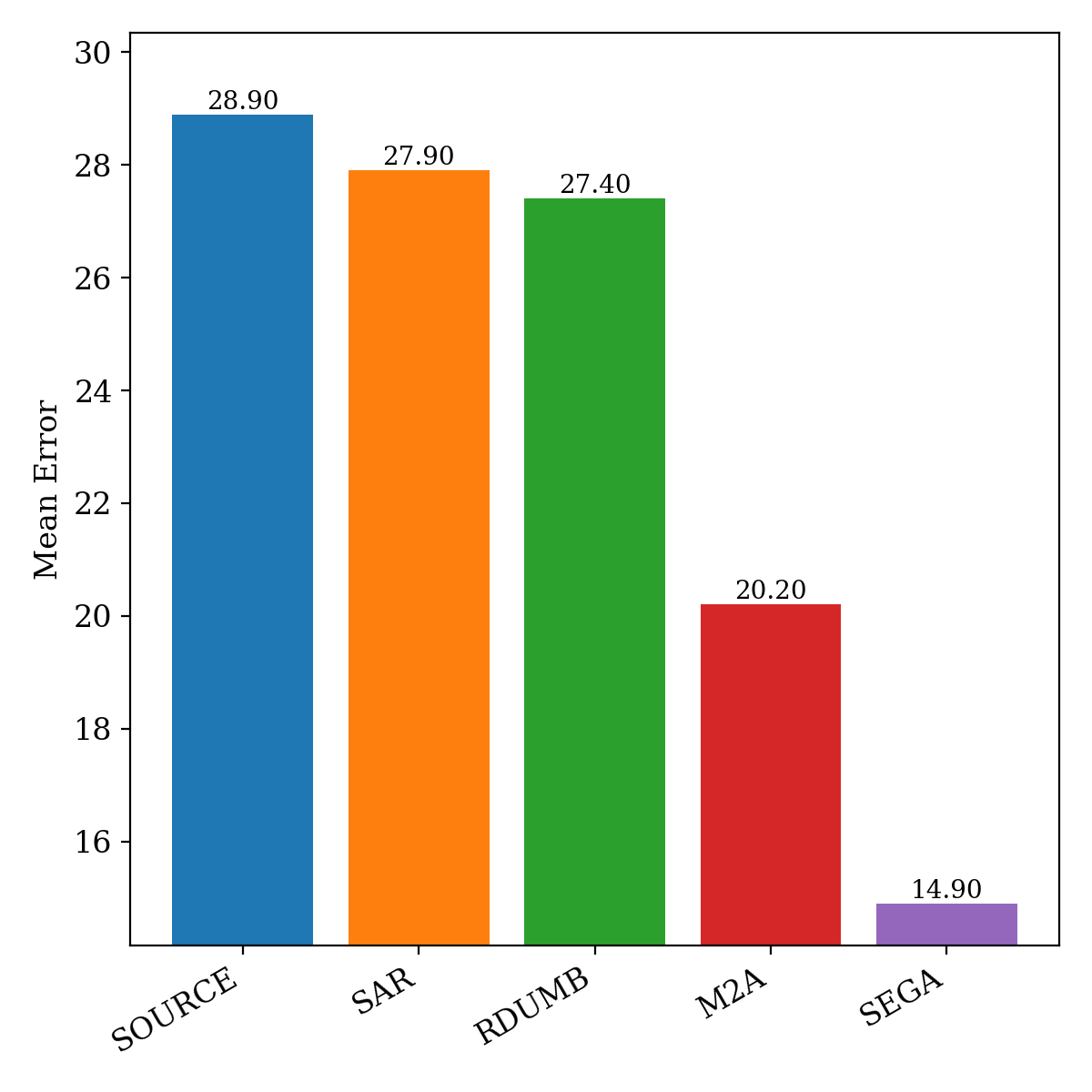}
    }

    \caption{Mean classification error rates (\%) on CIFAR100-C and FreshFish-C under CTTA (batch size 1) using ViT-Base. Entries report mean of three runs (seeds 1, 2, 3). Fine-grained results are available in the supplementary material.}
    \label{fig:benchmark}
\end{figure}

\paragraph{Benchmarks.} Table~\ref{tab:imagenetc_cifar10c_ctta_avg} shows that SEGA achieves the lowest mean error on the two primary strict online CTTA benchmarks, reaching $37.1\%$ on ImageNet-C and $10.8\%$ on CIFAR10-C, with the largest gains over the source model at $+18.7$ and $+17.4$~pp. The main advantage is stability on hard corruptions rather than isolated wins on easy ones. On ImageNet-C, SAR spikes to $82.0\%$ on Fog and $99.6\%$ on Contrast, while M2A remains competitive on several groups but rises to $51.6\%$ on Pixelate and $52.5\%$ on JPEG. SEGA wins $8$ of $15$ corruptions on each dataset and keeps ImageNet-C below $56\%$ on every corruption, which explains why it achieves the best stream mean.

\paragraph{Additional Benchmarks.} Fig.~\ref{fig:benchmark} shows the same trend on CIFAR100-C and FreshFish-C, where SEGA again attains the lowest mean error. This shows that the gains are not confined to ImageNet-C and CIFAR10-C, but extend to an additional corruption benchmark and to an aquaculture stream built from the same corruption protocol.

\begin{table}[ht!]
    \centering
    \caption{Continual Dynamic Change (CDC) on ImageNet-C and CIFAR10-C; GN-SN, DB-ZB, S-B, and C-JC are grouped by noise, blur, weather, and digital corruptions.}
    \label{tab:cdc}
    \footnotesize
    \setlength\tabcolsep{8pt}
    \begin{adjustbox}{width=\linewidth,center=\linewidth}
    \begin{tabular}{l|cccc|cc|cccc|cc}
        \toprule
        \multirow{3}{*}{Method} & \multicolumn{6}{c|}{\textbf{ImageNet-C}} & \multicolumn{6}{c}{\textbf{CIFAR10-C}} \\
        \cmidrule(lr){2-7} \cmidrule(lr){8-13}
         & GN-SN & DB-ZB & S-B & C-JC & Mean$\downarrow$ & Gain$\uparrow$ & GN-SN & DB-ZB & S-B & C-JC & Mean$\downarrow$ & Gain$\uparrow$ \\
         \midrule
        SOURCE          & 49.3 & 64.4 & 45.1 & 55.7 & 53.6     & 0.0   & 50.5 & 24.2 & 13.2 & 30.3 & 29.6 & 0.0 \\
        SAR             & 44.5 & 46.5 & 37.5 & 47.1 & 43.9     & +9.7  & 47.5 & 23.8 & 13.1 & 28.5 & 28.2 & +1.4 \\
        RDUMB           & 42.7 & 47.4 & \bf 33.9 & 41.9 & 41.5     & +12.1 & 23.0 & 16.6 & 9.5 & 17.5 & 16.7 & +12.9 \\
        M2A             & 78.8 & 93.8 & 99.0 & 86.8 & 89.6     & -36.0 & 18.1 & \bf 11.7 & \bf 7.0 & 12.9 & 12.4 & +17.2 \\
        SEGA            & \bf 40.6 & \bf 43.8 & 34.5 & \bf 39.7 & \bf 39.7     & +13.9 & \bf 15.8 & 12.7 & 7.3 & \bf 11.6 & \bf 11.8 & +17.8 \\

        \bottomrule
    \end{tabular}
    \end{adjustbox}
\end{table}
\paragraph{Continual Dynamic Change.} Table~\ref{tab:cdc} isolates the strongest stability test by interleaving corruption groups sample-wise. In this CDC setting, SEGA is the only method with the best mean on both datasets, reaching $39.7\%$ on ImageNet-C and $11.8\%$ on CIFAR10-C. The gap on ImageNet-C is especially important: M2A collapses to $89.6\%$ mean, whereas SEGA wins the noise, blur, and digital groups and remains close to RDumb on weather. On CIFAR10-C, M2A wins the blur and weather groups, but SEGA still achieves the best overall mean because its performance is more balanced across all four groups.

\begin{table}[t!]
    \centering
    \caption{Adaptation efficiency on ImageNet-C and CIFAR10-C under CTTA (batch size 1) with ViT-Base on a single AMD Instinct MI200 GPU. Forward/Backward entries use $K$ to denote $5000$ samples per corruption domain (e.g., $1K$, $3.41K$). Note: ImageNet-C and CIFAR10-C have 5,000 and 10,000 samples per domain, respectively.}
    \label{tab:efficiency}
    \footnotesize
    \setlength\tabcolsep{4pt}
    \begin{adjustbox}{width=\linewidth,center=\linewidth}
    \begin{tabular}{l|cccc|c|cccc|c}
        \toprule
        \multirow{3}{*}{Method} & \multicolumn{5}{c|}{\textbf{ImageNet-C}} & \multicolumn{5}{c}{\textbf{CIFAR10-C}} \\
        \cmidrule(lr){2-6} \cmidrule(lr){7-11}
         & Param(\%) & Time(s) & Forward & Backward & Error$\downarrow$ & Params(\%) & Time(s) & Forward & Backward & Error$\downarrow$ \\
         \midrule
        SOURCE          & 0    & 43.8  & 1K     & 0    & 55.8       & 0    & 103.7 & 2K     & 0    & 28.2 \\
        SAR             & 0.03 & 163.1 & 2K     & 1.56K & 46.1     & 0.03 & 442.7 & 4K     & 3.41K & 26.9 \\
        RES.TTA         & 0.03 & 236.8 & 3K     & 1.12K & 42.5     & 0.03 & 583.1 & 6K     & 2.20K & 25.9 \\
        RDUMB           & 0.05 & 87.13 & 1K     & 0.83K & 40.8     & 0.05 & 195.3 & 2K     & 1.40K & 15.5 \\
        M2A             & 0.03 & 231.4 & 3K     & 1K    & 42.4     & 0.03 & 627.7 & 6K     & 2K    & 11.1 \\
        SEGA            & 0.03 & 224.0 & 3K     & 0.69K & 37.1     & 0.03 & 604.2 & 6K     & 1.32K & 10.8 \\
        \bottomrule
    \end{tabular}
    \end{adjustbox}
\end{table}
\begin{table}[t]
    \centering\footnotesize
    \setlength\tabcolsep{4pt}
    \caption{Buffer-based adaptation vs.\ SEGA ($B{=}1$). Error (\%), avg.\ seeds 1/2/3. SEGA: 37.1\% (ImageNet-C), 10.8\% (CIFAR10-C). ``-'' = OOM.}
    \label{tab:buffer}
    \begin{tabular}{l|cccc}
        \toprule
        Dataset & $K{=}8$ & $K{=}16$ & $K{=}32$ & $K{=}64$ \\
        \midrule
        ImageNet-C & 42.3 & 50.1 & 51.0 & 49.7 \\
        CIFAR10-C & 16.6 & 18.8 & - & - \\
        \bottomrule
    \end{tabular}
\end{table}

\paragraph{Efficiency and Buffering.} Table~\ref{tab:efficiency} shows that SEGA does not win by spending a larger adaptation budget. It updates only $0.03\%$ of parameters, matching SAR and M2A, and uses the same number of forward passes as M2A, but cuts backward passes from $1K$ to $0.69K$ on ImageNet-C and from $2K$ to $1.32K$ on CIFAR10-C. This yields lower error than M2A at similar wall-clock time. Table~\ref{tab:buffer} further supports the strict online setting: even the smallest buffer, $K{=}8$, is still $5.2$~pp worse than SEGA on ImageNet-C and $5.8$~pp worse on CIFAR10-C, while larger buffers degrade further and $K{\geq}32$ runs out of memory on CIFAR10-C.

\begin{figure*}[t!]
    \centering
    \small
    \includegraphics[width=0.75\textwidth]{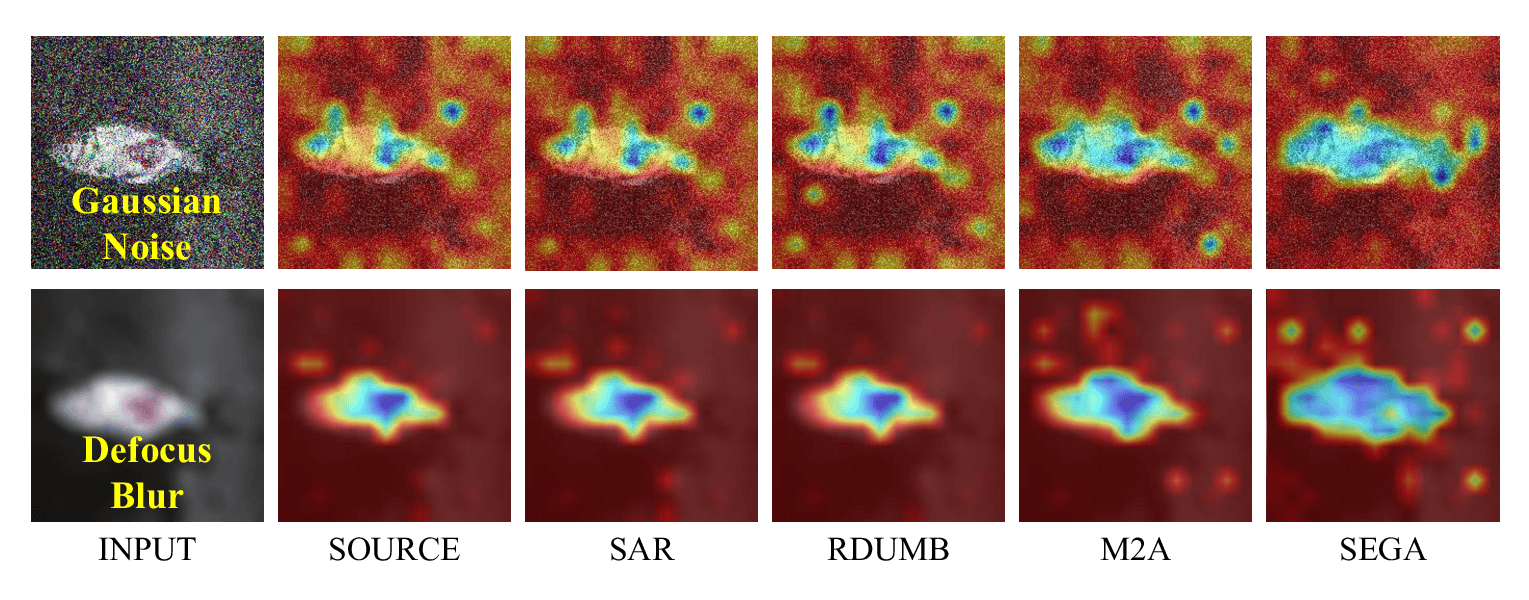}
    \caption{Visualization of class activation maps using GradCAM on corrupted samples from the FreshFish-C on severity level 5.}
    \label{fig:cam}
\end{figure*}

\begin{table}[t!]
    \centering
    \caption{Variable batch size (random between 1 to 128 per time step) under extended CTTA on FreshFish-C using ViT-Base. Entries report error rate (\%).}
    \label{tab:randbatch_lifelong}
    \footnotesize
    \setlength\tabcolsep{4pt}
    \begin{adjustbox}{width=\linewidth,center=\linewidth}
    \begin{tabular}{l|ccccc|cc}
        \toprule
        \multirow{3}{*}{Method} & \multicolumn{7}{c}{\textbf{FreshFish-C}} \\
        \cmidrule(lr){2-8}
         & Pass 1 & Pass 2 & Pass 3 & Pass 4 & Pass 5 & Mean$\downarrow$ & Gain$\uparrow$ \\
         \midrule
        SOURCE          & 28.9 & 28.9 & 28.9 & 28.9 & 28.9 & 28.9 & 0.0 \\
        SAR             & 28.4 & 28.4 & 28.2 & 28.5 & 28.4 & 28.4 & +0.5 \\
        RDUMB           & 27.8 & 27.7 & 27.8 & 29.1 & 29.1 & 28.3 & +0.6 \\
        M2A             & 27.7 & 26.0 & 24.4 & 23.4 & 22.4 & 24.8 & +4.1 \\
        SEGA            & \bf 22.5 & \bf 16.9 & \bf 13.5 & \bf 12.0 & \bf 11.2 & \bf 15.2 & +13.7 \\

        \bottomrule
    \end{tabular}
    \end{adjustbox}
\end{table}
\paragraph{FreshFish-C Transfer.} Figure~\ref{fig:cam} and Table~\ref{tab:randbatch_lifelong} show that SEGA remains useful beyond the fixed-$B{=}1$ regime within the same corruption-style proxy family for which it was tuned. Under variable batch sizes between $1$ and $128$, SEGA improves monotonically across five passes on FreshFish-C from $22.5\%$ to $11.2\%$, reducing the mean to $15.2\%$. By contrast, SAR and RDumb stay near $28.4\%$ and $28.3\%$, while M2A improves gradually but still averages $24.8\%$. The qualitative maps in Fig.~\ref{fig:cam} are consistent with this trend, showing attention that stays concentrated on fish regions rather than drifting into the background.

\begin{table}[t]
    \centering\footnotesize
    \setlength\tabcolsep{4pt}
    \caption{Augmentation comparison on ImageNet-C, CTTA ($B{=}1$), ViT-Base. Error (\%), avg.\ seeds 1/2/3.}
    \label{tab:augmentation}
    \begin{tabular}{l|c}
        \toprule
        Augmentation & Mean Error$\downarrow$ \\
        \midrule
        Random Crop & 50.5 \\
        Random Rotate & 63.8 \\
        Random Translate & 71.3 \\
        Random Erasing (Frequency) & 42.3 \\
        Random Erasing (Spatial) & \textbf{37.1} \\
        \bottomrule
    \end{tabular}
\end{table}

\begin{figure*}[t!]
    \centering
    \small
    \subfigure[Erasing ratio $A$]{%
        \includegraphics[width=0.21\textwidth]{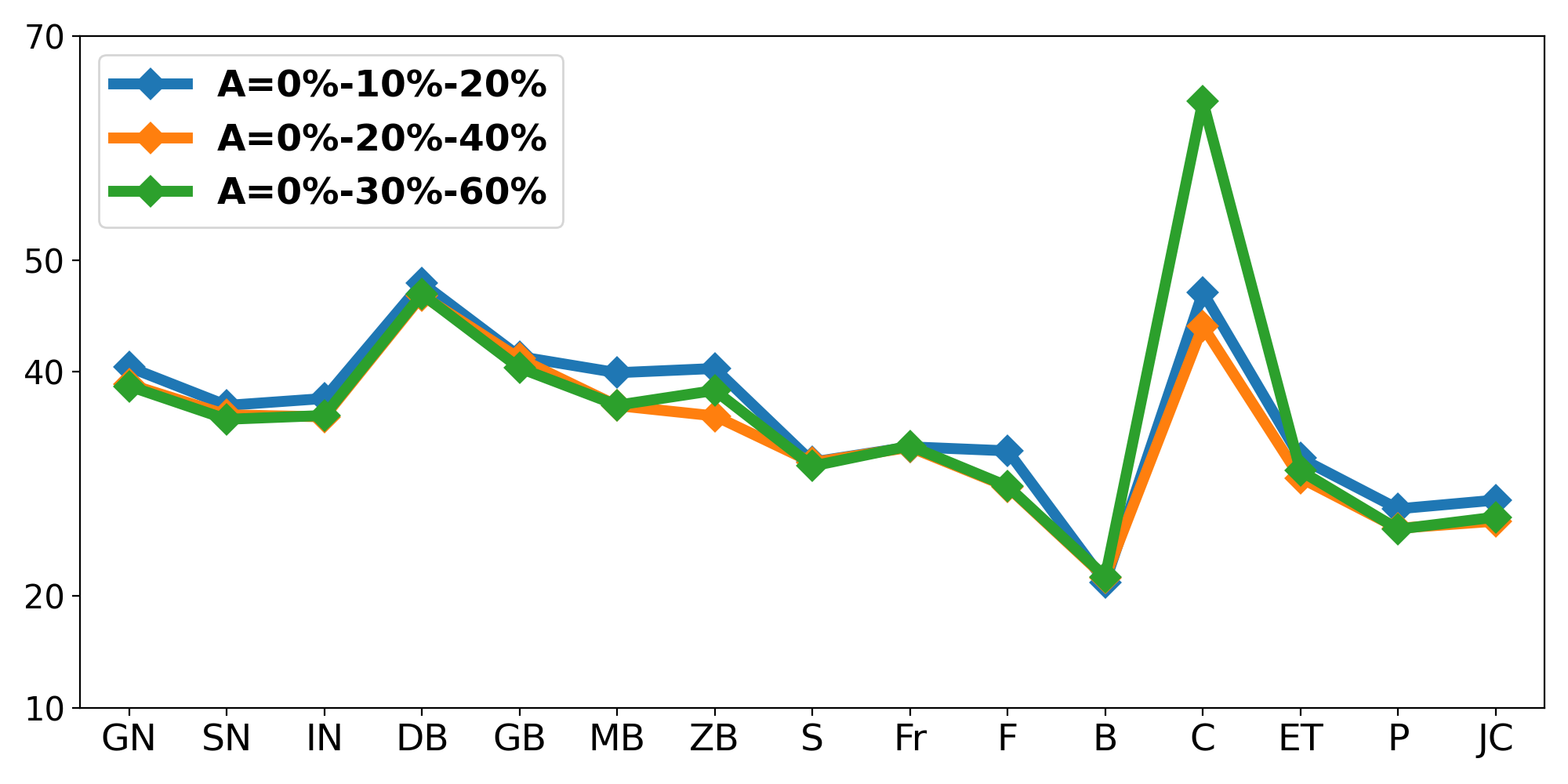}%
    }
    \subfigure[Erasing levels $N$]{%
        \includegraphics[width=0.21\textwidth]{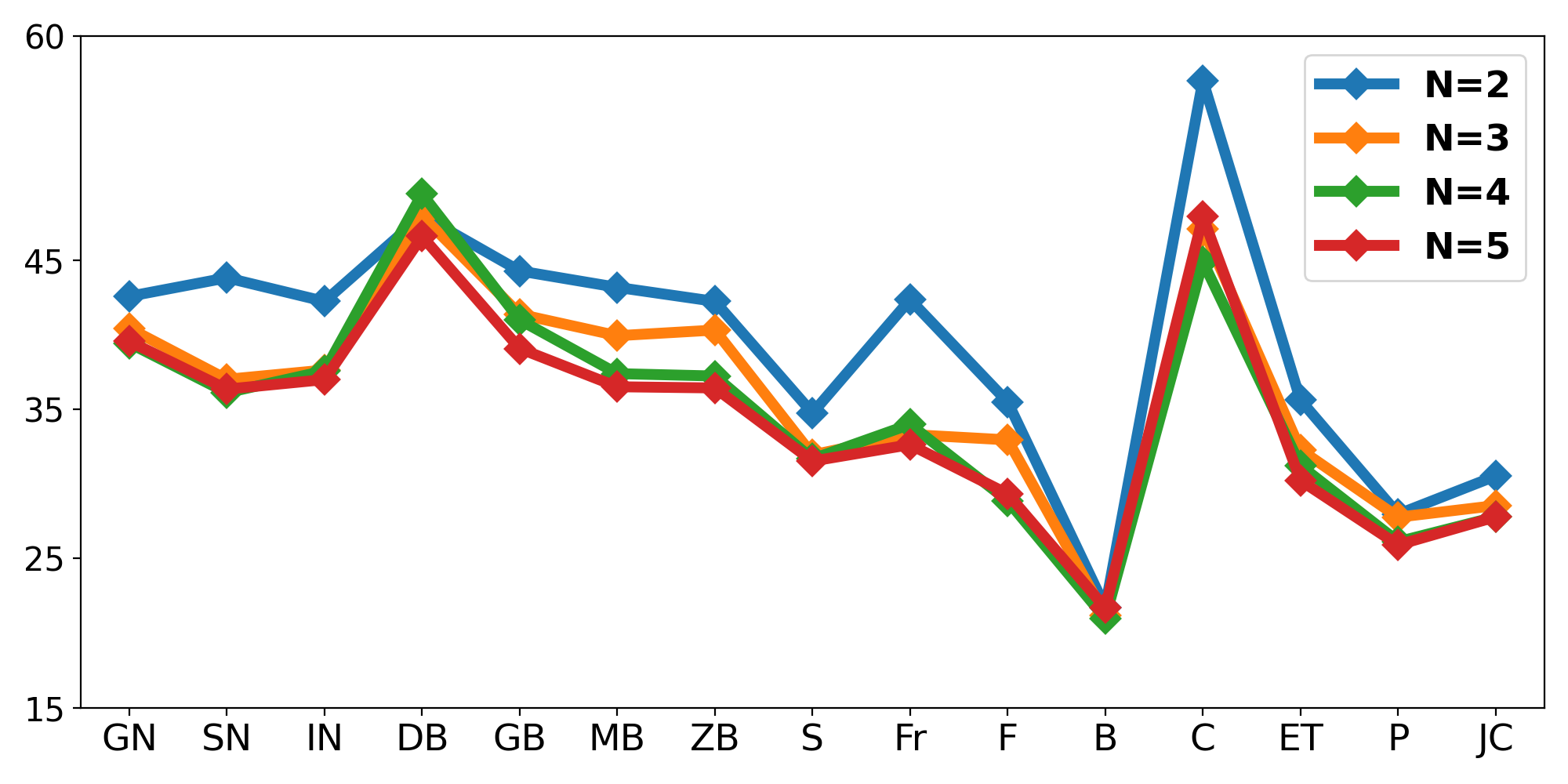}%
    }
    \subfigure[Quantile level $Q$]{%
        \includegraphics[width=0.21\textwidth]{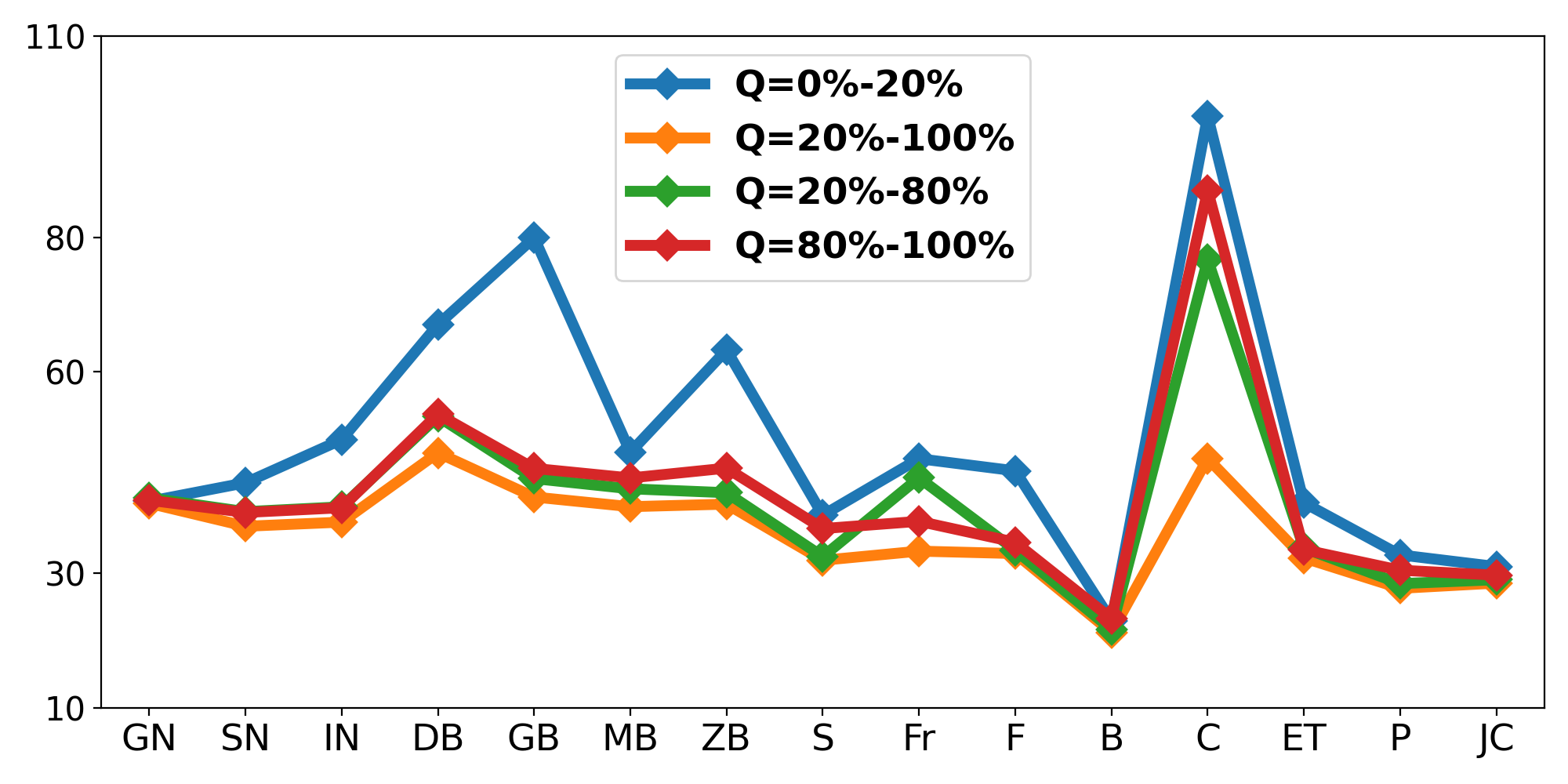}%
    }
    \subfigure[Loss coeff. $\lambda$]{%
        \includegraphics[width=0.21\textwidth]{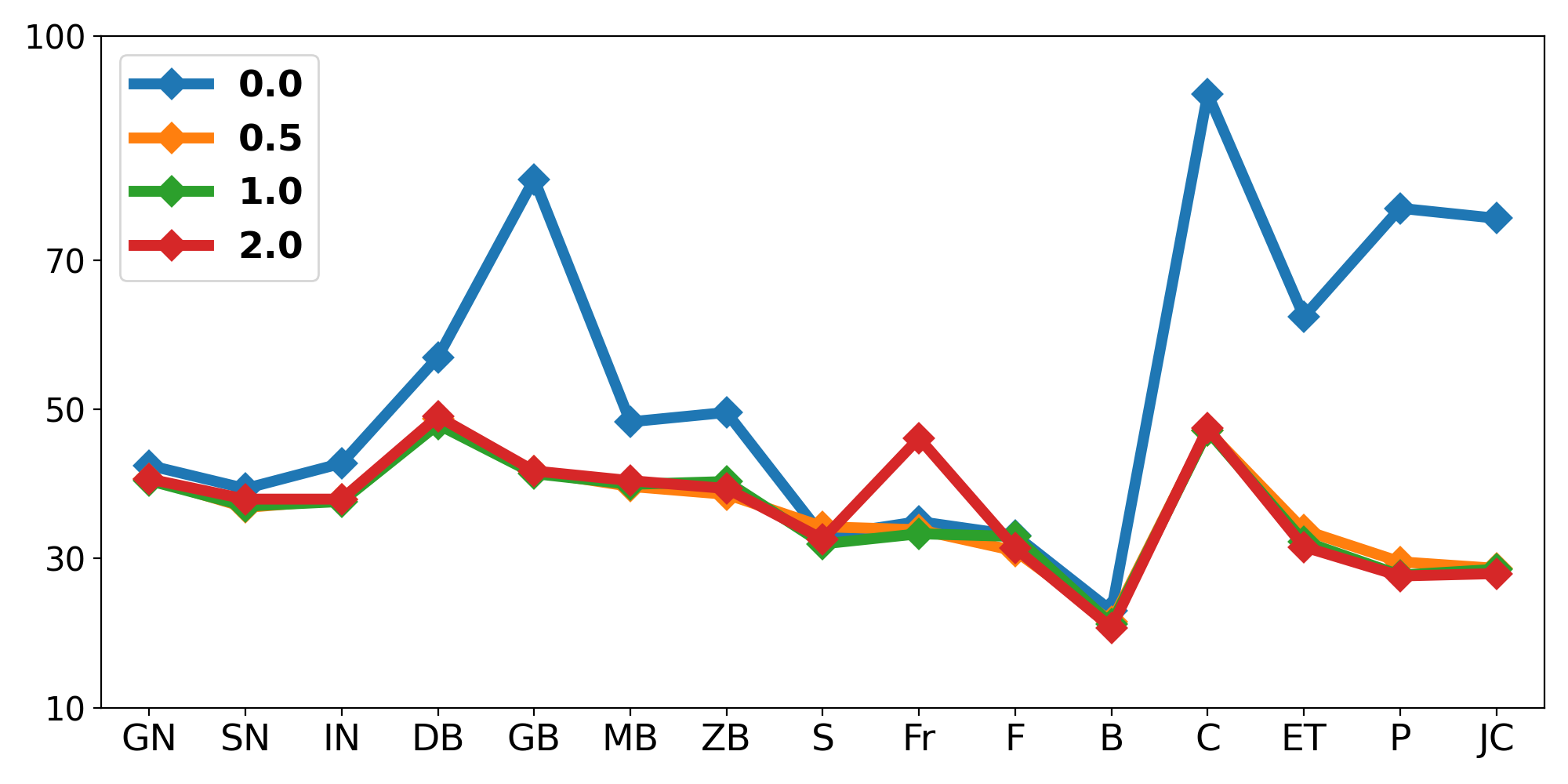}%
    }\\[0.75ex]
    \subfigure[Warm-up steps $U$]{%
        \includegraphics[width=0.21\textwidth]{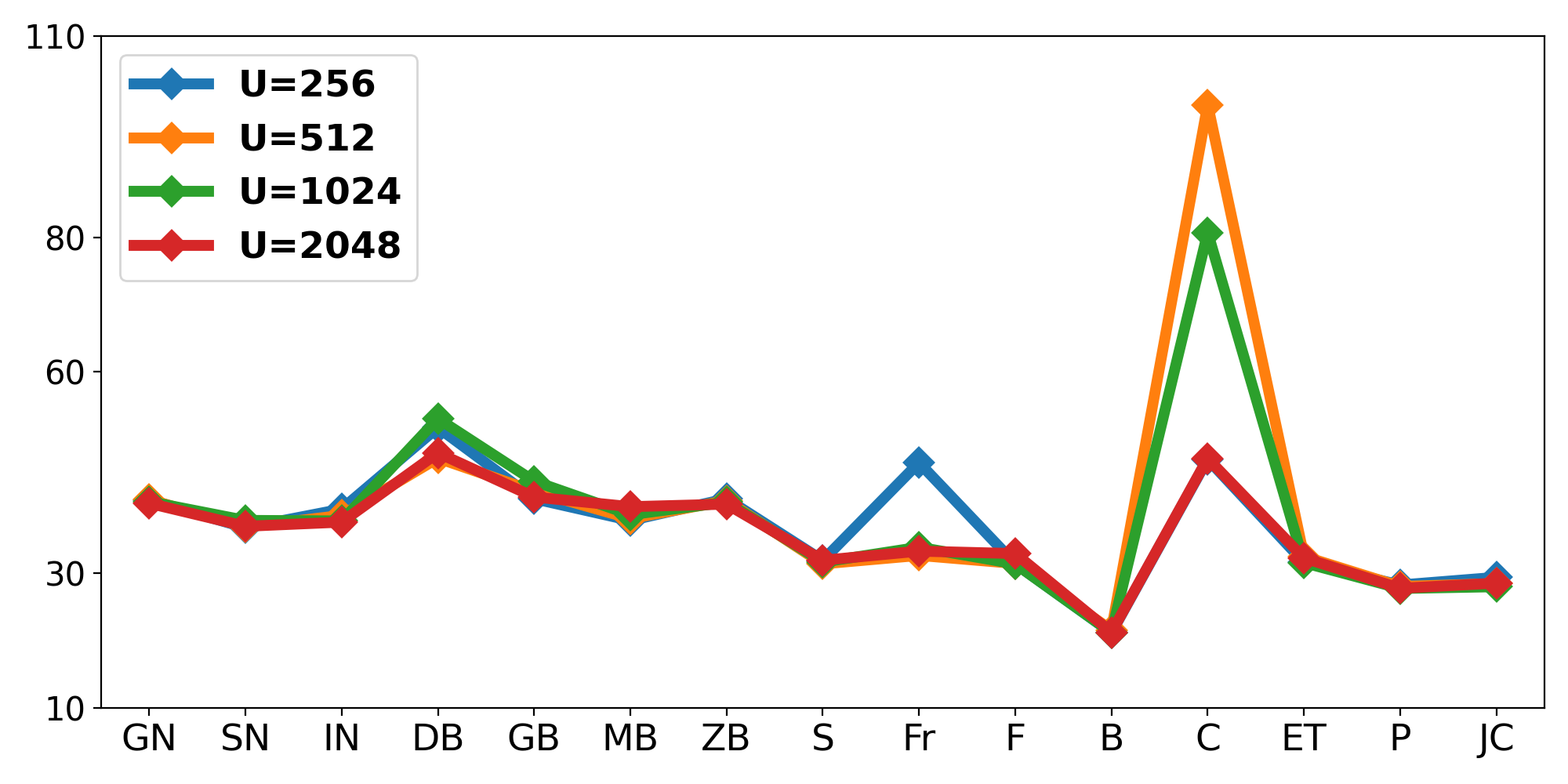}%
    }
    \subfigure[Deter. margin $D$]{%
        \includegraphics[width=0.21\textwidth]{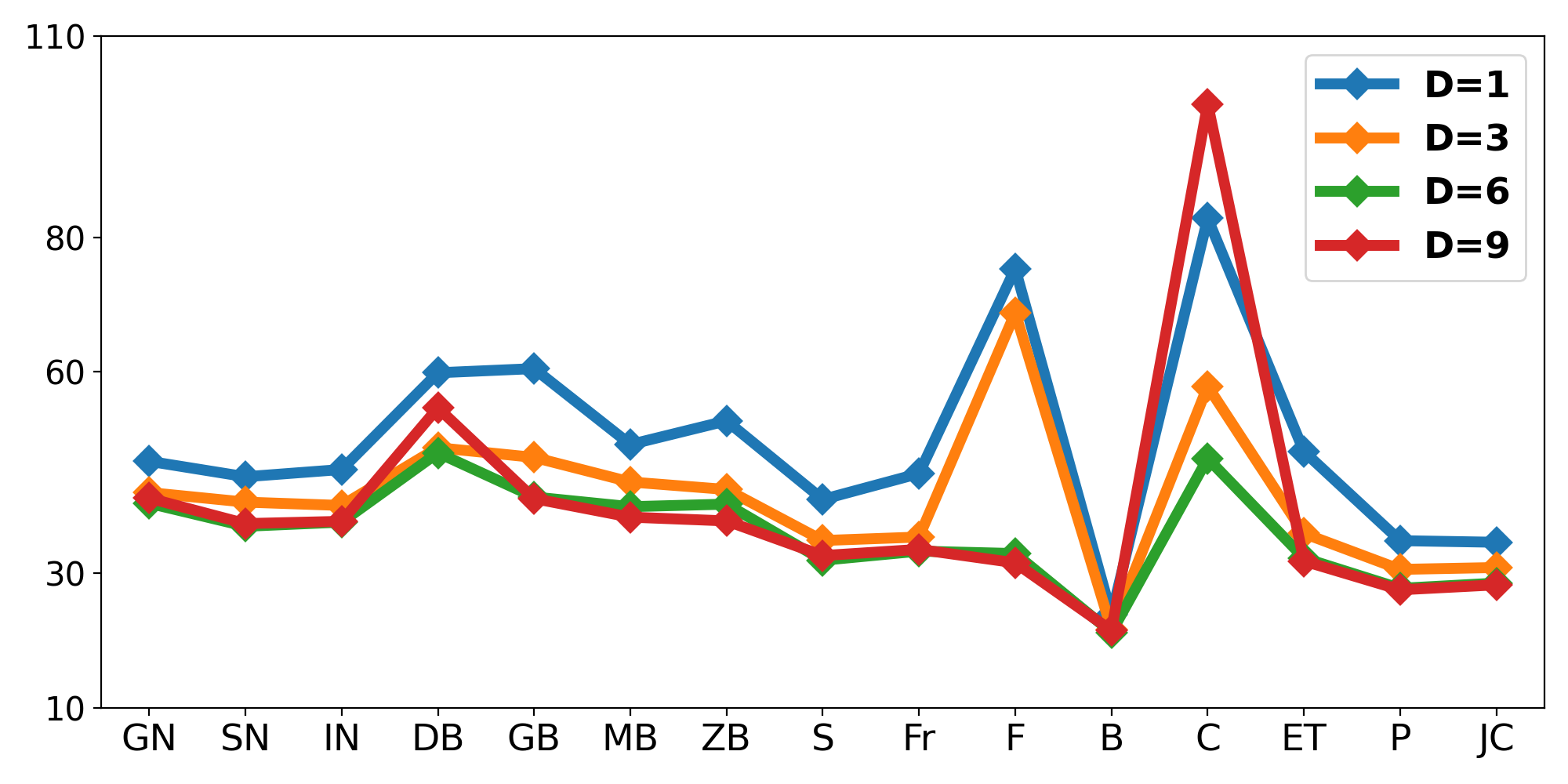}%
    }
    \subfigure[Min. act. steps $T$]{%
        \includegraphics[width=0.21\textwidth]{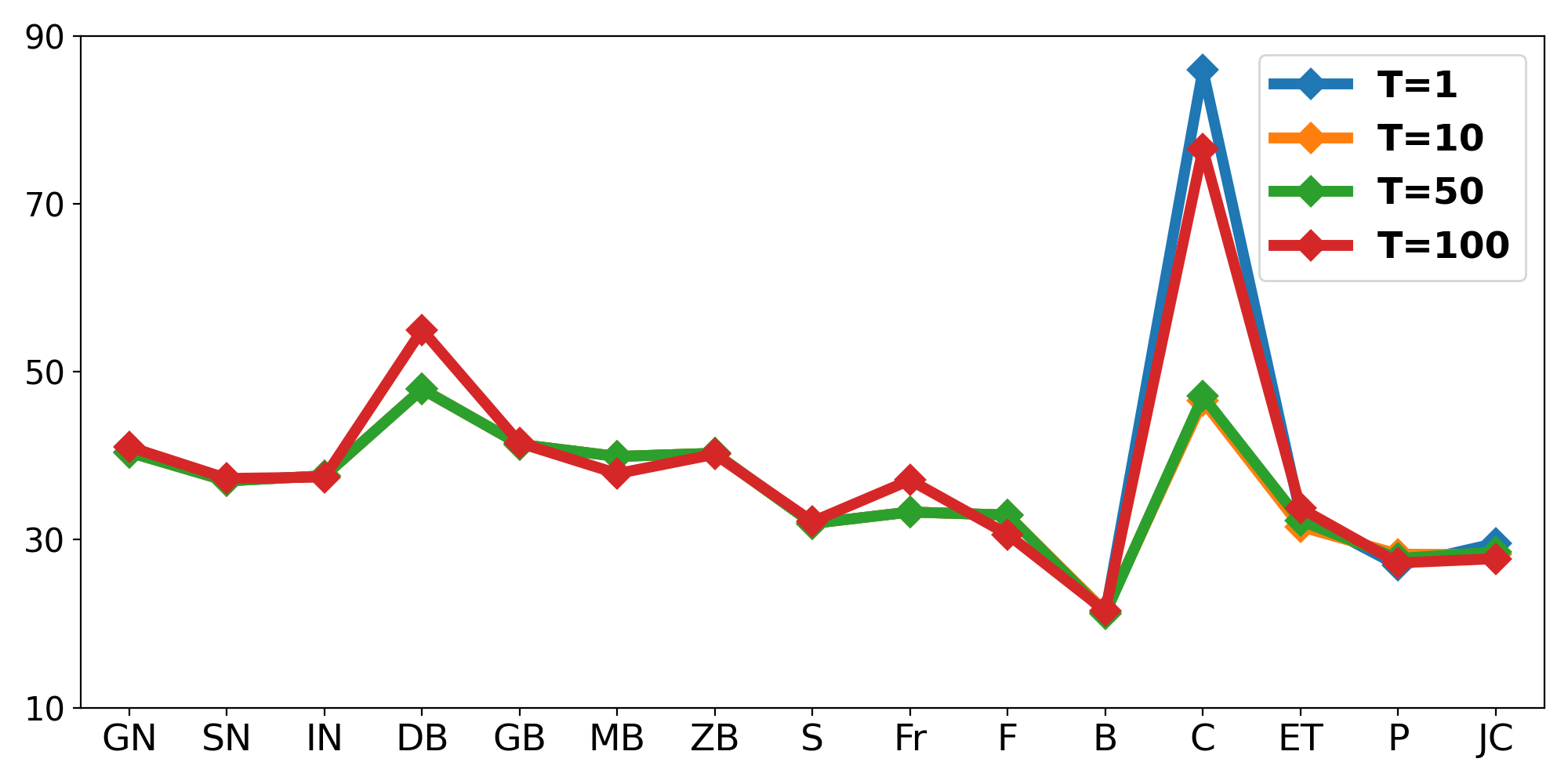}%
    }
    \subfigure[Batch size]{%
        \includegraphics[width=0.21\textwidth]{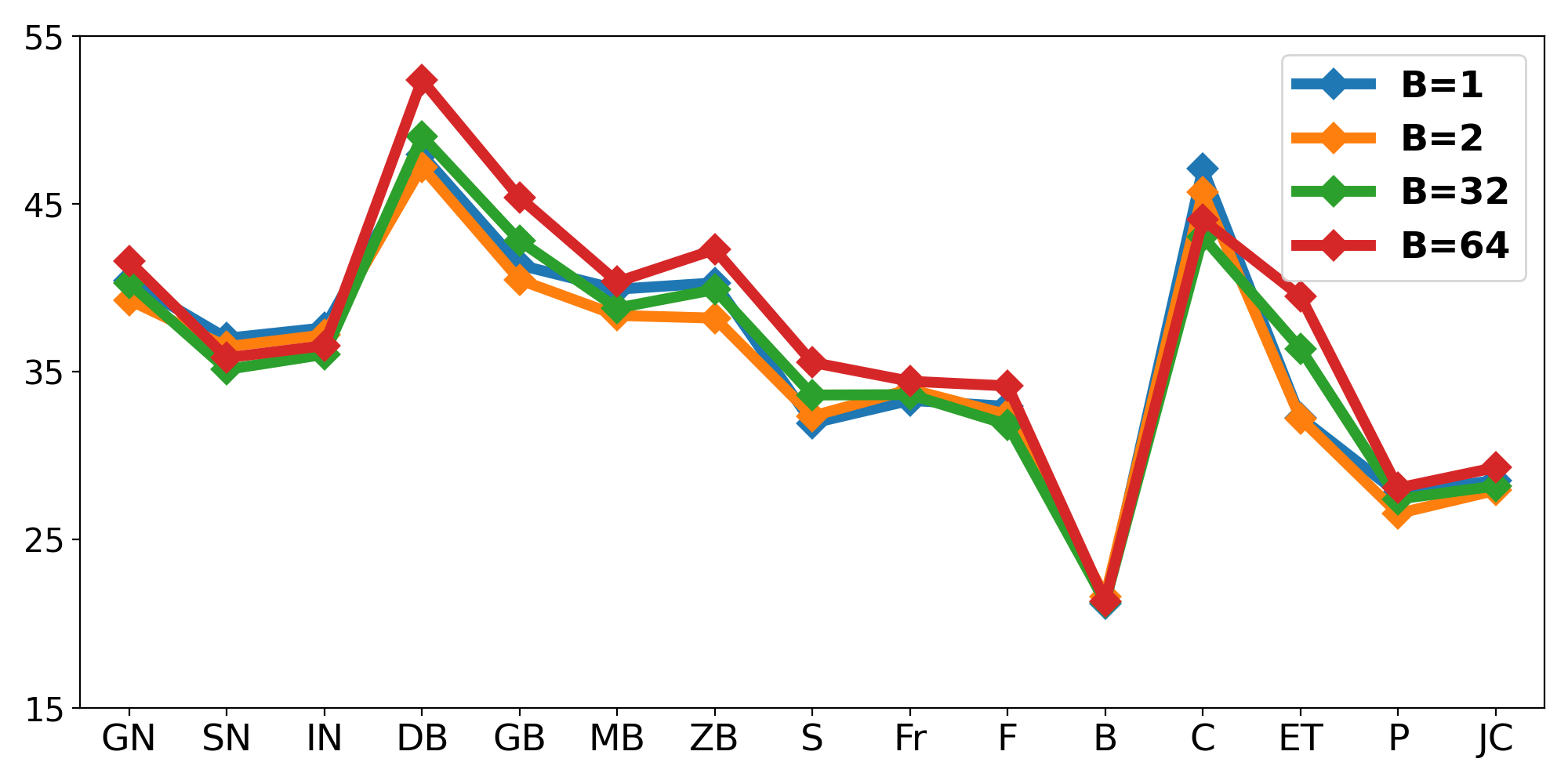}%
    }

    \caption{Hyperparameter ablations on SEGA with ImageNet-C under CTTA (batch size 1) using ViT-Base. Each panel shows the effect of a single SEGA hyperparameter (erasing ratio steps $A$, erasing levels $N$, quantile level $Q$, loss coefficient $\lambda$, warm-up steps $U$, deterioration margin $D$, minimum activation steps $T$, and batch size) on classification error across corruptions. Extended ablation results in supplementary material.}
    \label{fig:ablation}
\end{figure*}

\begin{table}[t!]
    \centering
    \caption{Ablation on SEGA with ImageNet-C and CIFAR10-C under CTTA ($B{=}1$). Entries report mean of three runs (seeds 1, 2, 3). Note: SEGA* and M2A (from Table~\ref{tab:imagenetc_cifar10c_ctta_avg}) does not have the same error because SEGA and M2A were individually tuned in learning rates and optimizer to find each methods best result at $B{=}1$.}
    \label{tab:sega_ablation}
    \footnotesize
    \setlength\tabcolsep{4pt}
    \begin{adjustbox}{width=\linewidth,center=\linewidth}
    \begin{tabular}{l|cccc|cccc}
        \toprule
        \multirow{3}{*}{Method} & \multicolumn{4}{c|}{\textbf{ImageNet-C}} & \multicolumn{4}{c}{\textbf{CIFAR10-C}} \\
        \cmidrule(lr){2-5} \cmidrule(lr){6-9}
         & Sens. Trend & Sens. Gate & Error$\downarrow$ & Gain$\uparrow$ & Sens. Trend & Sens. Gate & Error$\downarrow$ & Gain$\uparrow$ \\
         \midrule
        SOURCE         & \ding{55} & \ding{55} & 55.8     & 0.0   & \ding{55} & \ding{55} & 28.2     & 0.0\\
        SEGA*           & \ding{55} & \ding{55} & 62.2     & -6.4  & \ding{55} & \ding{55} & 17.4     & +10.8 \\
        SEGA           & \ding{55} & \ding{51} & 49.0     & +6.8  & \ding{55} & \ding{51} & 67.0     & -38.8 \\
        SEGA           & \ding{51} & \ding{55} & 37.7     & +18.1 & \ding{51} & \ding{55} & 11.1     & +17.1 \\
        SEGA           & \ding{51} & \ding{51} & 37.1     & +18.7 & \ding{51} & \ding{51} & 10.8     & +17.4 \\

        \bottomrule
    \end{tabular}
    \end{adjustbox}
\end{table}
\paragraph{Control Components.} Table~\ref{tab:sega_ablation} shows that the trend rule is the main reason SEGA remains stable. On ImageNet-C, removing both controls raises error to $62.2\%$, which is worse than the source model at $55.8\%$, while keeping only the trend rule already recovers $37.7\%$, within $0.6$~pp of full SEGA. On CIFAR10-C, the asymmetry is even sharper: removing the trend rule drives error to $67.0\%$, whereas trend-only reaches $11.1\%$ and nearly matches the full model at $10.8\%$. The gate therefore acts mainly as a selective refinement on top of the trend rule rather than as an independent stability mechanism.

\paragraph{Erasing Mechanism and Hyperparameters.} Table~\ref{tab:augmentation} shows that not all augmentations produce a useful sensitivity signal. Spatial block erasing achieves $37.1\%$ on ImageNet-C, outperforming frequency erasing at $42.3\%$ and geometric transforms such as crop ($50.5\%$), rotate ($63.8\%$), and translate ($71.3\%$). Fig.~\ref{fig:ablation} then shows that the best behavior comes from moderate settings rather than aggressive filtering: $N{=}3$ erasing levels provide a good trade-off, the $20\%$--$100\%$ band performs best, and a warm-up of about $U{=}2048$ is needed before gating becomes reliable. Together these results support the interpretation that SEGA is most effective on the studied corruption-style streams and backbones when erasing creates a smooth information-reduction channel that separates more stable from less stable samples.
\section{Conclusion}
\label{sec:conclusion}
In this work we presented strict online continual test-time adaptation, where the key decisions are when to recover and which samples should update the model. Our results show that per-sample sensitivity under multi-view erasing provides a practical feedback signal for single-backbone strict online CTTA on corruption-style streams, yielding more stable adaptation, lower long-horizon error, and better efficiency than entropy- and reset-based schemes across standard corruption benchmarks and corruption-generated aquaculture streams. The sensitivity gate further reduces unnecessary backward updates, and the trend rule prevents long-horizon collapse. The strongest evidence supports corruption-style streaming shifts and architectures for which erasing produces a useful sensitivity signal, especially transformer-style backbones. A limitation is that reset-to-source discards cross-domain adaptations, and exploring partial rollback or weight interpolation could extend SEGA to settings where knowledge accumulation across domains is essential.

{
    \small
    \bibliographystyle{ieeenat_fullname}
    \bibliography{main}
}

\end{document}


\maketitle

\tableofcontents

\section{Discussion}
\label{sec:appendix-discussion}
Across the main benchmarks, a consistent pattern is that SEGA's sensitivity-trend rule and erasing-based probe yield stable overall stream-mean improvements in strict online $B{=}1$ corruption streams, even though entropy- and reset-based baselines such as SAR and RDumb remain competitive on some corruption groups. The ImageNet-C and CIFAR10-C results, together with the extended aquaculture experiments, show that a single set of control hyperparameters tuned on ImageNet-C can be transferred unchanged to CIFAR10/100-C and corruption-generated aquaculture streams while still yielding stable long-horizon gains. This suggests that the sensitivity-driven control is not narrowly overfit to one dataset, even though it remains specialized to single-backbone, corruption-style shifts.

At the same time, the backbone and corruption-regime dependence visible in Table~\ref{tab:backbone}, the shuffled and lifelong CTTA tables, and the domain-sensitivity timelines indicates that SEGA's advantages hinge on regimes where block-wise erasing creates a clear separation between low- and high-instability behaviour. Transformer-style backbones with large receptive fields, such as ViT and ConvNeXt-B, appear particularly compatible with this probe, whereas CNNs can leak information around erased regions and reduce the usefulness of the sensitivity signal. The aquaculture streams and shuffled/lifelong schedules further highlight that SEGA behaves best when erasing exposes gradual drift rather than abrupt domain switches or extremely weak sensitivity contrast.

Finally, the extended ablations and efficiency analyses underscore both the promise and the limits of SEGA's information-theoretically motivated heuristics. The deterioration margin, quantile gate, warm-up horizon, and EMA weight together carve out a practical stability region in which elevated sensitivity tends to precede collapse on our benchmarks, yet the method still lacks a formal stability guarantee and the gate's marginal gains are smaller than those of the trend rule in some settings. A more complete theory would need to explain when erasing-induced sensitivity should track true risk across architectures and data regimes, and how to design control policies and erasing channels that remain robust beyond the corruption-style shifts and transformer backbones studied here.

\section{Additional Results}
\label{sec:additional-results}

\subsection{Backbone Analysis}
\label{sec:appendix-backbone}
\begin{table*}[t!]
    \centering
    \caption{Classification error rate (\%) on ImageNet-C for ResNet50-GN and ConvNext-B; GN-SN, DB-ZB, S-B, and C-JC are grouped by noise, blur, weather, and digital corruptions.}
    \label{tab:backbone}
    \tiny
    \setlength\tabcolsep{8pt}
    \begin{adjustbox}{width=\textwidth,center=\textwidth}
    \begin{tabular}{l|cccc|cc|cccc|cc}
        \toprule
        \multirow{3}{*}{Method} & \multicolumn{6}{c|}{\textbf{ResNet50-GN}} & \multicolumn{6}{c}{\textbf{ConvNext-B}} \\
        \cmidrule(lr){2-7} \cmidrule(lr){8-13}
         & GN-SN & DB-ZB & S-B & C-JC & Mean$\downarrow$ & Gain$\uparrow$ & GN-SN & DB-ZB & S-B & C-JC & Mean$\downarrow$ & Gain$\uparrow$ \\
         \midrule
        SOURCE          & 74.0 & 85.1 & 57.8 & 73.9 & 72.7     & 0.0   & 47.0 & 61.7 & 36.4 & 47.9 & 48.3 & 0.0 \\
        SAR             & \bf 65.0 & 85.7 & 48.7 & \bf 63.5 & \bf 65.7     & +7.0  & 36.5 & 61.1 & 28.9 & 52.9 & 44.9 & +3.4 \\
        RDUMB           & 76.0 & \bf 83.5 & 63.4 & 70.4 & 73.3     & -0.6  & 42.6 & 53.4 & 29.2 & 37.6 & 40.7 & +7.6 \\
        M2A             & 66.9 & 99.1 & 91.7 & 99.8 & 89.4     & -16.7 & \bf 36.4 & 93.7 & 99.7 & 99.8 & 82.4 & -34.1 \\
        SEGA            & 65.8 & 91.7 & \bf 46.5 & 67.5 & 67.8     & +4.9  & 37.1 & \bf 46.1 & \bf 27.8 & \bf 31.7 & \bf 35.7 & +12.6 \\

        \bottomrule
    \end{tabular}
    \end{adjustbox}
\end{table*}
Table~\ref{tab:backbone} shows that SEGA's block-wise erasing probe is naturally compatible with transformer-based backbones such as ViT and ConvNext-B, where it improves over the source model and prior CTTA baselines, whereas on ResNet50-GN gains are more mixed because spatial erasing can be less effective for CNNs due to information leakage~\citep{Li2022ArchitectureAgnosticMI} and SAR attains the lowest mean error. These results suggest that SEGA is best aligned with transformer-based backbones, while convolutional models may require modest retuning. Although SEGA exposes several control hyperparameters, in our experiments we tune $(A,N)$, $(Q_{\min},Q_{\max},U)$, and $(T,D)$ only on ImageNet-C and then transfer this single setting unchanged to CIFAR10/100-C and the aquaculture corruption streams without retuning. Our use of sustained high sensitivity as a proxy for collapse risk, and of a quantile-based information-stability band, should therefore be viewed as information-theoretically motivated heuristics (supported by the sensitivity timeline in Fig.~\ref{fig:method}(a) and our ablations) rather than formally proven properties or closed-form prescriptions for exact thresholds. Characterizing the optimal entropy-based band more precisely, and its dependence on architecture and stream type, remains an open direction.

\subsection{CTTA Benchmark Details}
\label{sec:appendix-ctta}
\begin{table*}[h!]
    \centering
    \small
    \caption{Fine-grained per-corruption ImageNet-C error rates (\%) for CTTA (batch size 1) using ViT-Base. Mean is the average across 15 corruption domains, each with 5000 test samples. Gain is the relative improvement over the source model. Entries report mean and standard deviation of three runs (seeds 1, 2, 3). }
    \label{tab:imagenetc_tta_continual_std}
    \setlength\tabcolsep{6pt}
    \resizebox{\linewidth}{!}{
    \begin{tabular}{l|ccc|cccc|cccc|cccc|cc}
        \toprule
        Time & \multicolumn{15}{c|}{$t\xrightarrow{\;}$}& \\ \hline
        Method &
        \rotatebox[origin=c]{0}{GN} & \rotatebox[origin=c]{0}{SN} & \rotatebox[origin=c]{0}{IN} & \rotatebox[origin=c]{0}{DB} & \rotatebox[origin=c]{0}{GB} & \rotatebox[origin=c]{0}{MB} & \rotatebox[origin=c]{0}{ZB} & \rotatebox[origin=c]{0}{S} & \rotatebox[origin=c]{0}{Fr} & \rotatebox[origin=c]{0}{F}  & \rotatebox[origin=c]{0}{B} & \rotatebox[origin=c]{0}{C} & \rotatebox[origin=c]{0}{ET} & \rotatebox[origin=c]{0}{P} & \rotatebox[origin=c]{0}{JC}
        & Mean$\downarrow$ & Gain$\uparrow$\\\hline
        SOURCE (2021)~\citep{dosovitskiy2021vit}   & 53.0$\pm0.0$ & 51.8$\pm0.0$ & 52.1$\pm0.0$ & 68.5$\pm0.0$ & 78.8$\pm0.0$ & 58.5$\pm0.0$ & 63.3$\pm0.0$ & 49.9$\pm0.0$ & 54.2$\pm0.0$ & 57.7$\pm0.0$ & 26.4$\pm0.0$ & 91.4$\pm0.0$ & 57.5$\pm0.0$ & 38.0$\pm0.0$ & 36.2$\pm0.0$ & 55.8 & 0.0\\
        ROTTA (2023)~\citep{Yuan2023RobustTA}      & 49.9$\pm0.0$ & 48.1$\pm0.1$ & 48.3$\pm0.2$ & 67.9$\pm0.0$ & 70.6$\pm0.3$ & 55.2$\pm0.0$ & 59.5$\pm0.5$ & 45.4$\pm0.1$ & 48.5$\pm0.6$ & 54.1$\pm0.7$ & 24.7$\pm0.1$ & 88.3$\pm0.1$ & 53.9$\pm0.2$ & 37.3$\pm0.1$ & 34.9$\pm0.2$ & 52.4 & +3.4\\
        SANTA (2023)~\citep{Chakrabarty2023SATASA} & 46.0$\pm0.0$ & 43.2$\pm0.0$ & 44.6$\pm0.0$ & 58.7$\pm0.0$ & 64.4$\pm0.0$ & 47.7$\pm0.0$ & 54.1$\pm0.0$ & 40.0$\pm0.0$ & 46.6$\pm0.0$ & 41.9$\pm0.0$ & 23.3$\pm0.0$ & 89.5$\pm0.0$ & 49.1$\pm0.0$ & 34.7$\pm0.0$ & 33.5$\pm0.0$ & 47.8 & +8.0\\
        TENT (2021)~\citep{DequanWangetal2021}     & 47.4$\pm0.2$ & 41.8$\pm0.1$ & 41.7$\pm0.0$ & 57.5$\pm0.4$ & 58.4$\pm0.2$ & 45.8$\pm0.2$ & 51.5$\pm0.4$ & 40.4$\pm0.1$ & 41.9$\pm0.0$ & 41.8$\pm0.2$ & 22.5$\pm0.1$ & 54.0$\pm0.1$ & 50.2$\pm0.3$ & 31.9$\pm0.2$ & 30.9$\pm0.0$ & 43.8 & +12.0\\
        RPL (2021)~\citep{Rusak2021IfYD}           & 46.6$\pm0.2$ & 40.6$\pm0.1$ & 40.6$\pm0.5$ & 55.7$\pm0.7$ & 55.9$\pm0.5$ & 44.6$\pm0.3$ & 51.1$\pm0.4$ & 41.0$\pm0.5$ & 42.5$\pm0.2$ & 40.9$\pm0.3$ & 23.5$\pm0.2$ & 69.0$\pm16.9$ & 64.6$\pm24.8$ & 54.9$\pm31.8$ & 54.3$\pm32.2$ & 48.4 & +7.4\\
        LCOTTA (2025)~\citep{Duan2025}             & 41.5$\pm0.3$ & 37.6$\pm0.1$ & 39.3$\pm0.5$ & 54.9$\pm1.4$ & 54.4$\pm0.4$ & 46.4$\pm0.1$ & 51.8$\pm0.5$ & 41.1$\pm0.7$ & 40.5$\pm0.1$ & 42.2$\pm0.3$ & 24.2$\pm0.2$ & 58.7$\pm0.5$ & 47.7$\pm2.4$ & 33.1$\pm0.6$ & 32.1$\pm0.1$ & 43.0 & +12.8\\
        RESERVOIRTTA (2025)~\citep{Vray2025ResTTA} & 41.9$\pm0.2$ & 39.5$\pm0.4$ & 39.5$\pm0.2$ & 50.0$\pm1.2$ & 47.5$\pm0.8$ & 45.0$\pm0.5$ & 49.8$\pm3.1$ & 39.0$\pm5.2$ & 39.4$\pm2.4$ & 37.3$\pm0.3$ & 21.9$\pm0.6$ & 88.7$\pm4.0$ & 38.5$\pm4.7$ & 29.3$\pm0.0$ & 29.9$\pm0.1$ & 42.5 & +13.3\\
        SAR (2023)~\citep{niu2023sar}              & 41.4$\pm0.5$ & 38.3$\pm0.4$ & 39.4$\pm0.5$ & 54.8$\pm1.8$ & 45.3$\pm1.0$ & 42.9$\pm2.3$ & 43.2$\pm1.2$ & 39.3$\pm1.8$ & 41.5$\pm8.6$ & 82.0$\pm16.1$ & 28.7$\pm6.5$ & 99.6$\pm0.2$ & 37.9$\pm6.7$ & 27.7$\pm0.5$ & 28.8$\pm0.4$ & 46.1 & +9.7\\
        RDUMB (2023)~\citep{Press2023RDumbAS}      & 44.0$\pm0.9$ & 42.4$\pm0.1$ & 42.4$\pm0.4$ & 49.5$\pm0.2$ & 48.2$\pm0.8$ & 43.6$\pm0.7$ & 46.2$\pm0.5$ & 35.1$\pm0.1$ & 37.9$\pm0.5$ & 37.0$\pm0.4$ & 21.6$\pm0.5$ & 69.4$\pm2.5$ & 37.4$\pm0.3$ & 27.9$\pm0.8$ & 29.5$\pm0.3$ & 40.8 & +15.0\\
        M2A (2026)~\citep{Doloriel2026Family}      & 41.2$\pm0.3$ & 35.8$\pm0.1$ & 36.2$\pm0.4$ & 49.6$\pm0.8$ & 44.9$\pm1.1$ & 38.7$\pm0.2$ & 41.1$\pm0.8$ & 34.1$\pm0.2$ & 33.7$\pm0.2$ & 32.7$\pm0.3$ & 21.2$\pm0.1$ & 64.6$\pm27.6$ & 57.4$\pm36.7$ & 51.6$\pm41.8$ & 52.5$\pm40.9$ & 42.4 & +13.4\\
        SEGA                                       & 40.5$\pm0.2$ & 37.3$\pm0.4$ & 38.3$\pm1.0$ & 47.7$\pm1.2$ & 42.1$\pm0.5$ & 38.9$\pm0.8$ & 39.4$\pm1.2$ & 31.9$\pm1.0$ & 39.7$\pm8.0$ & 36.3$\pm5.4$ & 21.1$\pm0.1$ & 55.6$\pm12.6$ & 31.6$\pm0.7$ & 27.7$\pm0.4$ & 28.4$\pm0.1$ & 37.1 & +18.7\\

        \bottomrule
    \end{tabular}%
    }

\end{table*}

\begin{table*}[h!]
    \centering
    \small
    \caption{Fine-grained per-corruption CIFAR10-C error rates (\%) for CTTA (batch size 1) using ViT-Base. Mean is the average across 15 corruption domains, each with 10000 test samples. Gain is the relative improvement over the source model. Entries report mean and standard deviation of three runs (seeds 1, 2, 3). }
    \label{tab:cifar10c_tta_continual_std}
    \setlength\tabcolsep{6pt}
    \resizebox{\linewidth}{!}{
    \begin{tabular}{l|ccc|cccc|cccc|cccc|cc}
        \toprule
        Time & \multicolumn{15}{c|}{$t\xrightarrow{\;}$}& \\ \hline
        Method &
        \rotatebox[origin=c]{0}{GN} & \rotatebox[origin=c]{0}{SN} & \rotatebox[origin=c]{0}{IN} & \rotatebox[origin=c]{0}{DB} & \rotatebox[origin=c]{0}{GB} & \rotatebox[origin=c]{0}{MB} & \rotatebox[origin=c]{0}{ZB} & \rotatebox[origin=c]{0}{S} & \rotatebox[origin=c]{0}{Fr} & \rotatebox[origin=c]{0}{F}  & \rotatebox[origin=c]{0}{B} & \rotatebox[origin=c]{0}{C} & \rotatebox[origin=c]{0}{ET} & \rotatebox[origin=c]{0}{P} & \rotatebox[origin=c]{0}{JC}
        & Mean$\downarrow$ & Gain$\uparrow$\\\hline
        SOURCE (2021)~\citep{dosovitskiy2021vit}   & 60.1$\pm$0.0 & 53.2$\pm$0.0 & 38.3$\pm$0.0 & 19.9$\pm$0.0 & 35.5$\pm$0.0 & 22.6$\pm$0.0 & 18.6$\pm$0.0 & 12.1$\pm$0.0 & 12.7$\pm$0.0 & 22.8$\pm$0.0 & 5.3$\pm$0.0 & 49.7$\pm$0.0 & 23.6$\pm$0.0 & 24.7$\pm$0.0 & 23.1$\pm$0.0 & 28.2 & 0.0\\
        ROTTA (2023)~\citep{Yuan2023RobustTA}      & 60.9$\pm$0.1 & 56.2$\pm$0.2 & 38.4$\pm$0.4 & 18.4$\pm$0.2 & 34.2$\pm$0.4 & 17.5$\pm$0.1 & 13.1$\pm$0.1 & 10.8$\pm$0.4 & 12.1$\pm$0.7 & 14.0$\pm$0.4 & 4.2$\pm$0.0 & 21.1$\pm$1.4 & 14.8$\pm$0.3 & 27.5$\pm$0.9 & 19.8$\pm$0.2 & 24.2 & +4.0\\
        TENT (2021)~\citep{DequanWangetal2021}     & 59.4$\pm$0.9 & 59.0$\pm$1.1 & 30.6$\pm$0.5 & 16.7$\pm$0.2 & 35.2$\pm$0.5 & 16.8$\pm$0.2 & 12.7$\pm$0.2 & 10.7$\pm$0.1 & 12.1$\pm$0.1 & 15.6$\pm$0.2 & 4.8$\pm$0.1 & 26.7$\pm$1.0 & 16.9$\pm$0.1 & 28.6$\pm$0.7 & 20.6$\pm$0.1 & 24.4 & +3.8\\
        RPL (2021)~\citep{Rusak2021IfYD}           & 52.3$\pm$1.6 & 38.7$\pm$3.0 & 27.1$\pm$2.4 & 18.6$\pm$1.0 & 33.4$\pm$0.9 & 19.3$\pm$2.4 & 13.7$\pm$0.6 & 10.8$\pm$1.0 & 10.4$\pm$0.5 & 16.9$\pm$0.7 & 4.6$\pm$0.0 & 27.4$\pm$6.7 & 18.1$\pm$0.9 & 29.7$\pm$1.8 & 21.3$\pm$0.6 & 22.8 & +5.4\\
        LCOTTA (2025)~\citep{Duan2025}             & 57.0$\pm$0.1 & 50.2$\pm$0.0 & 37.4$\pm$0.3 & 20.1$\pm$0.0 & 35.1$\pm$0.1 & 22.8$\pm$0.1 & 18.8$\pm$0.1 & 11.8$\pm$0.1 & 12.5$\pm$0.0 & 22.9$\pm$0.0 & 5.3$\pm$0.0 & 50.5$\pm$0.2 & 23.5$\pm$0.1 & 25.0$\pm$0.2 & 22.9$\pm$0.0 & 27.7 & +0.5\\
        RESERVOIRTTA (2025)~\citep{Vray2025ResTTA} & 63.2$\pm$0.7 & 59.7$\pm$0.4 & 38.5$\pm$0.1 & 13.5$\pm$0.0 & 42.7$\pm$0.1 & 16.5$\pm$0.1 & 14.0$\pm$0.0 & 13.5$\pm$0.0 & 15.2$\pm$0.0 & 13.3$\pm$0.0 & 4.5$\pm$0.0 & 15.4$\pm$0.1 & 23.5$\pm$0.0 & 27.3$\pm$0.0 & 28.4$\pm$0.0 & 25.9 & +2.3\\
        SAR (2023)~\citep{niu2023sar}              & 55.0$\pm$0.4 & 49.4$\pm$0.1 & 38.0$\pm$0.2 & 19.5$\pm$0.0 & 35.1$\pm$0.1 & 22.3$\pm$0.1 & 18.3$\pm$0.0 & 12.0$\pm$0.0 & 12.7$\pm$0.0 & 22.4$\pm$0.1 & 5.3$\pm$0.0 & 41.7$\pm$0.4 & 23.1$\pm$0.0 & 24.2$\pm$0.0 & 23.0$\pm$0.1 & 26.9 & +1.3\\
        RDUMB (2023)~\citep{Press2023RDumbAS}      & 27.7$\pm$1.1 & 24.0$\pm$0.6 & 15.1$\pm$0.6 & 11.5$\pm$0.4 & 28.0$\pm$1.0 & 13.1$\pm$0.8 & 10.8$\pm$0.5 & 9.3$\pm$0.3 & 9.2$\pm$0.1 & 13.0$\pm$0.4 & 5.0$\pm$0.5 & 16.8$\pm$0.9 & 17.1$\pm$1.3 & 12.7$\pm$0.3 & 19.4$\pm$1.0 & 15.5 & +12.7\\
        M2A (2026)~\citep{Doloriel2026Family}      & 23.4$\pm$0.7 & 13.9$\pm$0.4 & 11.7$\pm$0.1 & 9.6$\pm$0.0 & 22.9$\pm$0.3 & 9.7$\pm$0.0 & 6.3$\pm$0.1 & 7.4$\pm$0.1 & 5.9$\pm$0.0 & 8.1$\pm$0.1 & 4.3$\pm$0.1 & 6.5$\pm$0.0 & 12.4$\pm$0.1 & 10.0$\pm$0.1 & 15.6$\pm$0.4 & 11.1 & +17.1\\
        SEGA                                       & 15.8$\pm$0.5 & 14.0$\pm$0.0 & 9.8$\pm$0.5 & 9.3$\pm$0.0 & 22.8$\pm$0.3 & 9.4$\pm$0.5 & 7.7$\pm$0.6 & 7.5$\pm$0.0 & 7.5$\pm$0.2 & 8.8$\pm$0.1 & 4.4$\pm$0.2 & 8.4$\pm$0.3 & 13.2$\pm$0.9 & 9.9$\pm$0.2 & 13.8$\pm$1.1 & 10.8 & +17.4\\

        \bottomrule
    \end{tabular}%
    }

\end{table*}

\begin{table*}[h!]
    \centering
    \small
    \caption{Fine-grained per-corruption CIFAR100-C and FreshFish-C error rates (\%) for CTTA (batch size 1) using ViT-Base. Mean is the average across 15 corruption domains, each with 10000 test samples. Gain is the relative improvement over the source model. Entries report mean and standard deviation of three runs (seeds 1, 2, 3). }
    \label{tab:benchmark_ctta_std}
    \setlength\tabcolsep{6pt}
    \resizebox{\linewidth}{!}{
    \begin{tabular}{l|ccc|cccc|cccc|cccc|c}
        \toprule
        Time & \multicolumn{15}{c|}{$t\xrightarrow{\;}$}& \\ \hline
        Method &
        \rotatebox[origin=c]{0}{GN} & \rotatebox[origin=c]{0}{SN} & \rotatebox[origin=c]{0}{IN} & \rotatebox[origin=c]{0}{DB} & \rotatebox[origin=c]{0}{GB} & \rotatebox[origin=c]{0}{MB} & \rotatebox[origin=c]{0}{ZB} & \rotatebox[origin=c]{0}{S} & \rotatebox[origin=c]{0}{Fr} & \rotatebox[origin=c]{0}{F}  & \rotatebox[origin=c]{0}{B} & \rotatebox[origin=c]{0}{C} & \rotatebox[origin=c]{0}{ET} & \rotatebox[origin=c]{0}{P} & \rotatebox[origin=c]{0}{JC}
        & Mean$\downarrow$ \\\hline
        \multicolumn{17}{c}{\textbf{CIFAR100-C}}\\\hline
        SOURCE (2021)~\citep{dosovitskiy2021vit}   & 55.0$\pm$0.0 & 51.5$\pm$0.0 & 26.9$\pm$0.0 & 24.0$\pm$0.0 & 60.5$\pm$0.0 & 29.0$\pm$0.0 & 21.4$\pm$0.0 & 21.1$\pm$0.0 & 25.0$\pm$0.0 & 35.1$\pm$0.0 & 11.8$\pm$0.0 & 34.8$\pm$0.0 & 43.1$\pm$0.0 & 56.0$\pm$0.0 & 35.9$\pm$0.0 & 35.4 \\
        SAR (2023)~\citep{niu2023sar}              & 97.6$\pm$0.7 & 98.3$\pm$0.5 & 93.7$\pm$8.5 & 88.5$\pm$12.9 & 97.8$\pm$1.3 & 98.4$\pm$0.4 & 98.3$\pm$0.5 & 75.5$\pm$40.1 & 85.9$\pm$22.3 & 98.9$\pm$0.3 & 98.5$\pm$0.2 & 98.8$\pm$0.4 & 98.6$\pm$0.1 & 98.6$\pm$0.1 & 98.7$\pm$0.2 & 95.1 \\
        RDUMB (2023)~\citep{Press2023RDumbAS}      & 42.1$\pm$1.8 & 38.3$\pm$0.6 & 19.8$\pm$0.5 & 21.9$\pm$0.1 & 43.2$\pm$0.9 & 24.2$\pm$1.0 & 20.2$\pm$1.7 & 19.8$\pm$0.8 & 20.3$\pm$0.3 & 26.9$\pm$1.4 & 12.1$\pm$0.1 & 19.2$\pm$0.9 & 33.2$\pm$0.7 & 25.1$\pm$1.4 & 32.1$\pm$1.8 & 26.6 \\
        M2A (2026)~\citep{Doloriel2026Family}      & 51.8$\pm$29.7 & 49.8$\pm$34.8 & 44.0$\pm$38.9 & 45.9$\pm$37.5 & 55.3$\pm$30.9 & 46.3$\pm$37.3 & 43.4$\pm$39.4 & 44.4$\pm$38.6 & 43.6$\pm$39.1 & 45.6$\pm$37.8 & 40.7$\pm$41.2 & 43.1$\pm$39.5 & 51.3$\pm$33.8 & 46.0$\pm$37.5 & 51.1$\pm$33.9 & 46.8 \\
        SEGA                                       & 32.6$\pm$0.8 & 29.4$\pm$1.3 & 17.7$\pm$0.5 & 23.2$\pm$1.6 & 38.7$\pm$2.1 & 23.5$\pm$0.3 & 18.4$\pm$0.6 & 18.4$\pm$0.4 & 18.4$\pm$0.1 & 21.9$\pm$0.2 & 11.6$\pm$0.1 & 16.6$\pm$0.1 & 29.9$\pm$1.5 & 20.9$\pm$0.2 & 30.9$\pm$0.3 & 23.5 \\
        \midrule

        \multicolumn{17}{c}{\textbf{FreshFish-C}}\\\hline
        SOURCE (2021)~\citep{dosovitskiy2021vit}    & 24.0$\pm$0.0 & 20.7$\pm$0.0 & 21.3$\pm$0.0 & 37.6$\pm$0.0 & 27.1$\pm$0.0 & 30.1$\pm$0.0 & 20.6$\pm$0.0 & 36.3$\pm$0.0 & 35.4$\pm$0.0 & 65.3$\pm$0.0 & 12.1$\pm$0.0 & 85.6$\pm$0.0 & 10.6$\pm$0.0 & 3.7$\pm$0.0 & 3.1$\pm$0.0 & 28.9 \\
        SAR (2023)~\citep{niu2023sar}              & 22.1$\pm$0.5 & 19.3$\pm$0.1 & 18.9$\pm$0.3 & 35.8$\pm$0.9 & 26.7$\pm$0.1 & 28.6$\pm$0.0 & 19.9$\pm$0.2 & 33.4$\pm$0.4 & 33.8$\pm$0.3 & 65.2$\pm$0.4 & 11.9$\pm$0.1 & 85.6$\pm$0.0 & 10.4$\pm$0.1 & 3.6$\pm$0.1 & 3.1$\pm$0.0 & 27.9 \\
        RDUMB (2023)~\citep{Press2023RDumbAS}      & 20.2$\pm$2.6 & 16.9$\pm$2.1 & 17.4$\pm$1.0 & 38.8$\pm$1.2 & 26.0$\pm$0.4 & 26.8$\pm$0.1 & 19.2$\pm$1.4 & 34.7$\pm$1.0 & 35.0$\pm$0.2 & 63.3$\pm$1.8 & 11.2$\pm$0.3 & 85.6$\pm$0.0 & 8.9$\pm$0.3 & 4.0$\pm$0.5 & 3.2$\pm$0.3 & 27.4 \\
        M2A (2026)~\citep{Doloriel2026Family}      & 16.2$\pm$0.8 & 11.5$\pm$0.1 & 9.6$\pm$0.4 & 28.2$\pm$0.7 & 20.0$\pm$0.2 & 12.1$\pm$0.8 & 10.4$\pm$0.7 & 22.0$\pm$0.5 & 21.3$\pm$0.6 & 37.7$\pm$0.5 & 8.6$\pm$0.6 & 84.1$\pm$0.1 & 15.8$\pm$0.2 & 2.3$\pm$0.5 & 2.7$\pm$0.1 & 20.2 \\
        SEGA                                       & 10.2$\pm$0.5 & 5.8$\pm$0.4 & 3.0$\pm$0.4 & 15.3$\pm$5.1 & 8.0$\pm$0.2 & 8.3$\pm$1.1 & 6.7$\pm$0.9 & 20.2$\pm$0.6 & 9.6$\pm$0.7 & 35.0$\pm$17.8 & 6.4$\pm$0.8 & 85.4$\pm$0.2 & 6.1$\pm$1.2 & 2.3$\pm$0.3 & 1.5$\pm$0.4 & 14.9 \\

        \bottomrule
    \end{tabular}%
    }

\end{table*}

Tables~\ref{tab:imagenetc_tta_continual_std}, \ref{tab:cifar10c_tta_continual_std}, and \ref{tab:benchmark_ctta_std} report fine-grained corruption-wise results for our continual CTTA benchmarks, listing per-corruption ImageNet-C and CIFAR10-C error rates and gains for ViT-Base under strict CTTA and aggregating mean performance across corruption domains and datasets. These decompositions clarify how SEGA and prior CTTA methods trade off stability and accuracy across specific corruption types beyond the headline mean metrics.

\subsection{Erasing Mechanism Analysis}
\label{sec:appendix-erasing}

\begin{figure*}[h!]
    \centering
    \small
    \includegraphics[width=0.9\textwidth]{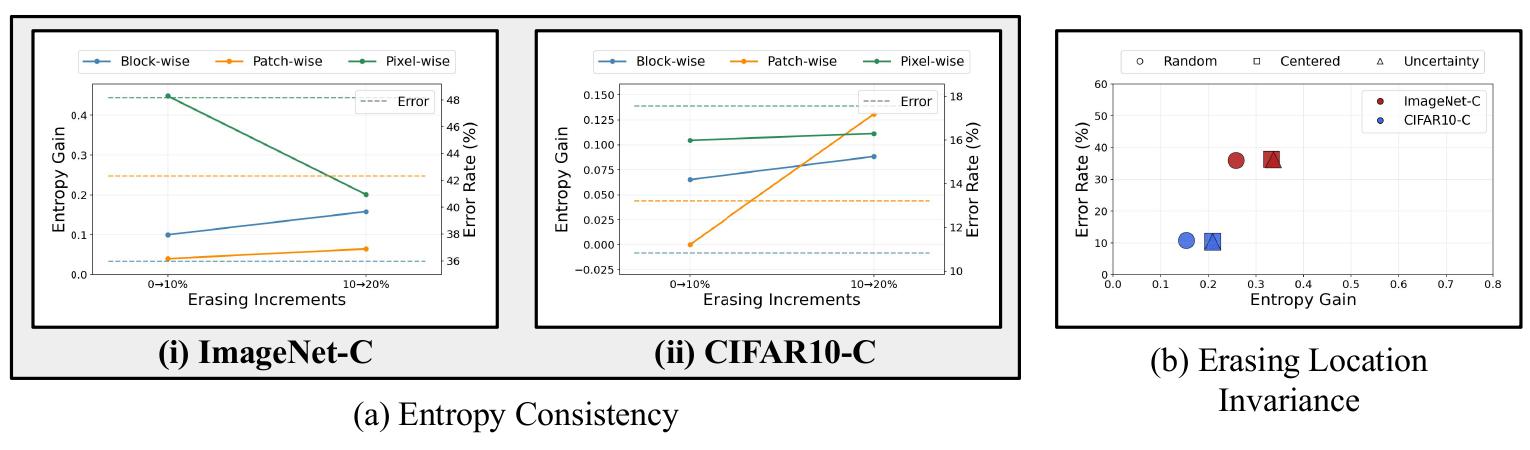}
    \caption{Analysis of SEGA's erasing mechanism. (a) \textbf{Entropy Consistency:} on ImageNet-C and CIFAR10-C, block-wise erasing yields the most consistent and well-behaved entropy response across erasing increments and attains the lowest error, with moderate, monotonic entropy gains per erasing step, whereas pixel and patch erasing produces inconsistent entropy jumps and higher error. (b) \textbf{Erasing Location Invariance:} for block erasing, error rates for random, centered, and uncertainty-guided block locations cluster within each dataset, indicating that the entropy-based control signal is largely insensitive to the exact block location.}
    \label{fig:erasing}
\end{figure*}

\begin{table*}[h!]
    \centering
    \small
    \caption{SEGA ablation on ImageNet-C and CIFAR10-C CTTA (batch size 1) comparing erasing families with ViT-Base; entries are error rates (\%). Mean is over 15 corruption domains. Gain is relative improvement over the source model. Entries report mean$\pm$std over three runs (seeds 1, 2, 3). }
    \label{tab:erasing_family}
    \setlength\tabcolsep{6pt}
    \resizebox{\linewidth}{!}{
    \begin{tabular}{l|ccc|cccc|cccc|cccc|c}
        \toprule
        Time & \multicolumn{15}{c|}{$t\xrightarrow{\;}$}& \\ \hline
        Family &
        \rotatebox[origin=c]{0}{GN} & \rotatebox[origin=c]{0}{SN} & \rotatebox[origin=c]{0}{IN} & \rotatebox[origin=c]{0}{DB} & \rotatebox[origin=c]{0}{GB} & \rotatebox[origin=c]{0}{MB} & \rotatebox[origin=c]{0}{ZB} & \rotatebox[origin=c]{0}{S} & \rotatebox[origin=c]{0}{Fr} & \rotatebox[origin=c]{0}{F}  & \rotatebox[origin=c]{0}{B} & \rotatebox[origin=c]{0}{C} & \rotatebox[origin=c]{0}{ET} & \rotatebox[origin=c]{0}{P} & \rotatebox[origin=c]{0}{JC}
        & Mean$\downarrow$ \\\hline
        \multicolumn{17}{c}{\textbf{ImageNet-C}}\\\hline
        Block & 40.5$\pm0.2$ & 37.3$\pm0.4$ & 38.3$\pm1.0$ & 47.7$\pm1.2$ & 42.1$\pm0.5$ & 38.9$\pm0.8$ & 39.4$\pm1.2$ & 31.9$\pm1.0$ & 39.7$\pm8.0$ & 36.3$\pm5.4$ & 21.1$\pm0.1$ & 55.6$\pm12.6$ & 31.6$\pm0.7$ & 27.7$\pm0.4$ & 28.4$\pm0.1$ & 37.1 \\
        Pixel & 42.0$\pm$0.3 & 39.1$\pm$0.4 & 41.7$\pm$1.5 & 64.3$\pm$4.7 & 54.0$\pm$7.3 & 71.1$\pm$17.6 & 65.3$\pm$7.7 & 56.3$\pm$37.8 & 54.9$\pm$34.8 & 38.2$\pm$1.2 & 23.0$\pm$0.5 & 75.3$\pm$20.8 & 74.1$\pm$4.0 & 29.9$\pm$0.2 & 31.6$\pm$0.8 & 50.7\\
        Patch & 41.5$\pm$0.4 & 39.1$\pm$0.3 & 41.0$\pm$0.8 & 66.3$\pm$5.1 & 49.1$\pm$4.5 & 45.1$\pm$4.2 & 57.4$\pm$1.9 & 32.9$\pm$0.2 & 38.4$\pm$3.9 & 33.6$\pm$0.9 & 22.0$\pm$0.3 & 82.0$\pm$14.6 & 32.9$\pm$0.7 & 28.6$\pm$0.4 & 28.9$\pm$0.6 & 42.6\\
        \midrule

        \multicolumn{17}{c}{\textbf{CIFAR10-C}}\\\hline
        Block & 15.8$\pm$0.5 & 14.0$\pm$0.0 & 9.8$\pm$0.5 & 9.3$\pm$0.0 & 22.8$\pm$0.3 & 9.4$\pm$0.5 & 7.7$\pm$0.6 & 7.5$\pm$0.0 & 7.5$\pm$0.2 & 8.8$\pm$0.1 & 4.4$\pm$0.2 & 8.4$\pm$0.3 & 13.2$\pm$0.9 & 9.9$\pm$0.2 & 13.8$\pm$1.1 & 10.8 \\
        Pixel & 16.7$\pm$1.0 & 15.2$\pm$0.7 & 9.1$\pm$0.4 & 10.5$\pm$0.3 & 57.3$\pm$3.9 & 11.6$\pm$1.1 & 9.4$\pm$0.4 & 8.0$\pm$0.3 & 7.1$\pm$0.2 & 8.0$\pm$0.2 & 4.8$\pm$0.1 & 9.7$\pm$0.6 & 15.0$\pm$0.6 & 64.5$\pm$3.0 & 15.9$\pm$0.6 & 17.5\\
        Patch & 24.5$\pm$3.4 & 19.5$\pm$2.4 & 12.5$\pm$0.5 & 9.9$\pm$0.2 & 27.5$\pm$3.5 & 10.9$\pm$0.3 & 9.1$\pm$0.3 & 8.6$\pm$0.5 & 8.5$\pm$0.2 & 11.1$\pm$0.2 & 4.6$\pm$0.1 & 10.5$\pm$1.5 & 14.6$\pm$0.4 & 12.3$\pm$0.5 & 15.7$\pm$0.3 & 13.3 \\

        \bottomrule
    \end{tabular}%
    }

\end{table*}

\begin{table*}[h!]
    \centering
    \small
    \caption{SEGA ablation on ImageNet-C and CIFAR10-C CTTA (batch size 1) comparing erasing-selection strategies with ViT-Base; entries are error rates (\%). Mean is over 15 corruption domains. Gain is relative improvement over the source model. Entries report mean$\pm$std over three runs (seeds 1, 2, 3). }
    \label{tab:erasing_selection}
    \setlength\tabcolsep{6pt}
    \resizebox{\linewidth}{!}{
    \begin{tabular}{l|ccc|cccc|cccc|cccc|c}
        \toprule
        Time & \multicolumn{15}{c|}{$t\xrightarrow{\;}$}& \\ \hline
        Selection &
        \rotatebox[origin=c]{0}{GN} & \rotatebox[origin=c]{0}{SN} & \rotatebox[origin=c]{0}{IN} & \rotatebox[origin=c]{0}{DB} & \rotatebox[origin=c]{0}{GB} & \rotatebox[origin=c]{0}{MB} & \rotatebox[origin=c]{0}{ZB} & \rotatebox[origin=c]{0}{S} & \rotatebox[origin=c]{0}{Fr} & \rotatebox[origin=c]{0}{F}  & \rotatebox[origin=c]{0}{B} & \rotatebox[origin=c]{0}{C} & \rotatebox[origin=c]{0}{ET} & \rotatebox[origin=c]{0}{P} & \rotatebox[origin=c]{0}{JC}
        & Mean$\downarrow$ \\\hline
        \multicolumn{17}{c}{\textbf{ImageNet-C}}\\\hline
        Random      & 40.5$\pm0.2$ & 37.3$\pm0.4$ & 38.3$\pm1.0$ & 47.7$\pm1.2$ & 42.1$\pm0.5$ & 38.9$\pm0.8$ & 39.4$\pm1.2$ & 31.9$\pm1.0$ & 39.7$\pm8.0$ & 36.3$\pm5.4$ & 21.1$\pm0.1$ & 55.6$\pm12.6$ & 31.6$\pm0.7$ & 27.7$\pm0.4$ & 28.4$\pm0.1$ & 37.1 \\
        Center      & 40.6$\pm$0.2 & 37.1$\pm$0.2 & 38.7$\pm$0.5 & 49.7$\pm$2.6 & 42.3$\pm$0.6 & 39.0$\pm$0.9 & 38.6$\pm$1.1 & 33.1$\pm$1.0 & 34.0$\pm$1.4 & 53.5$\pm$38.4 & 21.7$\pm$0.4 & 48.4$\pm$2.3 & 32.9$\pm$0.4 & 26.9$\pm$0.9 & 28.1$\pm$0.6 & 37.6\\
        Uncertainty & 40.7$\pm$0.3 & 38.0$\pm$0.4 & 39.1$\pm$0.7 & 47.5$\pm$1.0 & 41.2$\pm$0.2 & 37.7$\pm$0.2 & 39.0$\pm$1.4 & 32.6$\pm$0.7 & 41.4$\pm$11.2 & 54.4$\pm$17.3 & 21.1$\pm$0.1 & 61.4$\pm$22.7 & 32.6$\pm$1.2 & 26.4$\pm$0.5 & 28.1$\pm$0.4 & 38.7\\
        \midrule

        \multicolumn{17}{c}{\textbf{CIFAR10-C}}\\\hline
        Random      & 15.8$\pm$0.5 & 14.0$\pm$0.0 & 9.8$\pm$0.5 & 9.3$\pm$0.0 & 22.8$\pm$0.3 & 9.4$\pm$0.5 & 7.7$\pm$0.6 & 7.5$\pm$0.0 & 7.5$\pm$0.2 & 8.8$\pm$0.1 & 4.4$\pm$0.2 & 8.4$\pm$0.3 & 13.2$\pm$0.9 & 9.9$\pm$0.2 & 13.8$\pm$1.1 & 10.8 \\
        Center      & 17.5$\pm$1.0 & 14.3$\pm$0.3 & 9.3$\pm$0.5 & 8.3$\pm$0.3 & 20.2$\pm$0.5 & 9.5$\pm$0.2 & 7.5$\pm$0.1 & 7.3$\pm$0.2 & 7.3$\pm$0.2 & 8.3$\pm$0.2 & 4.0$\pm$0.1 & 8.4$\pm$0.4 & 13.6$\pm$0.2 & 9.9$\pm$0.4 & 14.8$\pm$0.1 & 10.7\\
        Uncertainty & 15.4$\pm$0.4 & 13.3$\pm$0.7 & 9.8$\pm$0.2 & 8.8$\pm$0.1 & 20.0$\pm$1.5 & 9.7$\pm$0.1 & 7.7$\pm$0.4 & 7.6$\pm$0.3 & 7.1$\pm$0.3 & 8.9$\pm$0.1 & 4.3$\pm$0.2 & 8.0$\pm$0.3 & 13.4$\pm$0.1 & 9.9$\pm$0.1 & 14.8$\pm$0.6 & 10.6 \\

        \bottomrule
    \end{tabular}%
    }

\end{table*}

Fig.~\ref{fig:erasing} together with Tables~\ref{tab:erasing_family} and \ref{tab:erasing_selection} analyze SEGA's erasing mechanism across families and placement strategies, showing that block erasing yields smoother, more monotonic entropy trajectories and consistently lower error than pixel or patch erasing, while different block placements behave similarly. These results motivate random block erasing as a simple default that provides a stable, informative entropy gradient without requiring delicate tuning of the erasing pattern.

\subsection{Domain Sensitivity Timelines}
\label{sec:appendix-domain-sensitivity}
\begin{figure*}[h!]
    \centering
    \subfigure[GN]{
        \includegraphics[width=0.30\linewidth]{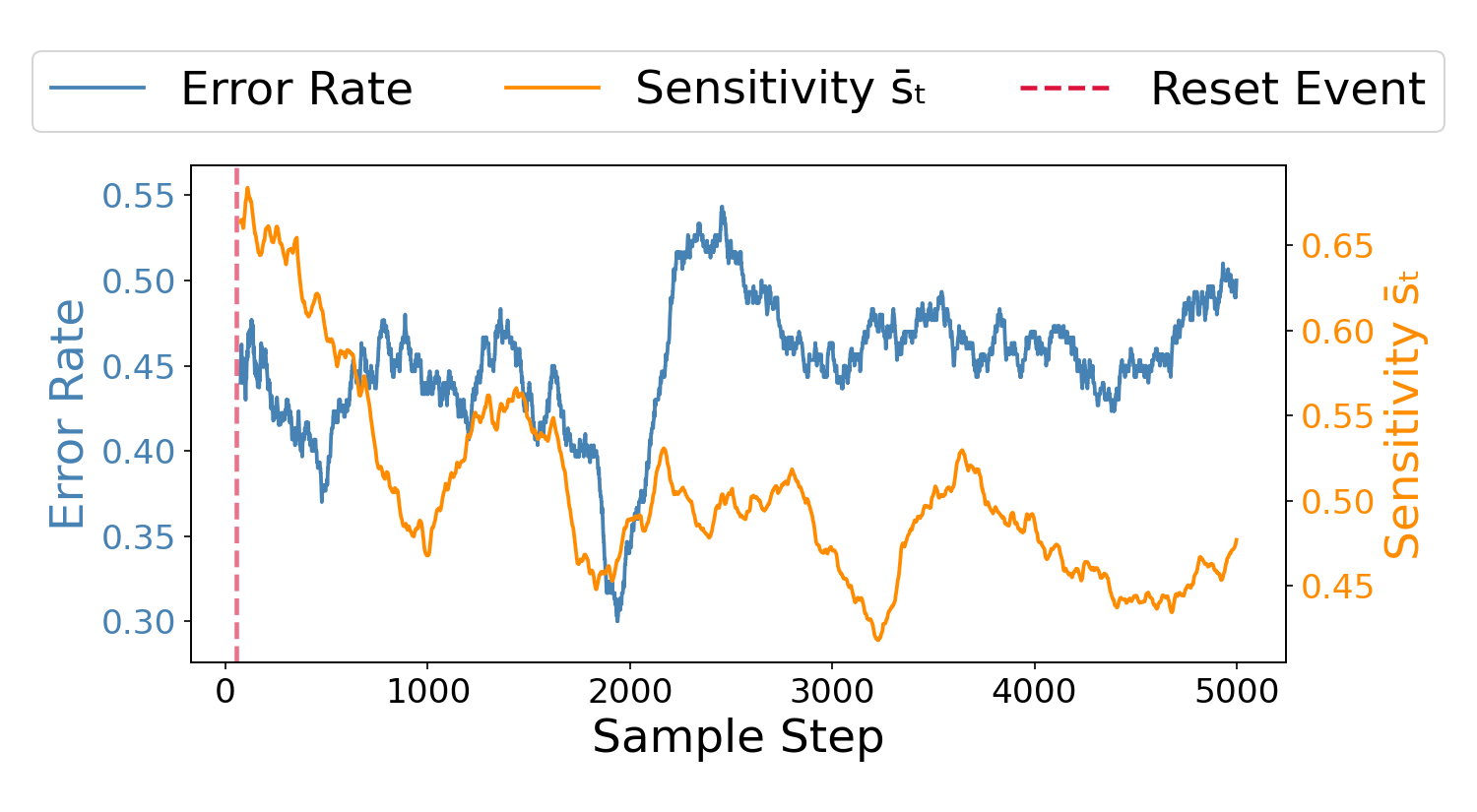}
    }
    \subfigure[SN]{
        \includegraphics[width=0.30\linewidth]{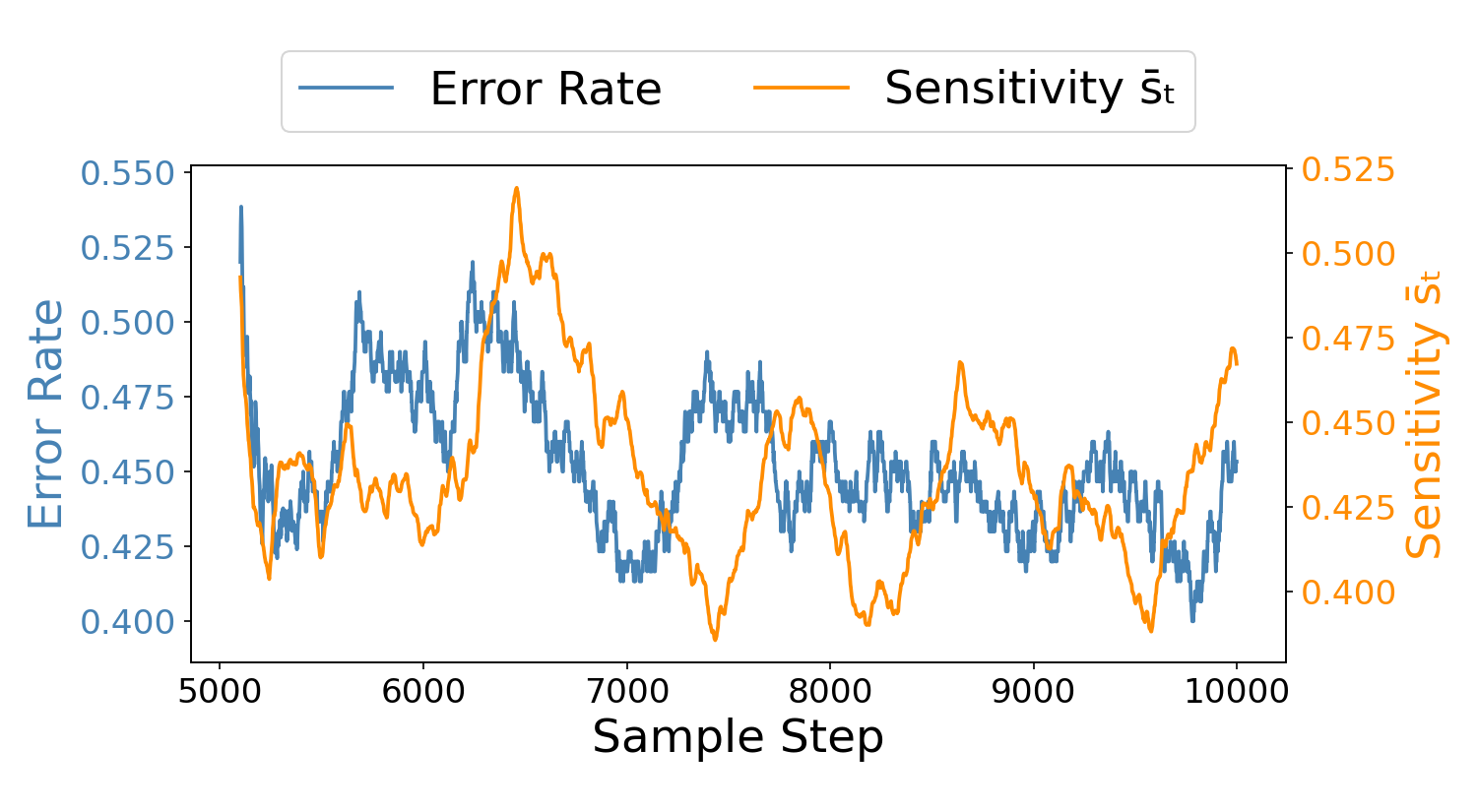}
    }
    \subfigure[IN]{
        \includegraphics[width=0.30\linewidth]{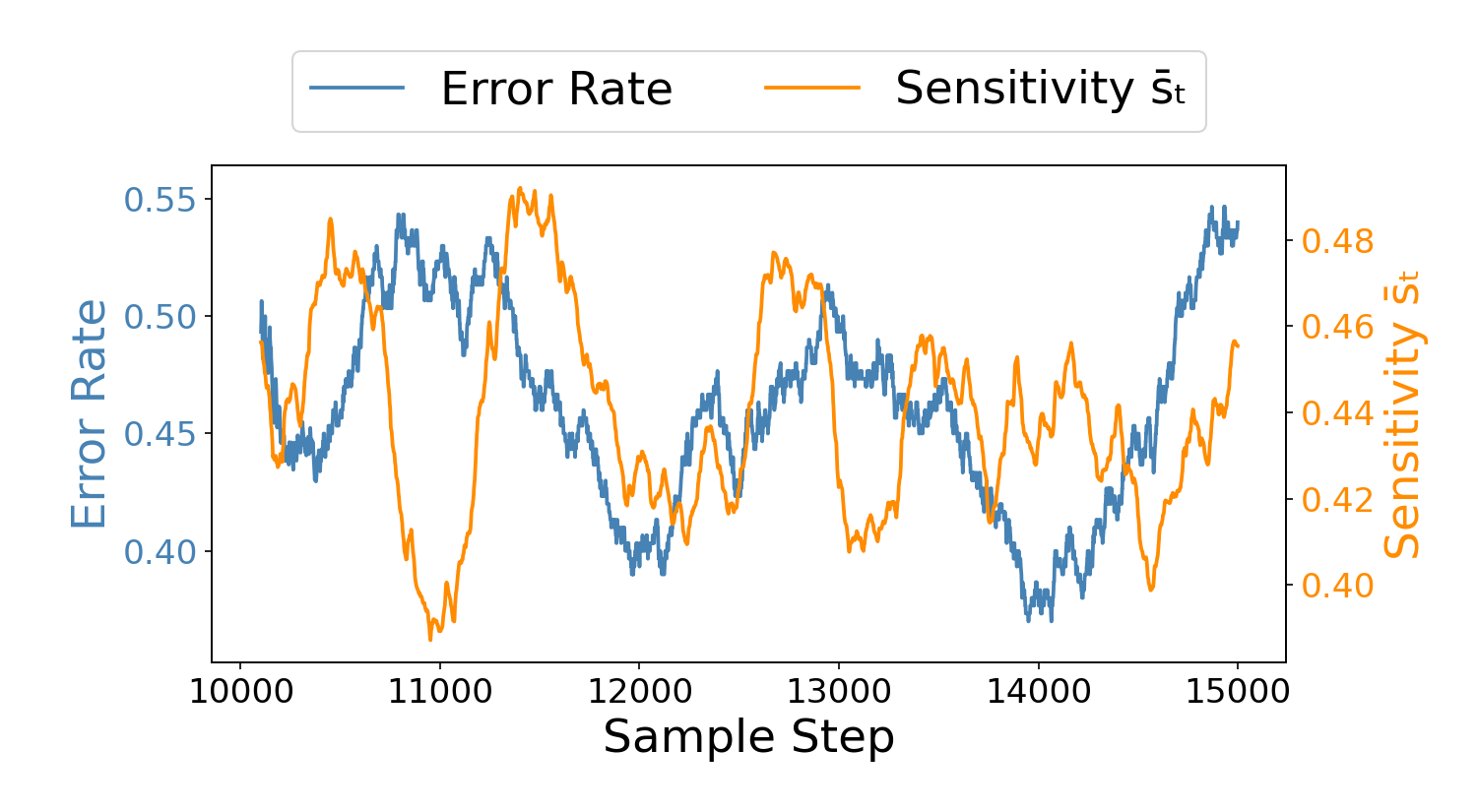}
    }\\[2pt]

    \subfigure[DB]{
        \includegraphics[width=0.30\linewidth]{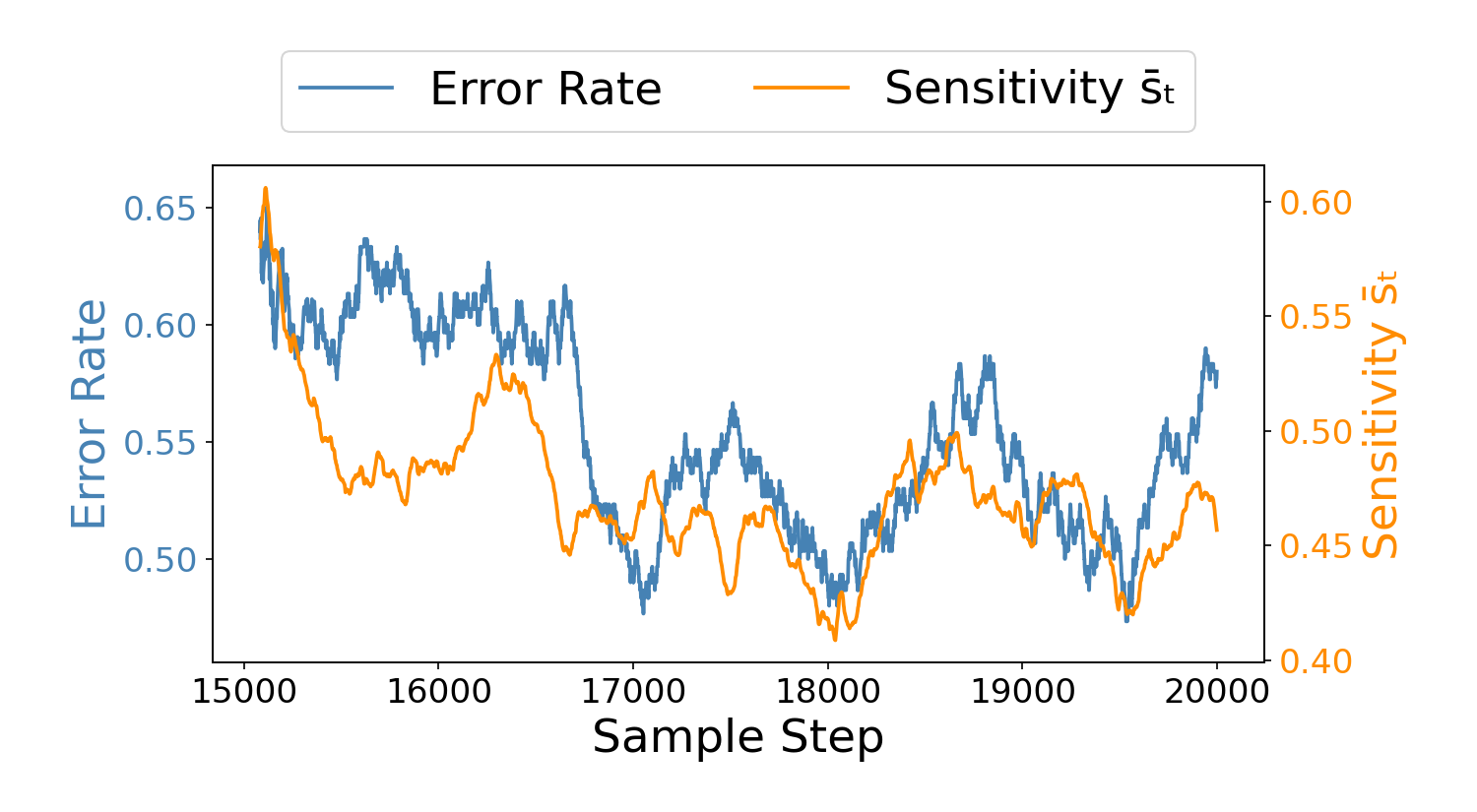}
    }
    \subfigure[GB]{
        \includegraphics[width=0.30\linewidth]{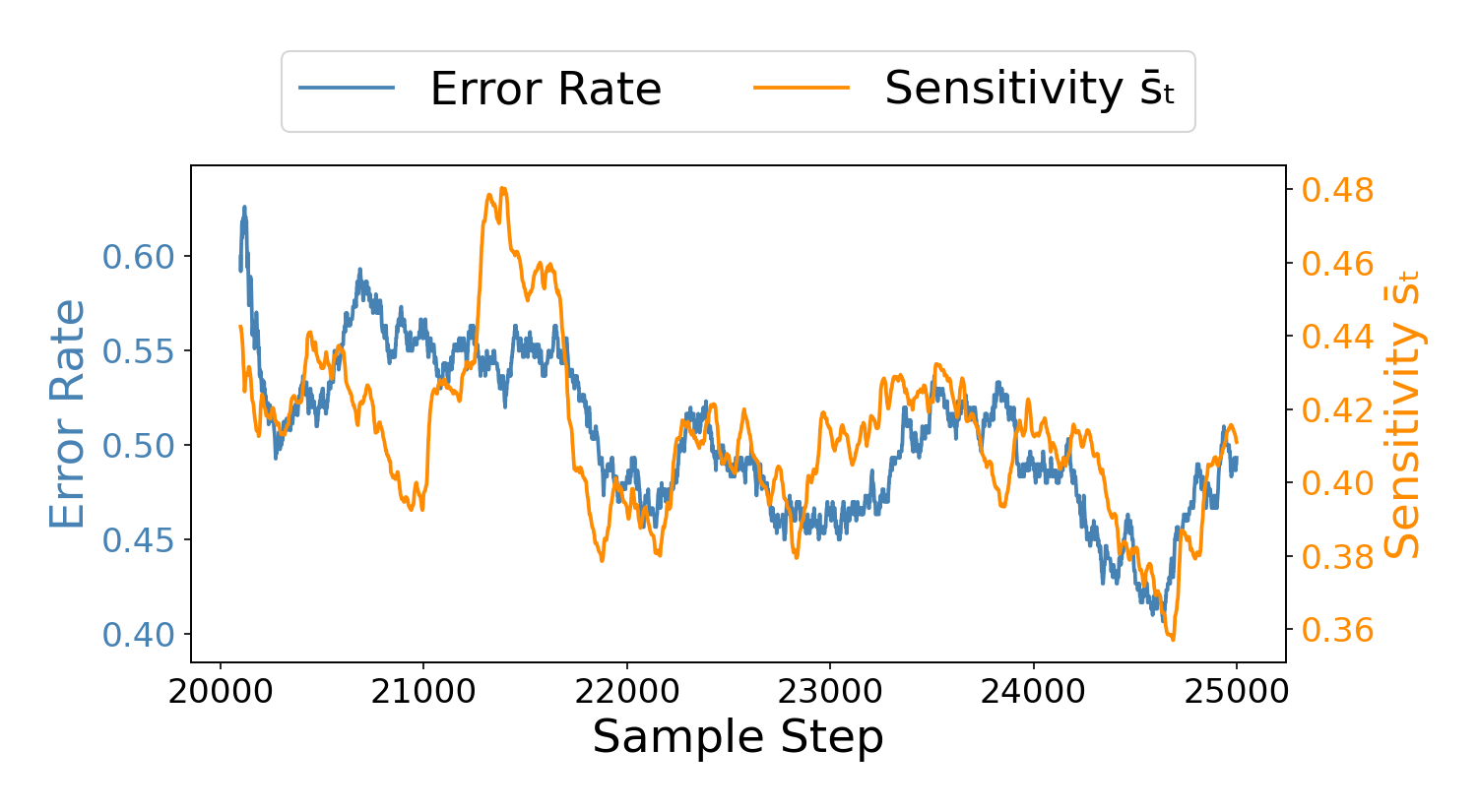}
    }
    \subfigure[MB]{
        \includegraphics[width=0.30\linewidth]{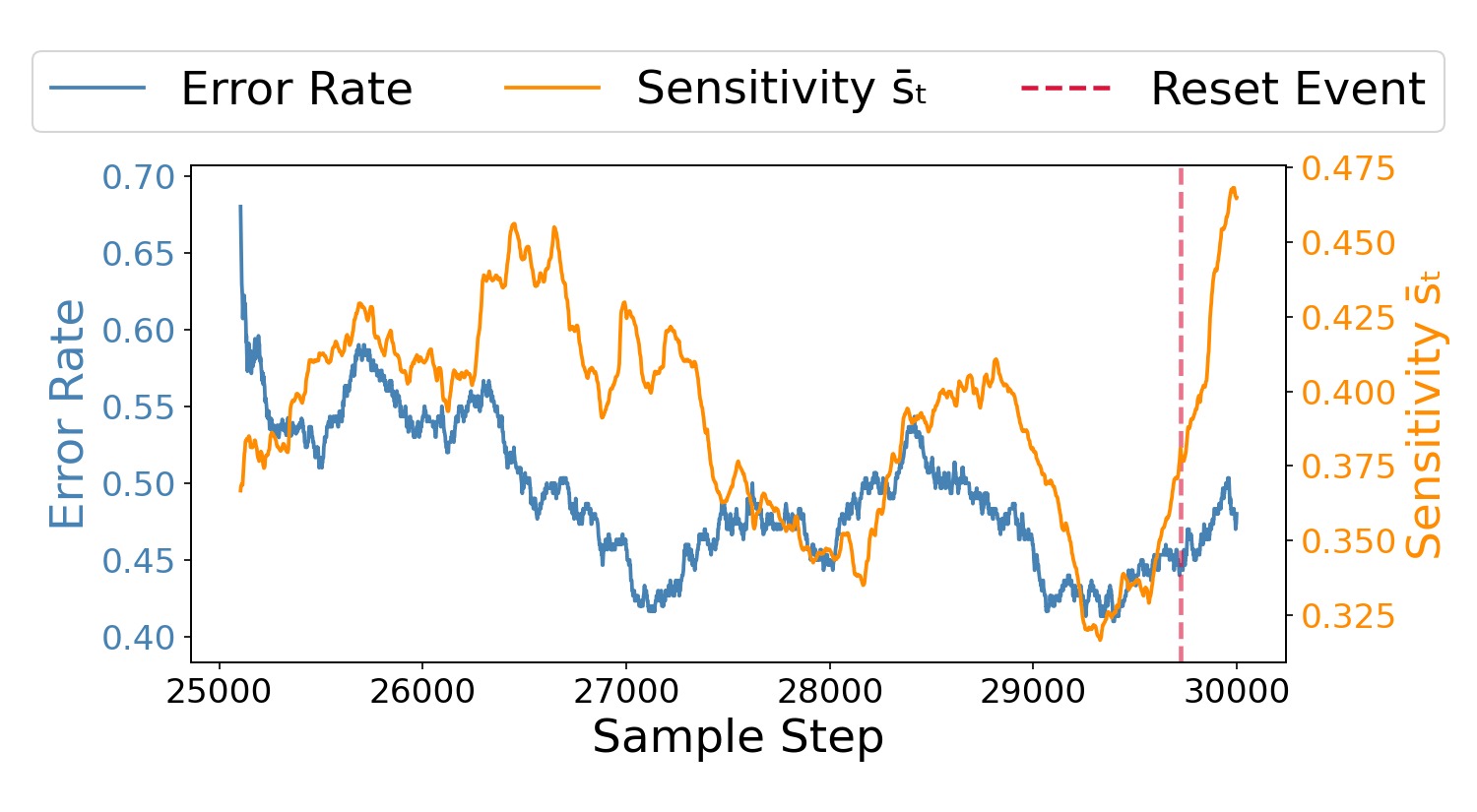}
    }\\[2pt]

    \subfigure[ZB]{
        \includegraphics[width=0.30\linewidth]{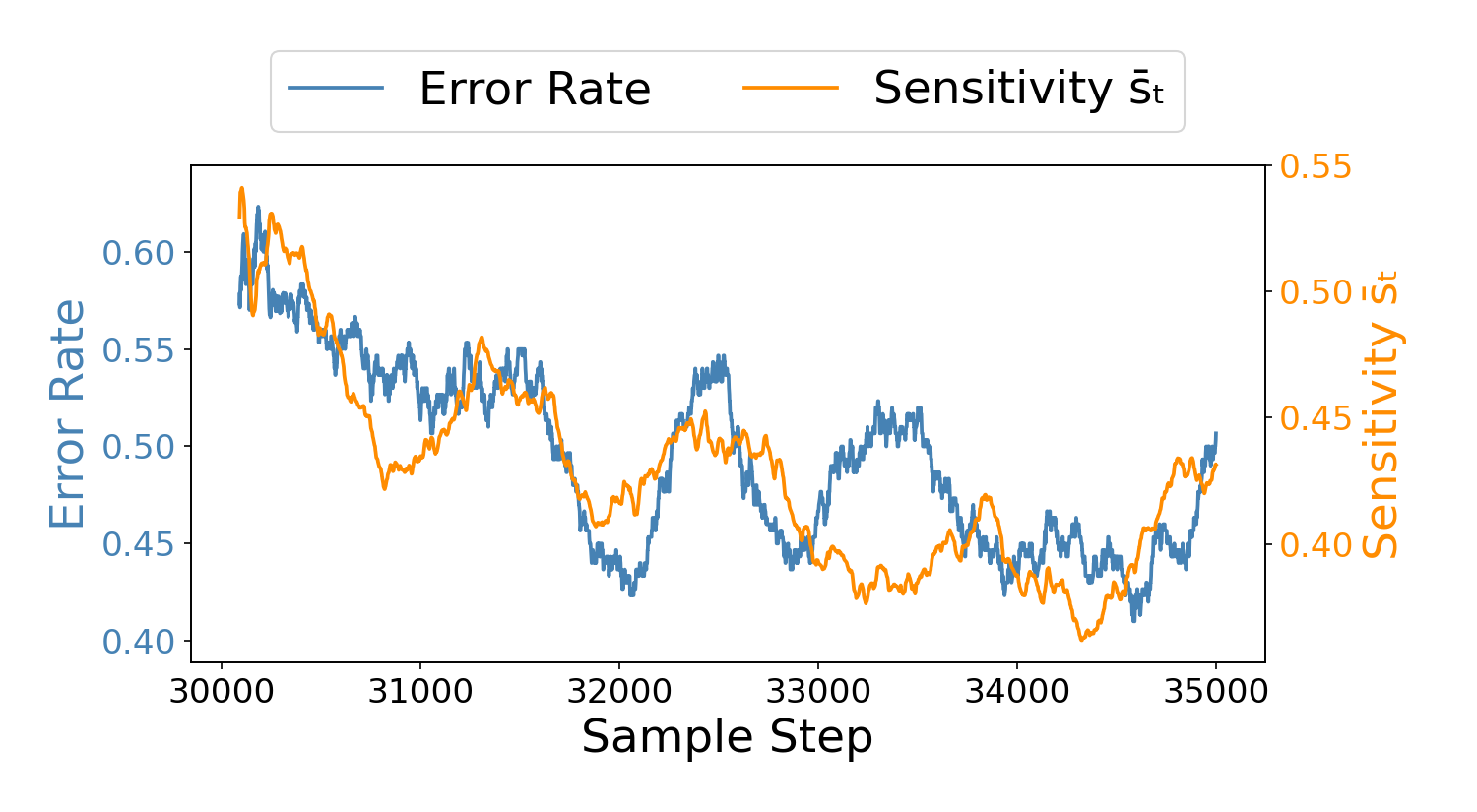}
    }
    \subfigure[S]{
        \includegraphics[width=0.30\linewidth]{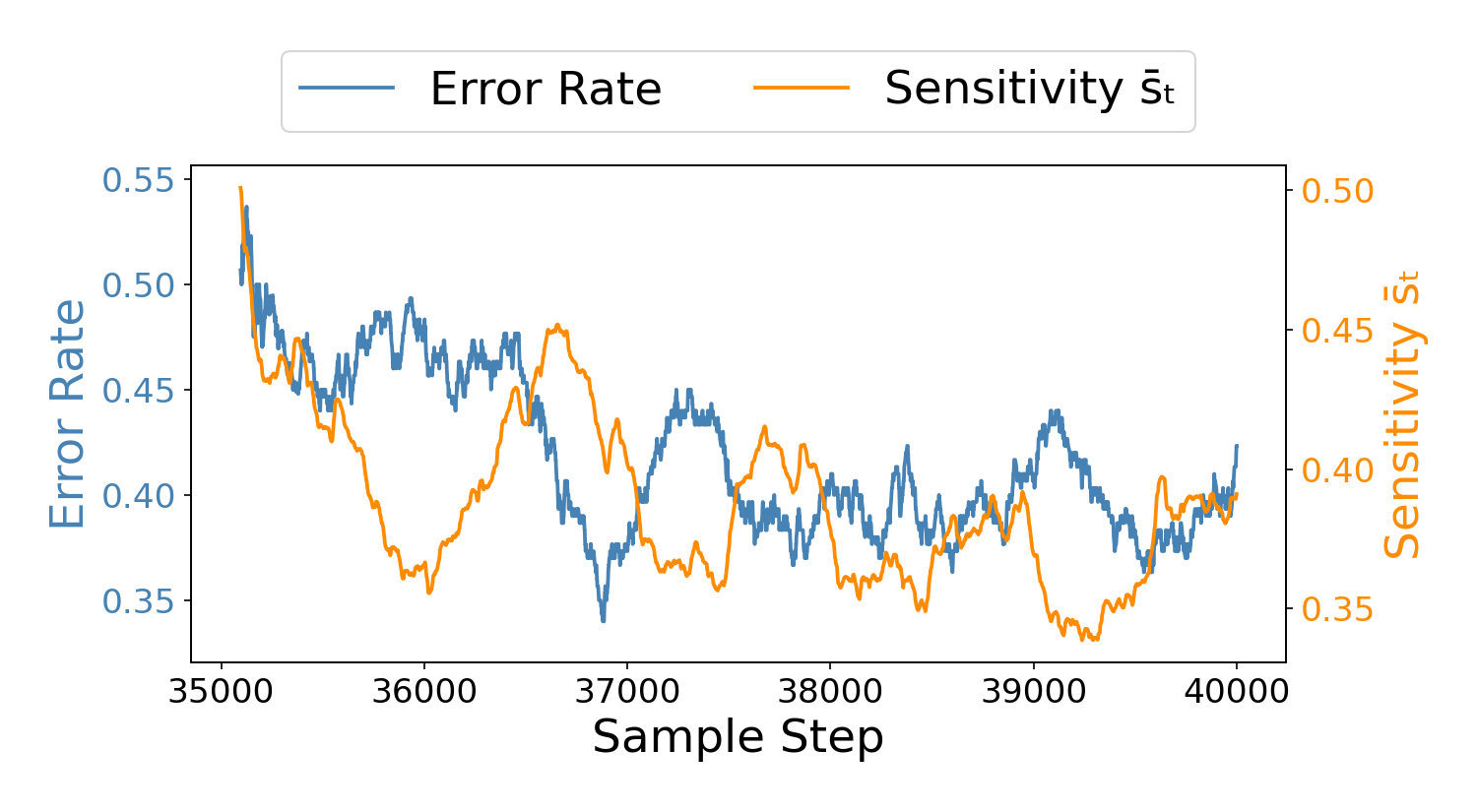}
    }
    \subfigure[Fr]{
        \includegraphics[width=0.30\linewidth]{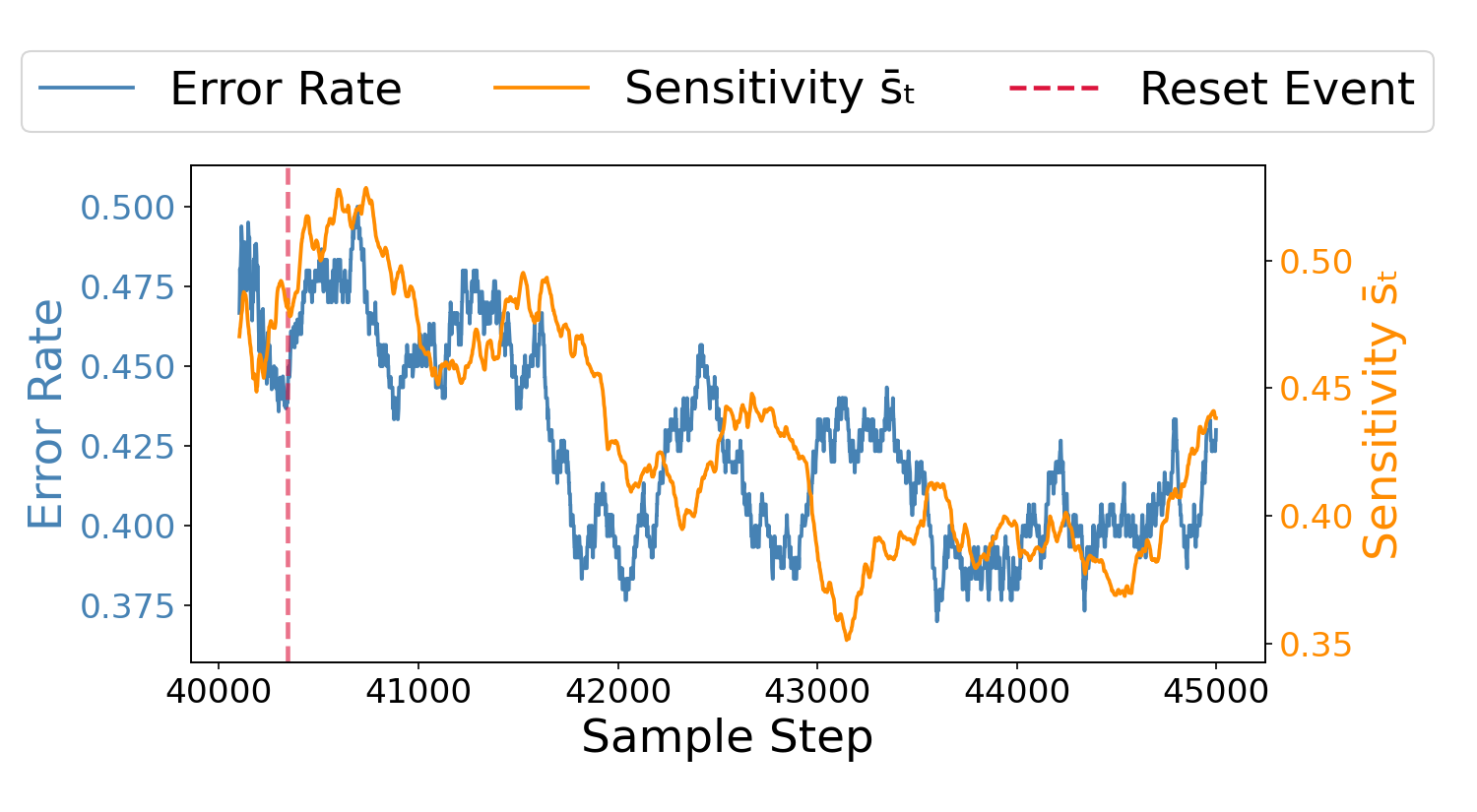}
    }\\[2pt]

    \subfigure[F]{
        \includegraphics[width=0.30\linewidth]{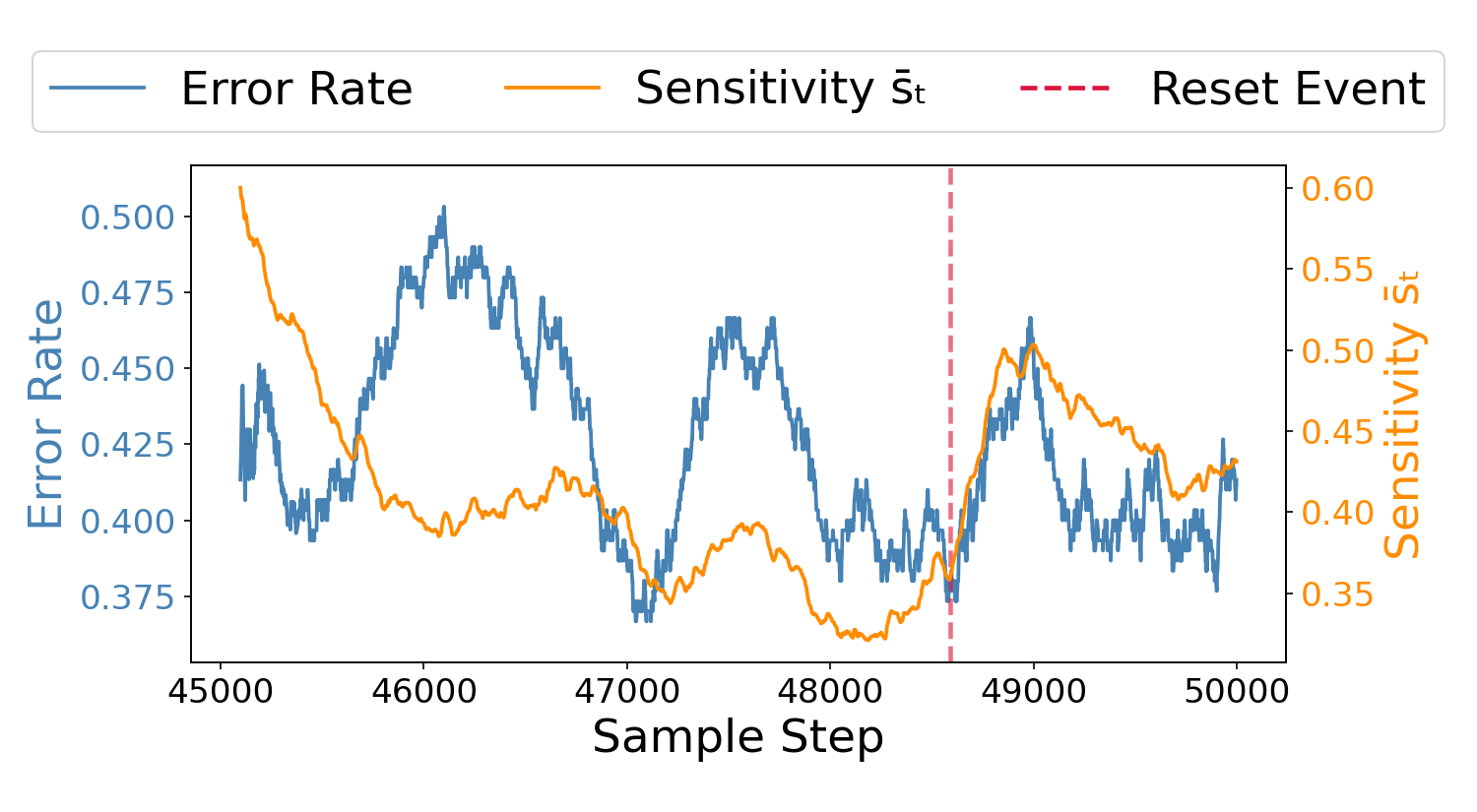}
    }
    \subfigure[B]{
        \includegraphics[width=0.30\linewidth]{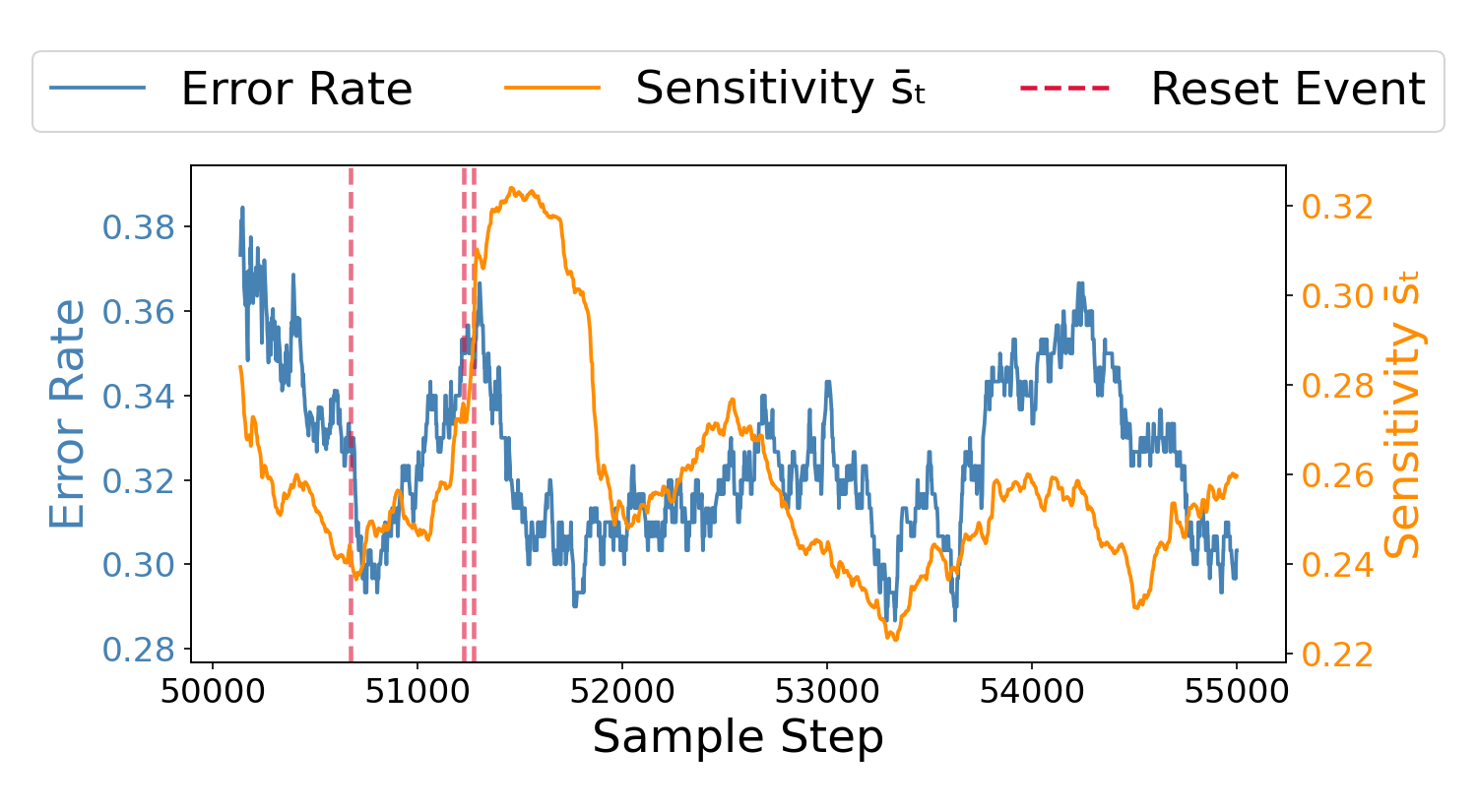}
    }
    \subfigure[C]{
        \includegraphics[width=0.30\linewidth]{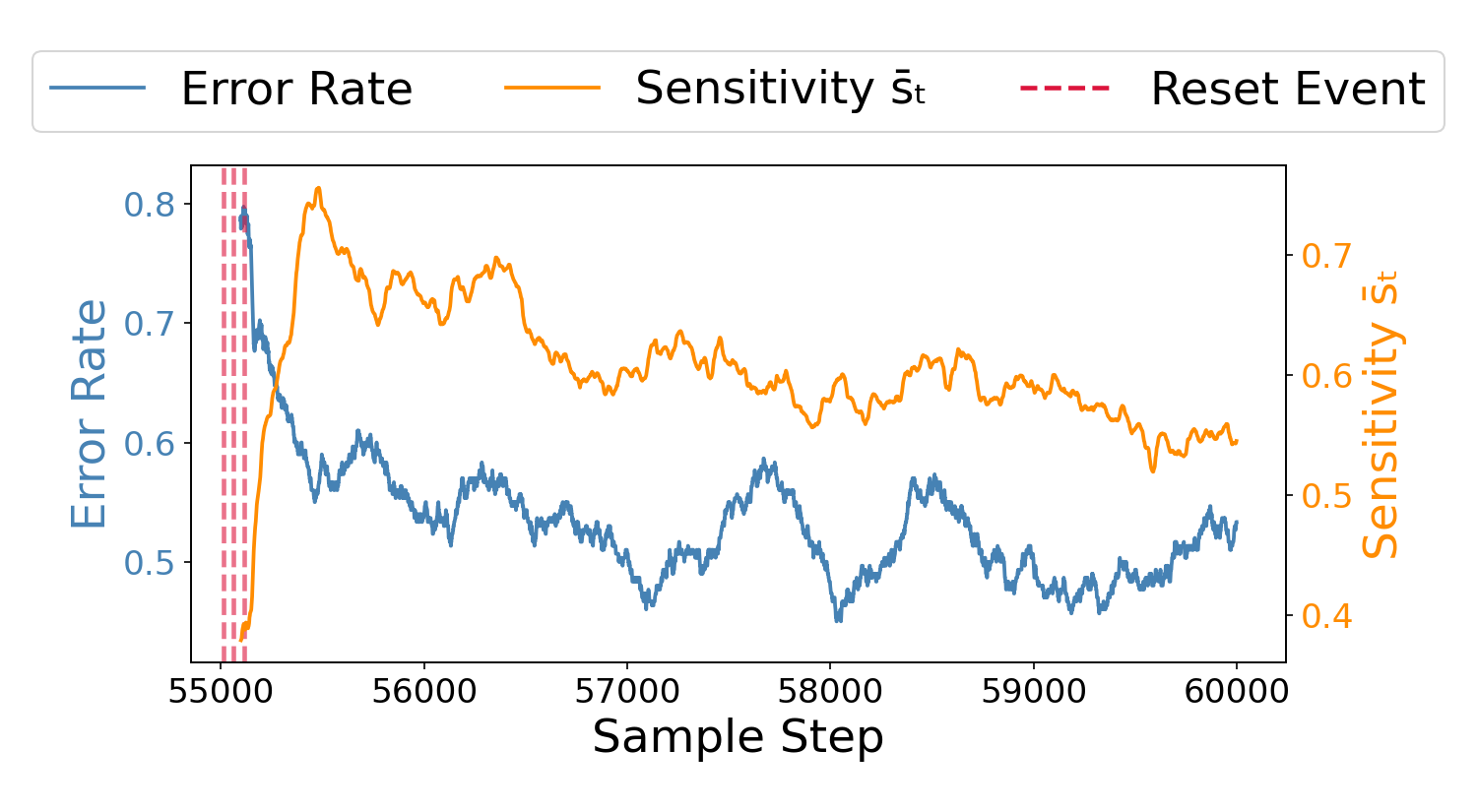}
    }\\[2pt]

    \subfigure[ET]{
        \includegraphics[width=0.30\linewidth]{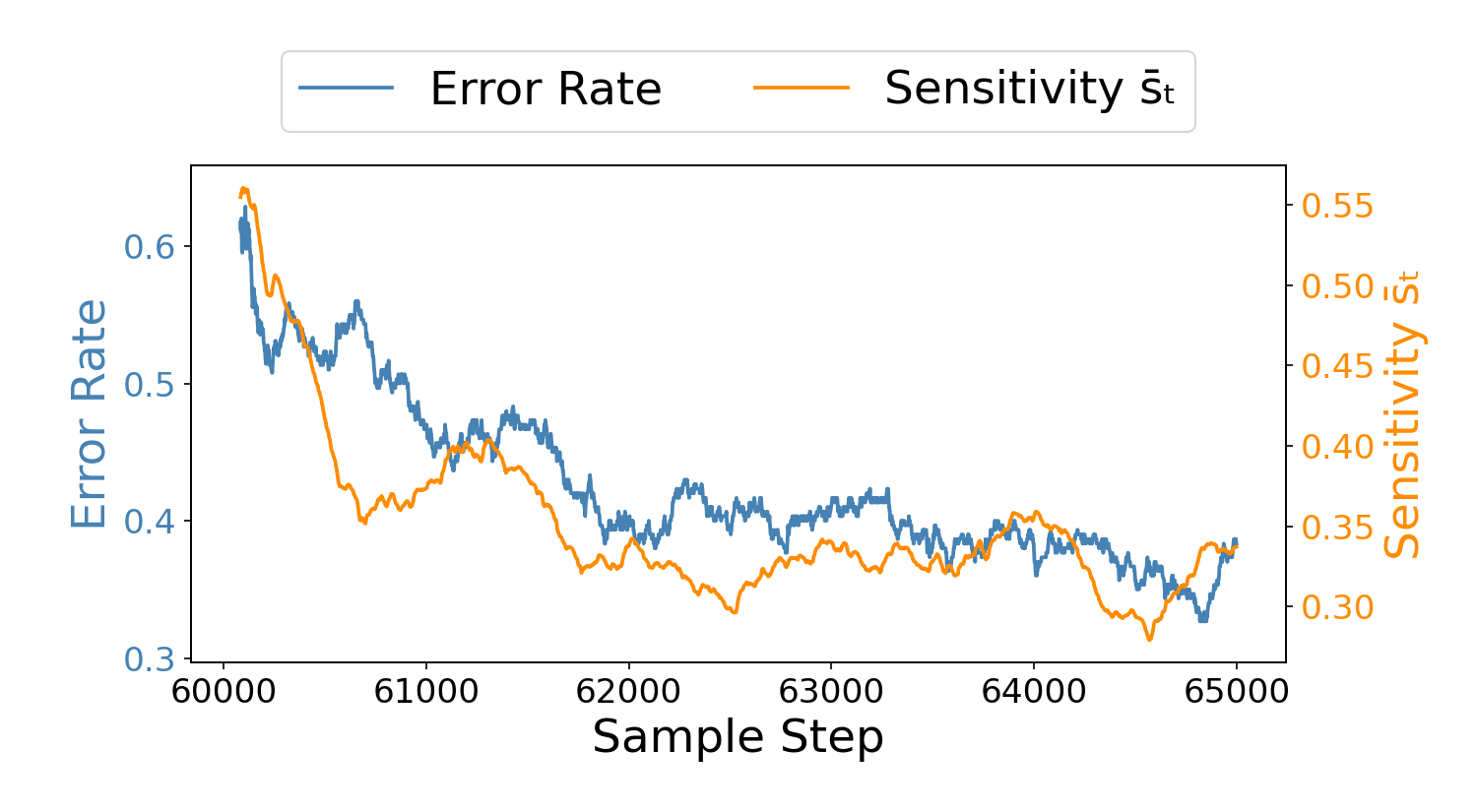}
    }
    \subfigure[P]{
        \includegraphics[width=0.30\linewidth]{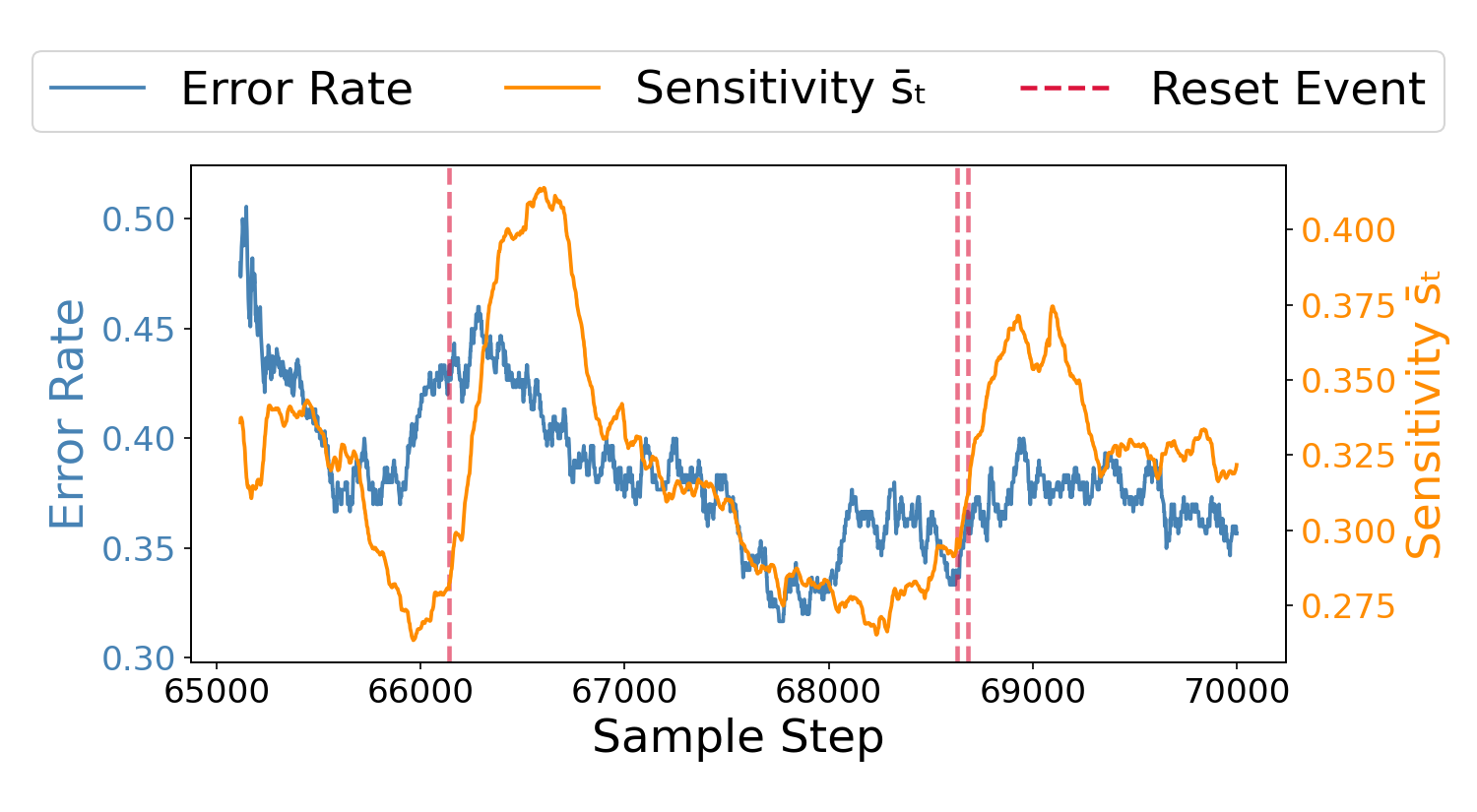}
    }
    \subfigure[JC]{
        \includegraphics[width=0.30\linewidth]{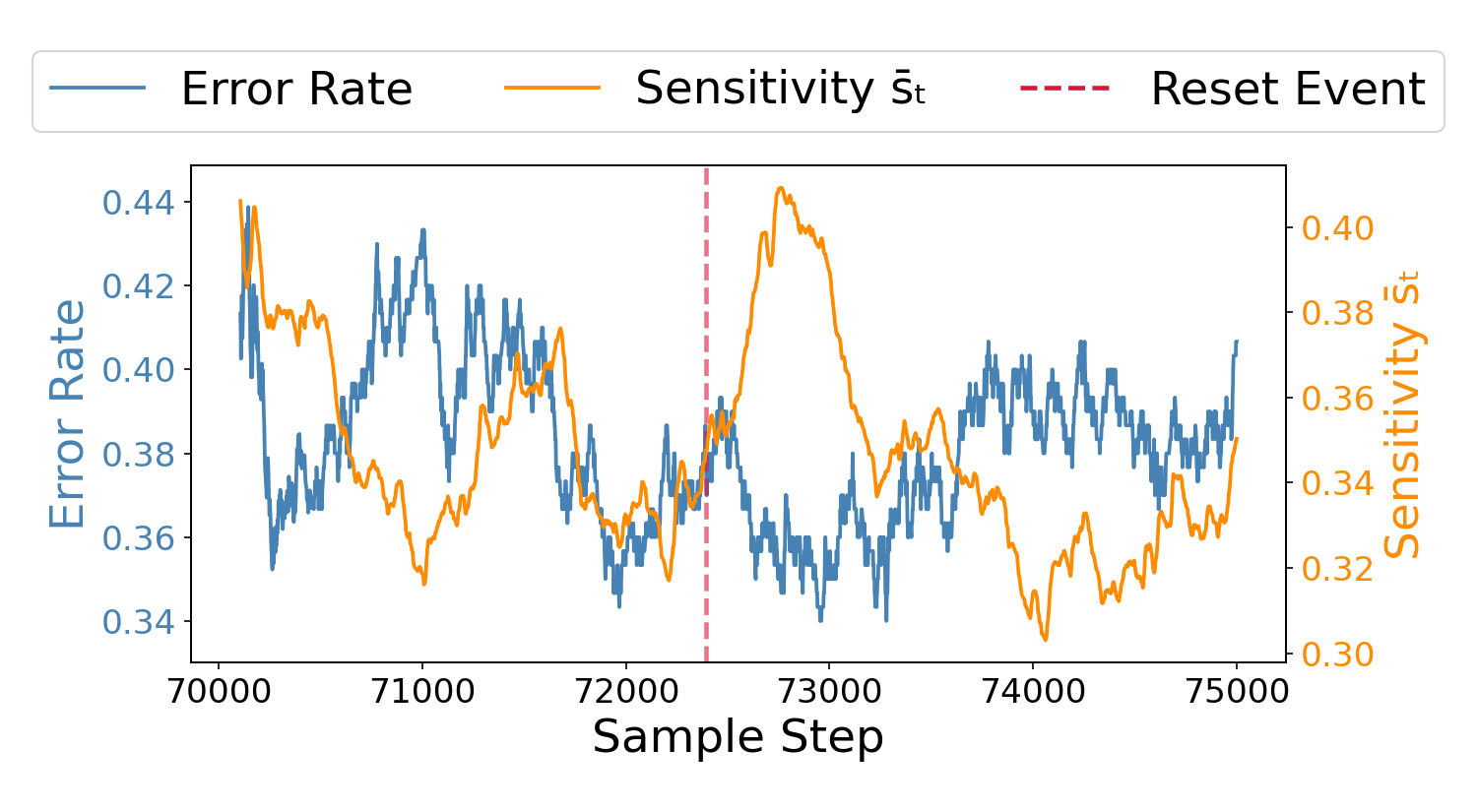}
    }

    \caption{Per-domain operational sensitivity timelines computed by SEGA on ImageNet-C under strict online CTTA ($B{=}1$). Each subplot shows the per-sample sensitivity trajectory within a single corruption domain.}
    \label{fig:domain_sensitivity}
\end{figure*}

Fig.~\ref{fig:domain_sensitivity} visualizes the per-domain operational sensitivity trajectories that SEGA computes during strict online CTTA on FreshFish-C. Each subplot shows the per-sample sensitivity score $s_b$ within a single corruption domain, revealing how the entropy response under erasing varies across noise, blur, weather, and compression artifacts. Harder domains such as Gaussian noise and pixelate exhibit elevated and more variable sensitivity, whereas milder shifts like brightness produce flatter trajectories. These timelines illustrate the empirical basis for the stability budget and quantile gate: the trend rule detects sustained upward drift in these per-domain signals, while the gate filters out the lowest-sensitivity samples that contribute little useful gradient.

\begin{table*}[h!]
    \centering
    \caption{Shuffled domain CTTA using ViT-Base (batch size 1) on ImageNet-C and CIFAR10-C using ViT-Base. Entries report classification error rate (\%).}
    \label{tab:shuffle}
    \tiny
    \setlength\tabcolsep{4pt}
    \begin{adjustbox}{width=\textwidth,center=\textwidth}
    \begin{tabular}{l|ccccc|c|ccccc|c}
        \toprule
        \multirow{3}{*}{Method} & \multicolumn{6}{c|}{\textbf{ImageNet-C}} & \multicolumn{6}{c}{\textbf{CIFAR10-C}} \\
        \cmidrule(lr){2-7} \cmidrule(lr){8-13}
        & Pass 1 & Pass 2 & Pass 3 & Pass 4 & Pass 5 & Mean$\downarrow$ & Pass 1 & Pass 2 & Pass 3 & Pass 4 & Pass 5 & Mean$\downarrow$ \\
        \midrule
        SOURCE          & 55.8 & 55.8 & 55.8 & 55.8 & 55.8 & 55.8     & 28.2 & 28.2 & 28.2 & 28.2 & 28.2 & 28.2 \\
        SAR             & 45.3 & 96.6 & 96.7 & 53.8 & 58.2 & 70.1     & 26.9 & 27.0 & 26.9 & 26.9 & 26.9 & 26.9 \\
        RDUMB           & 40.5 & 40.5 & 40.5 & 40.5 & 40.5 & 40.5     & 15.9 & 15.9 & 15.9 & 15.9 & 15.9 & 15.9 \\
        M2A             & 37.0 & 36.9 & 37.3 & 85.8 & 86.9 & 56.8     & 10.5 & 11.6 & 11.0 & 11.4 & 11.4 & 11.2 \\
        SEGA            & 41.1 & 36.4 & 42.3 & 41.2 & 41.1 & 40.4     & 10.8 & 11.1 & 10.8 & 11.2 & 11.3 & 11.0 \\

        \bottomrule
    \end{tabular}
    \end{adjustbox}
\end{table*}
\begin{table}[t!]
    \centering
    \caption{Lifelong CTTA (batch size 1) on FreshFish-C using ViT-Base. Entries report classification error rate (\%).}
    \label{tab:lifelong}
    \footnotesize
    \setlength\tabcolsep{4pt}
    \begin{adjustbox}{width=\linewidth,center=\linewidth}
    \begin{tabular}{l|ccccc|c}
        \toprule
        \multirow{3}{*}{Method} & \multicolumn{6}{c}{\textbf{FreshFish-C}} \\
        \cmidrule(lr){2-7}
         & Pass 1 & Pass 2 & Pass 3 & Pass 4 & Pass 5 & Mean$\downarrow$ \\
         \midrule
        SOURCE          & 28.9 & 28.9 & 28.9 & 28.9 & 28.9 & 28.9 \\
        SAR             & 27.8 & 27.8 & 27.8 & 27.8 & 27.8 & 27.8 \\
        RDUMB           & 27.1 & 26.8 & 27.1 & 26.8 & 27.1 & 27.0 \\
        M2A             & 20.2 & 16.2 & 14.4 & 13.5 & 12.8 & 15.4 \\
        SEGA            & 14.5 & 14.0 & 13.2 & 12.9 & 17.6 & 14.4 \\

        \bottomrule
    \end{tabular}
    \end{adjustbox}
\end{table}
\subsection{Shuffled and Lifelong CTTA}
\label{sec:appendix-shuffle-lifelong}
We probe stability under domain recursion and extended horizons. Table~\ref{tab:shuffle} reports error across five shuffled passes through ImageNet-C and CIFAR10-C, where corruption domains are revisited in random order, highlighting how SAR, RDumb, M2A, and SEGA behave when the domain schedule is no longer aligned with a single forward sweep. Table~\ref{tab:lifelong} then evaluates lifelong CTTA on the FreshFish-C aquaculture stream over five passes, summarizing how methods cope with very long, non-stationary streams. Together these results show that SEGA maintains competitive or improved mean error without relying on fixed domain ordering or short evaluation horizons.

\subsection{Carla CTTA Analysis}
\label{sec:appendix-carla}
\begin{table*}[h!]
    \centering
    \small
    \caption{Semantic segmentation mIoU (\%) results on Carla dataset under CTTA. We use DeepLabV2 as the segmentation model.}
    \label{tab:carla_table}
    \setlength\tabcolsep{4pt}
    \resizebox{0.95\linewidth}{!}{
    \begin{tabular}{l|cccccccccccccc|c}
        \toprule
        Method  &
        \rotatebox[origin=c]{65}{Road} & \rotatebox[origin=c]{65}{Sidewalk} & \rotatebox[origin=c]{65}{Building} & \rotatebox[origin=c]{65}{Wall} & \rotatebox[origin=c]{65}{Fence} & \rotatebox[origin=c]{65}{Pole} & \rotatebox[origin=c]{65}{TrafficLight} & \rotatebox[origin=c]{65}{TrafficSign} & \rotatebox[origin=c]{65}{Vegetation} & \rotatebox[origin=c]{65}{Terrain}  & \rotatebox[origin=c]{65}{Sky} & \rotatebox[origin=c]{65}{Person} & \rotatebox[origin=c]{65}{Vehicle} & \rotatebox[origin=c]{65}{Roadline} & Mean$\uparrow$\\\hline
        SOURCE & 85.5$\pm$0.0 & 62.7$\pm$0.0 & 52.7$\pm$0.0 & 46.2$\pm$0.0 & 24.3$\pm$0.0 & 40.8$\pm$0.0 & 55.9$\pm$0.0 & 51.9$\pm$0.0 & 51.5$\pm$0.0 & 24.1$\pm$0.0 & 15.7$\pm$0.0 & 66.9$\pm$0.0 & 81.7$\pm$0.0 & 78.6$\pm$0.0 & 52.8\\
        SAR    & 87.7$\pm$0.3 & 77.0$\pm$0.0 & 70.5$\pm$0.0 & 44.2$\pm$0.2 & 20.7$\pm$0.1 & 45.6$\pm$0.1 & 66.7$\pm$0.1 & 57.2$\pm$0.0 & 58.9$\pm$0.1 & 24.6$\pm$0.2 & 36.8$\pm$0.1 & 68.1$\pm$0.0 & 65.4$\pm$0.2 & 71.9$\pm$0.1 & 56.8\\
        RDUMB  & 86.5$\pm$0.0 & 77.2$\pm$0.0 & 71.4$\pm$0.0 & 43.8$\pm$0.0 & 20.1$\pm$0.0 & 45.0$\pm$0.0 & 66.6$\pm$0.0 & 57.1$\pm$0.0 & 58.7$\pm$0.0 & 23.9$\pm$0.0 & 38.6$\pm$0.0 & 67.9$\pm$0.0 & 62.5$\pm$0.0 & 71.7$\pm$0.0 & 56.5\\
        M2A    & 92.9$\pm$0.2 & 78.5$\pm$0.0 & 74.5$\pm$0.0 & 46.8$\pm$0.1 & 19.4$\pm$0.1 & 45.4$\pm$0.0 & 67.1$\pm$0.1 & 57.5$\pm$0.0 & 60.3$\pm$0.1 & 24.1$\pm$0.0 & 38.7$\pm$0.1 & 68.4$\pm$0.1 & 81.3$\pm$0.7 & 73.9$\pm$0.1 & 59.2\\
        SEGA   & 92.9$\pm$0.2 & 78.5$\pm$0.0 & 74.6$\pm$0.0 & 46.8$\pm$0.1 & 19.3$\pm$0.1 & 45.4$\pm$0.0 & 67.1$\pm$0.1 & 57.5$\pm$0.0 & 60.3$\pm$0.1 & 24.0$\pm$0.0 & 38.8$\pm$0.1 & 68.3$\pm$0.1 & 81.3$\pm$0.6 & 73.9$\pm$0.0 & 59.2\\
        \bottomrule
    \end{tabular}%
    }
\end{table*}
\begin{figure*}[h!]
    \centering
    \subfigure[Input]{
        \includegraphics[width=0.3\linewidth]{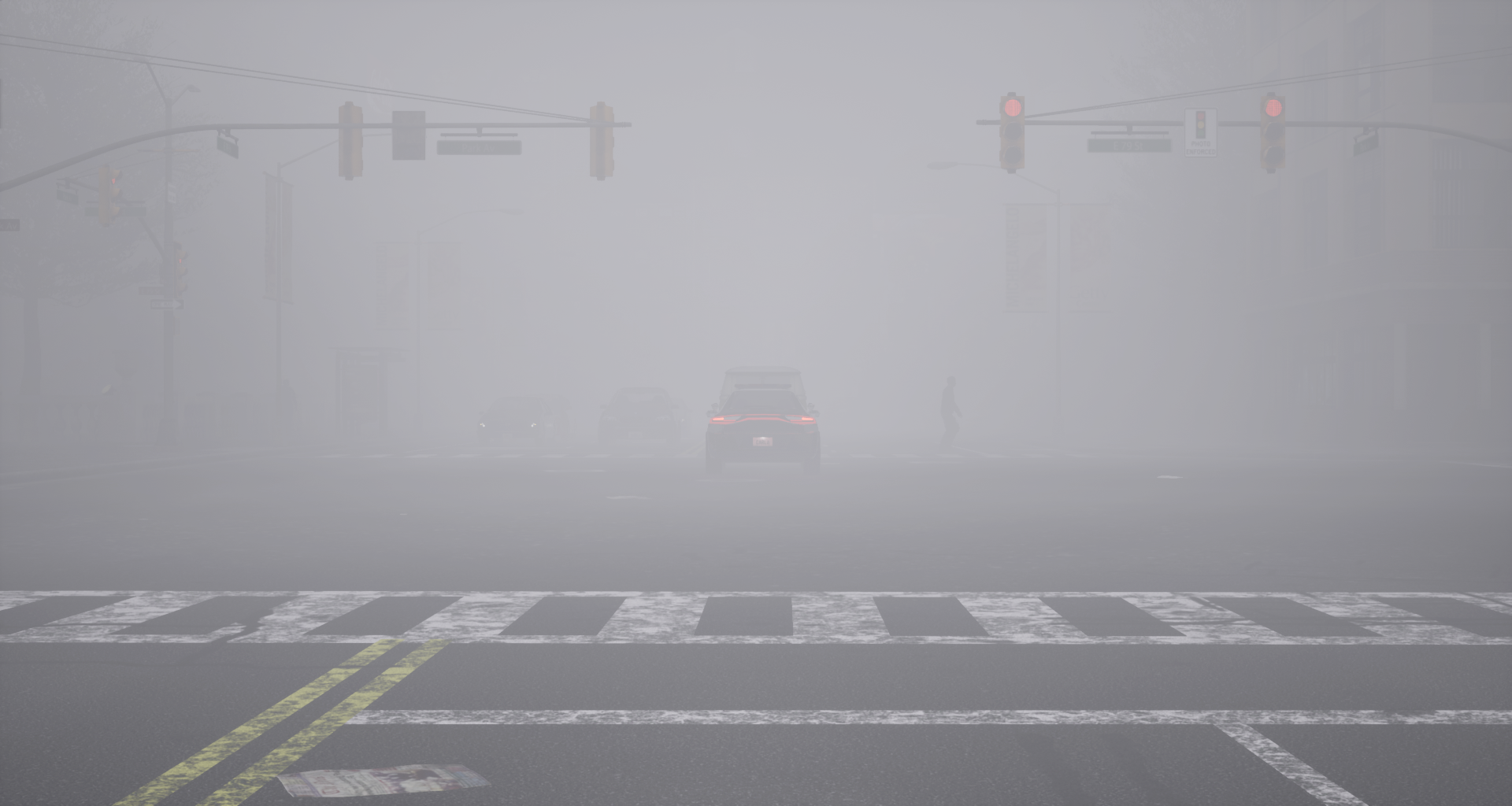}
    }
    \subfigure[Source]{
        \includegraphics[width=0.3\linewidth]{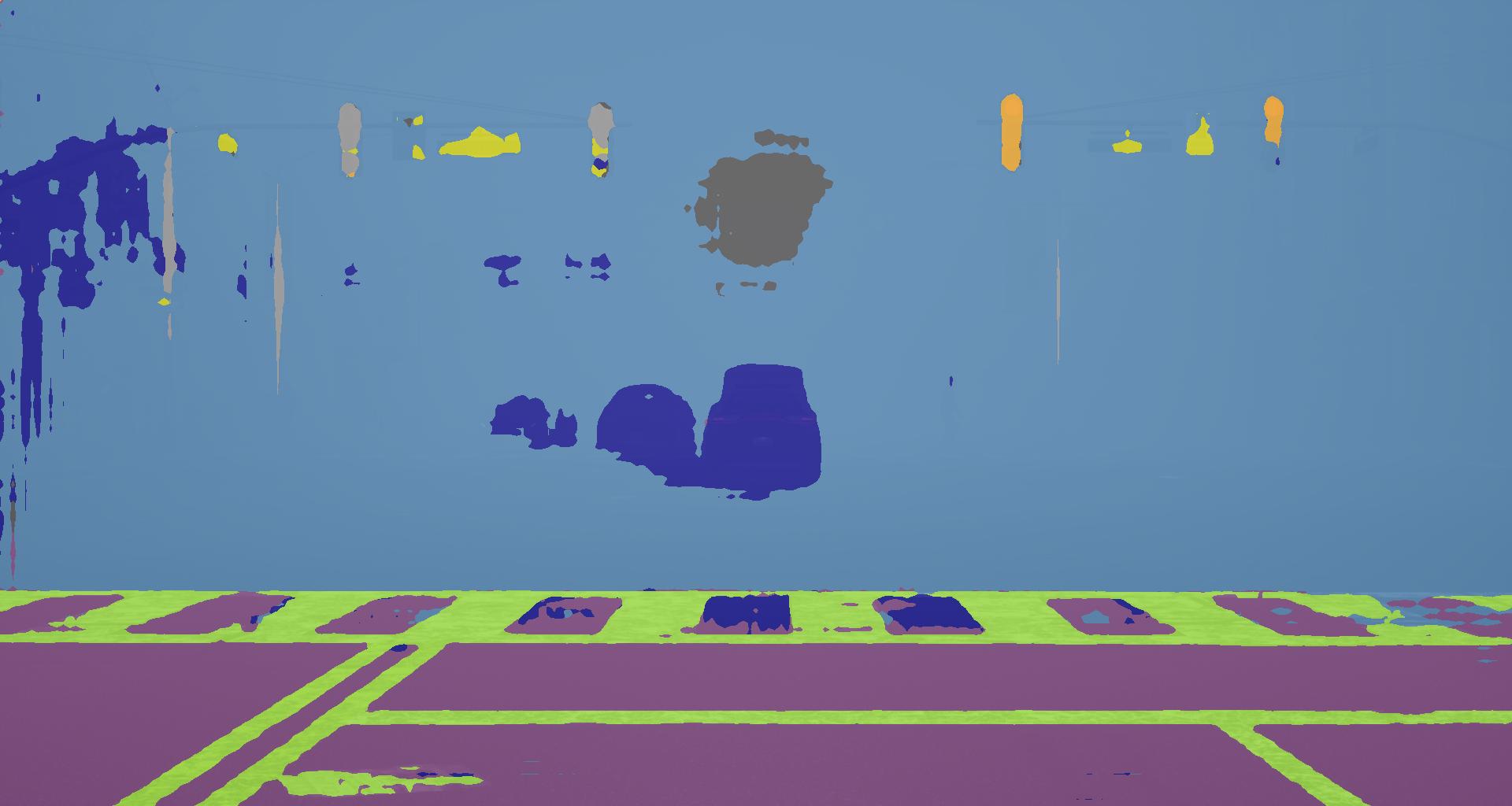}
    }
    \subfigure[SAR]{
        \includegraphics[width=0.3\linewidth]{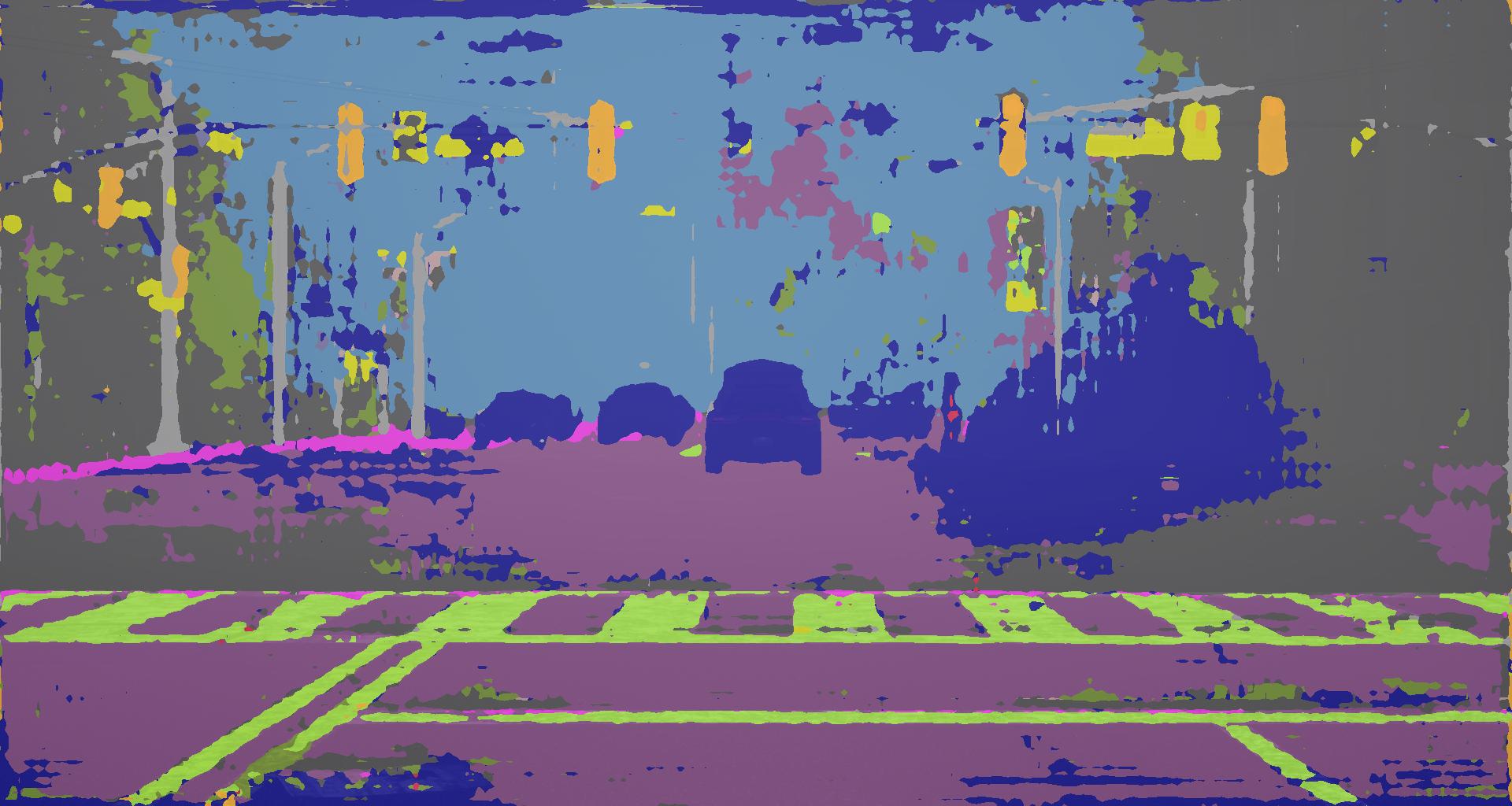}
    }\\[2pt]

    \subfigure[RDUMB]{
        \includegraphics[width=0.3\linewidth]{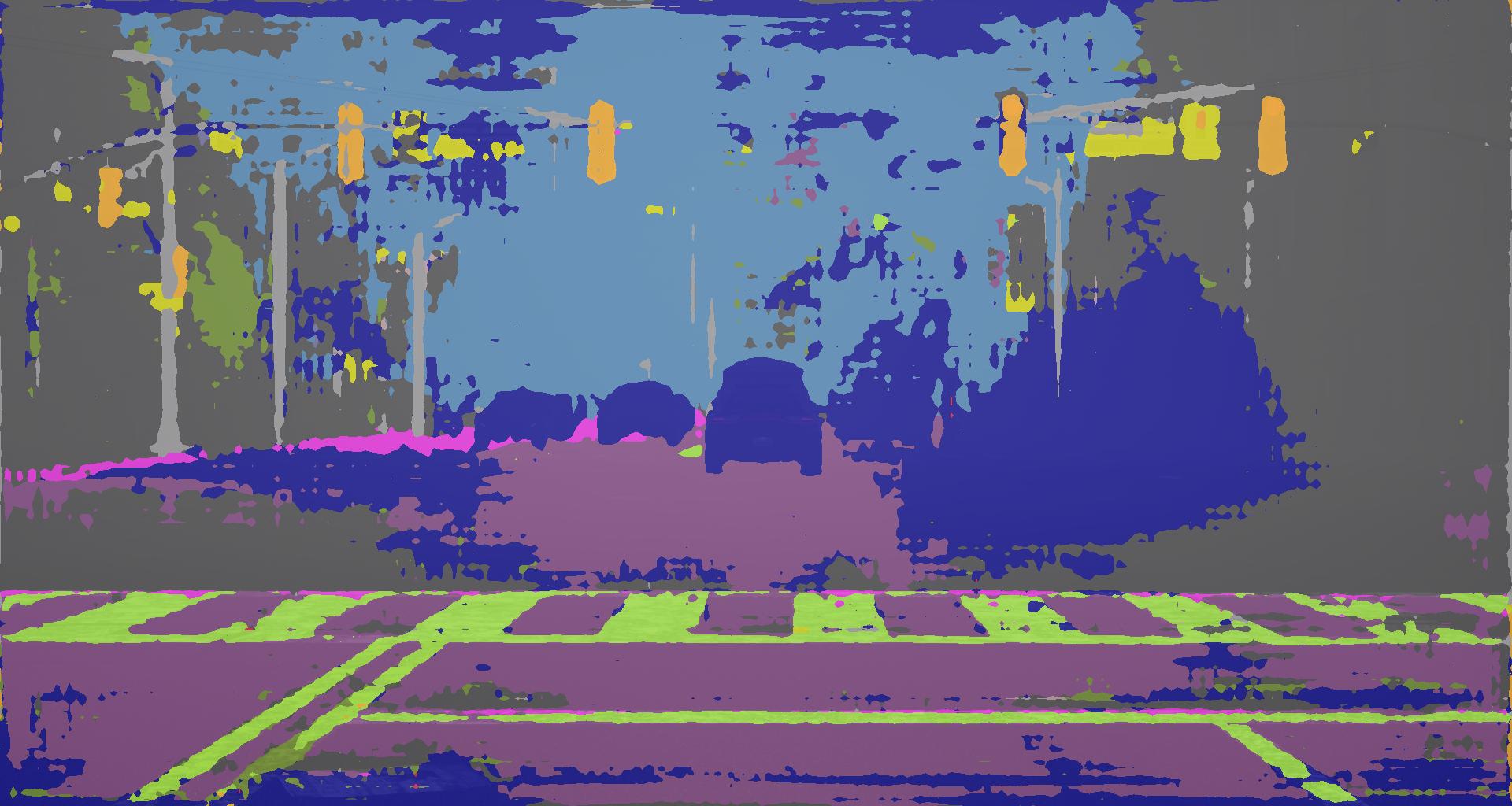}
    }
    \subfigure[M2A]{
        \includegraphics[width=0.3\linewidth]{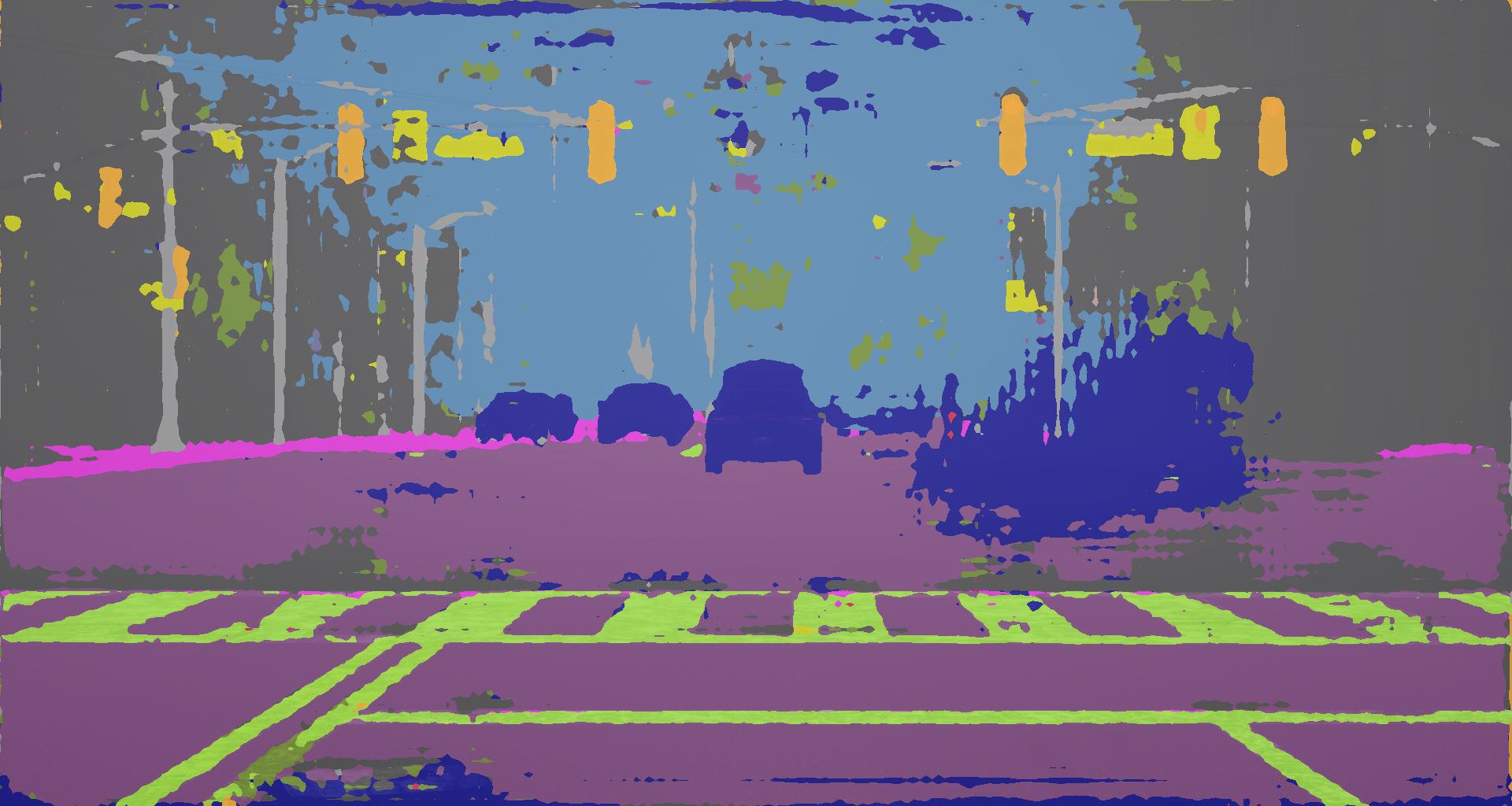}
    }
    \subfigure[SEGA]{
        \includegraphics[width=0.3\linewidth]{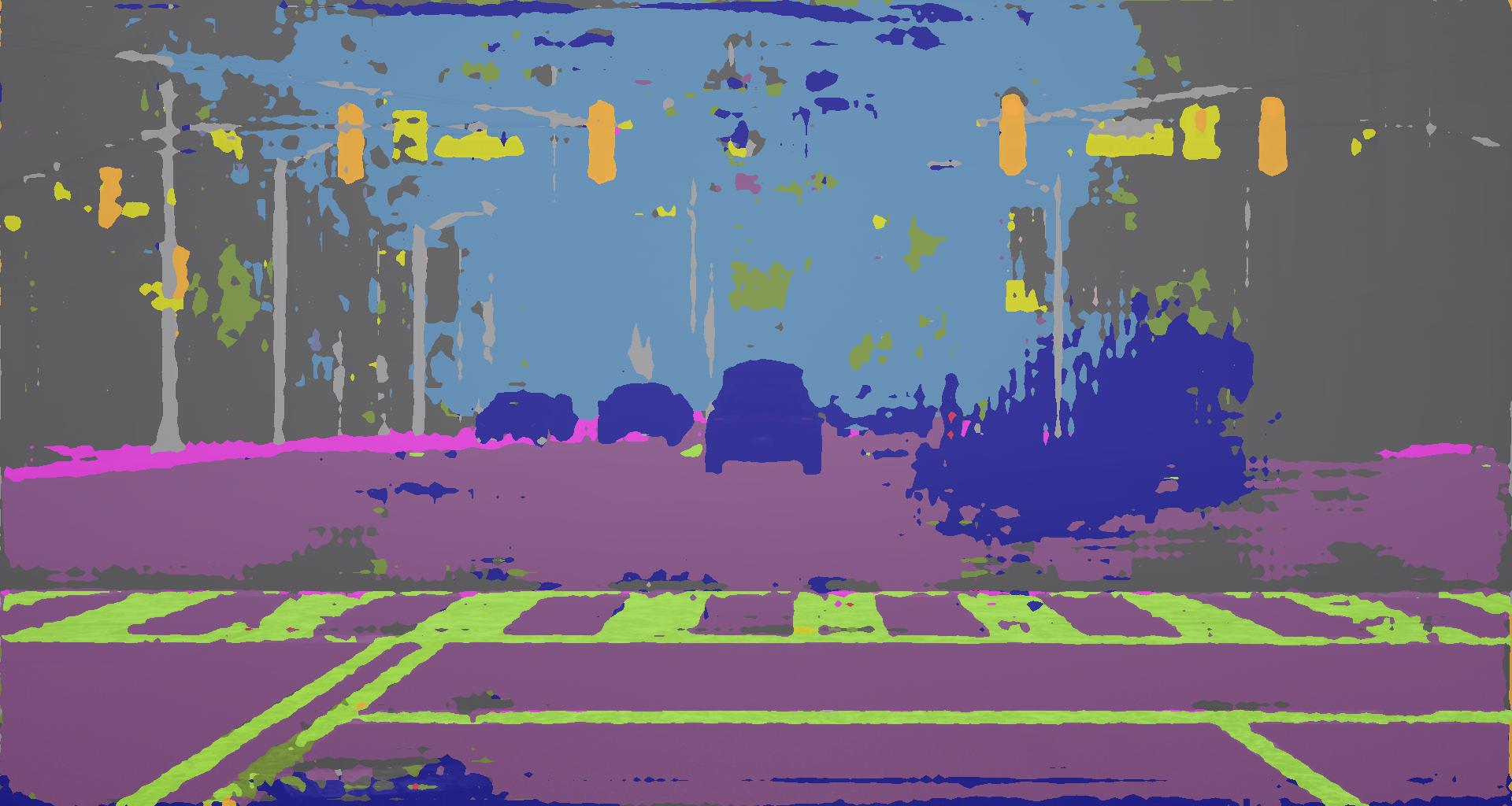}
    }

    \caption{Visualization of the segmentation results in the Carla dataset under CTTA. We use DeepLabV2 as the segmentation model.}
    \label{fig:carla_visuals}
\end{figure*}

\begin{table*}[h!]
    \centering
    \small
    \caption{Hyperparameter tuning of learning rates using Adam optimizer. Entries report semantic segmentation mIoU (\%) results on Carla dataset under CTTA. We use DeepLabV2 as the segmentation model.}
    \label{tab:carla_table}
    \setlength\tabcolsep{6pt}
    \resizebox{0.9\linewidth}{!}{
    \begin{tabular}{l|l|cccccccccccccc|c}
        \toprule
        Method & LR &
        \rotatebox[origin=c]{65}{Road} & \rotatebox[origin=c]{65}{Sidewalk} & \rotatebox[origin=c]{65}{Building} & \rotatebox[origin=c]{65}{Wall} & \rotatebox[origin=c]{65}{Fence} & \rotatebox[origin=c]{65}{Pole} & \rotatebox[origin=c]{65}{TrafficLight} & \rotatebox[origin=c]{65}{TrafficSign} & \rotatebox[origin=c]{65}{Vegetation} & \rotatebox[origin=c]{65}{Terrain}  & \rotatebox[origin=c]{65}{Sky} & \rotatebox[origin=c]{65}{Person} & \rotatebox[origin=c]{65}{Vehicle} & \rotatebox[origin=c]{65}{Roadline} & Mean$\uparrow$\\\midrule

        \multirow{3}{*}{\rotatebox[origin=c]{45}{SAR}}
        & $10^{-3}$ & 81.6 & 76.9 & 72.2 & 44.8 & 18.7 & 44.0 & 65.3 & 56.0 & 57.5 & 23.8 & 41.8 & 67.2 & 57.6 & 69.7 & 55.5\\
        & $10^{-4}$ & 86.3 & 77.2 & 71.7 & 43.9 & 19.9 & 44.9 & 66.6 & 57.1 & 58.6 & 23.9 & 39.0 & 67.9 & 62.4 & 71.6 & 56.5\\
        & $10^{-5}$ & 87.9 & 77.0 & 70.6 & 44.3 & 20.6 & 45.5 & 66.7 & 57.2 & 59.0 & 24.7 & 36.9 & 68.1 & 65.7 & 71.9 & 56.9\\
        \midrule

        \multirow{3}{*}{\rotatebox[origin=c]{45}{RDUMB}}
        & $10^{-3}$ & 86.0 & 77.1 & 71.9 & 43.4 & 18.1 & 44.6 & 66.0 & 56.7 & 58.1 & 23.2 & 39.0 & 67.7 & 65.1 & 71.4 & 56.3\\
        & $10^{-4}$ & 86.5 & 77.2 & 71.5 & 43.8 & 19.9 & 45.0 & 66.6 & 57.1 & 58.7 & 23.8 & 38.6 & 67.9 & 62.7 & 71.7 & 56.5\\
        & $10^{-5}$ & 86.5 & 77.2 & 71.4 & 43.8 & 20.1 & 45.0 & 66.6 & 57.1 & 58.7 & 23.9 & 38.6 & 67.9 & 62.5 & 71.7 & 56.5\\
        \midrule

        \multirow{3}{*}{\rotatebox[origin=c]{45}{M2A}}
        & $10^{-3}$ & 63.3 & 48.0 & 68.8 & 39.3 & 8.2 & 21.6 & 52.5 & 38.5 & 50.5 & 11.8 & 29.5 & 56.8 & 47.0 & 58.6 & 42.5\\
        & $10^{-4}$ & 91.8 & 78.9 & 76.7 & 46.7 & 17.2 & 44.6 & 66.6 & 56.5 & 58.4 & 22.9 & 42.8 & 67.8 & 80.0 & 74.2 & 58.9\\
        & $10^{-5}$ & 89.2 & 77.4 & 71.3 & 44.7 & 20.6 & 45.7 & 67.0 & 57.6 & 59.3 & 24.3 & 36.5 & 68.4 & 70.2 & 72.6 & 57.5\\
        \midrule

        \multirow{3}{*}{\rotatebox[origin=c]{45}{SEGA}}
        & $10^{-3}$ & 63.5 & 52.2 & 70.4 & 38.1 & 9.1 & 22.6 & 51.5 & 39.7 & 48.5 & 12.0 & 29.1 & 56.8 & 48.0 & 61.1 & 43.0\\
        & $10^{-4}$ & 91.8 & 78.8 & 76.3 & 46.6 & 17.4 & 44.8 & 66.6 & 56.6 & 58.5 & 23.0 & 42.3 & 67.8 & 80.0 & 74.1 & 58.9\\
        & $10^{-5}$ & 89.2 & 77.4 & 71.3 & 44.6 & 20.6 & 45.7 & 67.0 & 57.6 & 59.3 & 24.3 & 36.5 & 68.4 & 70.0 & 72.5 & 57.5\\
        \bottomrule
    \end{tabular}%
    }
\end{table*}
\begin{table*}[h!]
    \centering
    \small
    \caption{Hyperparameter tuning of learning rates using SGD optimizer. Entries report semantic segmentation mIoU (\%) results on Carla dataset under CTTA. We use DeepLabV2 as the segmentation model.}
    \label{tab:carla_table}
    \setlength\tabcolsep{6pt}
    \resizebox{0.9\linewidth}{!}{
    \begin{tabular}{l|l|cccccccccccccc|c}
        \toprule
        Method & LR &
        \rotatebox[origin=c]{65}{Road} & \rotatebox[origin=c]{65}{Sidewalk} & \rotatebox[origin=c]{65}{Building} & \rotatebox[origin=c]{65}{Wall} & \rotatebox[origin=c]{65}{Fence} & \rotatebox[origin=c]{65}{Pole} & \rotatebox[origin=c]{65}{TrafficLight} & \rotatebox[origin=c]{65}{TrafficSign} & \rotatebox[origin=c]{65}{Vegetation} & \rotatebox[origin=c]{65}{Terrain}  & \rotatebox[origin=c]{65}{Sky} & \rotatebox[origin=c]{65}{Person} & \rotatebox[origin=c]{65}{Vehicle} & \rotatebox[origin=c]{65}{Roadline} & Mean$\uparrow$\\\midrule

        \multirow{3}{*}{\rotatebox[origin=c]{45}{SAR}}
        & $10^{-3}$ & 81.6 & 76.9 & 72.2 & 44.8 & 18.7 & 44.0 & 65.3 & 56.0 & 57.5 & 23.8 & 41.8 & 67.2 & 57.6 & 69.7 & 55.5\\
        & $10^{-4}$ & 86.3 & 77.2 & 71.7 & 43.9 & 19.9 & 44.9 & 66.6 & 57.1 & 58.6 & 23.9 & 39.0 & 67.9 & 62.4 & 71.6 & 56.5\\
        & $10^{-5}$ & 87.9 & 77.0 & 70.6 & 44.3 & 20.6 & 45.5 & 66.7 & 57.2 & 59.0 & 24.7 & 36.9 & 68.1 & 65.7 & 71.9 & 56.9\\
        \midrule

        \multirow{3}{*}{\rotatebox[origin=c]{45}{RDUMB}}
        & $10^{-3}$ & 86.5 & 77.2 & 71.4 & 43.8 & 20.1 & 45.0 & 66.6 & 57.1 & 58.7 & 23.9 & 38.6 & 67.9 & 62.5 & 71.7 & 56.5\\
        & $10^{-4}$ & 86.5 & 77.2 & 71.4 & 43.8 & 20.1 & 45.0 & 66.6 & 57.1 & 58.7 & 23.9 & 38.6 & 67.9 & 62.5 & 71.7 & 56.5\\
        & $10^{-5}$ & 86.5 & 77.2 & 71.4 & 43.8 & 20.1 & 45.0 & 66.6 & 57.1 & 58.7 & 23.9 & 38.6 & 67.9 & 62.5 & 71.7 & 56.5\\
        \midrule

        \multirow{3}{*}{\rotatebox[origin=c]{45}{M2A}}
        & $10^{-3}$ & 93.0 & 78.5 & 74.5 & 46.6 & 19.3 & 45.3 & 67.1 & 57.5 & 60.3 & 24.0 & 38.6 & 68.3 & 81.7 & 73.9 & 59.2\\
        & $10^{-4}$ & 90.1 & 77.4 & 71.6 & 45.6 & 19.6 & 45.7 & 66.7 & 57.6 & 59.0 & 24.3 & 36.0 & 68.5 & 76.7 & 72.6 & 58.0\\
        & $10^{-5}$ & 87.1 & 77.2 & 71.5 & 44.1 & 20.2 & 45.2 & 66.8 & 57.2 & 58.8 & 24.0 & 38.3 & 68.1 & 63.9 & 71.9 & 56.7\\
        \midrule

        \multirow{3}{*}{\rotatebox[origin=c]{45}{SEGA}}
        & $10^{-3}$ & 93.0 & 78.5 & 74.5 & 46.6 & 19.3 & 45.3 & 67.1 & 57.5 & 60.3 & 24.1 & 38.6 & 68.3 & 81.6 & 73.9 & 59.2\\
        & $10^{-4}$ & 90.1 & 77.4 & 71.6 & 45.6 & 19.6 & 45.7 & 66.7 & 57.6 & 59.0 & 24.3 & 36.0 & 68.5 & 76.6 & 72.6 & 58.0\\
        & $10^{-5}$ & 87.1 & 77.3 & 71.5 & 44.1 & 20.2 & 45.2 & 66.8 & 57.3 & 58.8 & 24.0 & 38.3 & 68.1 & 63.9 & 71.9 & 56.7\\
        \bottomrule
    \end{tabular}%
    }
\end{table*}
We extend our evaluation to semantic segmentation on the Carla driving dataset under CTTA. The main Carla table reports per-class and mean mIoU for DeepLabV2 across road, infrastructure, object, and agent categories, comparing SOURCE, SAR, RDumb, M2A, and SEGA to show that sensitivity-guided adaptation maintains or improves segmentation quality across most classes while preserving strong mean performance. The accompanying learning-rate tuning tables sweep three learning rates for each method under SGD and Adam optimizers, revealing that SEGA closely tracks M2A at the best settings and that our chosen default learning rate lies near a stable performance plateau rather than a fragile optimum. Finally, Fig.~\ref{fig:carla_visuals} visualizes qualitative segmentation outputs for representative scenes, highlighting how SEGA reduces artifacts and preserves structure relative to entropy-only baselines, especially on road markings, vehicles, and thin objects.

\subsection{ImageNet-C Learning-Rate Tuning}
\label{sec:appendix-imagenetc-lr}
\begin{table*}[h!]
    \centering
    \small
    \caption{ImageNet-C hyperparameter tuning of learning rates with Adam optimizer. Entries report classification error rate (\%) under CTTA (batch size 1) using ViT-Base. Mean is the average across 15 corruption domains, each with 5000 test samples.}
    \label{tab:imagenetc_lrtune_adam}
    \setlength\tabcolsep{6pt}
    \resizebox{0.9\linewidth}{!}{
    \begin{tabular}{l|l|ccc|cccc|cccc|cccc|c}
        \toprule
        Method & LR &
        \rotatebox[origin=c]{0}{GN} & \rotatebox[origin=c]{0}{SN} & \rotatebox[origin=c]{0}{IN} & \rotatebox[origin=c]{0}{DB} & \rotatebox[origin=c]{0}{GB} & \rotatebox[origin=c]{0}{MB} & \rotatebox[origin=c]{0}{ZB} & \rotatebox[origin=c]{0}{S} & \rotatebox[origin=c]{0}{Fr} & \rotatebox[origin=c]{0}{F}  & \rotatebox[origin=c]{0}{B} & \rotatebox[origin=c]{0}{C} & \rotatebox[origin=c]{0}{ET} & \rotatebox[origin=c]{0}{P} & \rotatebox[origin=c]{0}{JC}
        & Mean$\downarrow$\\\midrule

        \multirow{3}{*}{\rotatebox[origin=c]{45}{ROTTA}}
        & $10^{-3}$ & 49.9 & 48.2 & 48.0 & 67.9 & 70.5 & 55.1 & 60.0 & 45.5 & 48.7 & 53.4 & 24.9 & 88.4 & 54.0 & 37.2 & 34.6 & 52.4 \\
        & $10^{-4}$ & 49.9 & 48.8 & 48.8 & 68.3 & 72.0 & 55.7 & 60.6 & 47.1 & 51.0 & 56.6 & 24.5 & 91.1 & 55.4 & 38.9 & 36.0 & 53.6 \\
        & $10^{-5}$ & 49.9 & 48.9 & 48.9 & 68.6 & 72.3 & 55.9 & 60.5 & 47.5 & 52.1 & 56.0 & 24.8 & 91.1 & 55.7 & 39.0 & 36.6 & 53.9 \\
        \midrule

        \multirow{3}{*}{\rotatebox[origin=c]{45}{RPL}}
        & $10^{-3}$ & 93.4 & 99.9 & 99.9 & 99.9 & 99.9 & 99.9 & 99.9 & 99.9 & 99.9 & 99.9 & 99.9 & 99.9 & 99.9 & 99.9 & 99.9 & 99.5 \\
        & $10^{-4}$ & 43.5 & 37.9 & 39.0 & 50.9 & 52.6 & 47.1 & 57.3 & 84.3 & 99.8 & 99.9 & 99.6 & 99.9 & 99.9 & 99.9 & 99.9 & 74.1 \\
        & $10^{-5}$ & 47.8 & 43.5 & 42.7 & 58.7 & 60.5 & 46.2 & 51.4 & 40.6 & 42.0 & 43.0 & 21.8 & 56.4 & 49.9 & 31.7 & 31.1 & 44.5 \\
        \midrule


        \multirow{3}{*}{\rotatebox[origin=c]{45}{SANTA}}
        & $10^{-3}$ & 46.0 & 44.4 & 45.8 & 63.2 & 66.3 & 49.2 & 55.4 & 41.7 & 47.7 & 46.6 & 24.2 & 92.6 & 49.1 & 35.2 & 35.1 & 49.5 \\
        & $10^{-4}$ & 45.6 & 43.2 & 45.1 & 60.0 & 66.4 & 48.5 & 55.1 & 41.1 & 47.4 & 45.3 & 22.6 & 90.6 & 50.6 & 35.8 & 33.9 & 48.7 \\
        & $10^{-5}$ & 47.8 & 44.0 & 44.6 & 61.4 & 64.8 & 49.4 & 54.2 & 41.5 & 45.7 & 45.9 & 22.7 & 84.6 & 49.7 & 35.5 & 34.0 & 48.4 \\
        \midrule

        \multirow{3}{*}{\rotatebox[origin=c]{45}{TENT}}
        & $10^{-3}$ & 85.4 & 99.9 & 99.9 & 99.9 & 99.9 & 99.9 & 99.9 & 99.9 & 99.9 & 99.9 & 99.9 & 99.9 & 99.9 & 99.9 & 99.9 & 99.0 \\
        & $10^{-4}$ & 43.0 & 38.3 & 39.3 & 50.7 & 52.6 & 50.0 & 95.2 & 99.7 & 99.9 & 99.9 & 99.9 & 99.9 & 99.9 & 99.9 & 99.9 & 77.9 \\
        & $10^{-5}$ & 47.2 & 41.9 & 41.7 & 57.1 & 58.2 & 45.6 & 51.0 & 40.5 & 41.9 & 42.1 & 22.6 & 54.0 & 49.9 & 31.8 & 30.9 & 43.8 \\
        \midrule

        \multirow{3}{*}{\rotatebox[origin=c]{45}{LCOTTA}}
        & $10^{-3}$ & 50.7 & 49.6 & 55.5 & 82.3 & 70.2 & 65.6 & 66.8 & 63.0 & 56.5 & 73.5 & 35.4 & 99.3 & 57.6 & 48.0 & 45.2 & 61.3 \\
        & $10^{-4}$ & 41.8 & 37.5 & 38.9 & 54.7 & 54.9 & 48.5 & 52.1 & 40.3 & 39.1 & 41.1 & 23.1 & 57.7 & 47.7 & 32.1 & 31.0 & 42.7 \\
        & $10^{-5}$ & 47.1 & 42.2 & 41.9 & 60.8 & 61.3 & 45.9 & 51.4 & 39.5 & 41.1 & 43.0 & 21.9 & 58.8 & 47.9 & 32.2 & 30.5 & 44.4 \\
        \midrule

        \multirow{3}{*}{\rotatebox[origin=c]{45}{RES.TTA}}
        & $10^{-3}$ & 42.0 & 39.9 & 39.7 & 48.4 & 46.9 & 44.3 & 45.9 & 46.1 & 42.6 & 37.2 & 22.6 & 83.2 & 34.3 & 29.2 & 29.9 & 42.1 \\
        & $10^{-4}$ & 42.0 & 39.9 & 39.7 & 48.4 & 46.9 & 44.3 & 45.9 & 46.1 & 42.6 & 37.2 & 22.6 & 83.2 & 34.8 & 29.2 & 29.8 & 42.2 \\
        & $10^{-5}$ & 42.0 & 39.9 & 39.7 & 48.4 & 46.9 & 44.3 & 45.9 & 46.1 & 42.6 & 37.2 & 22.6 & 83.2 & 34.3 & 28.7 & 29.5 & 42.1 \\
        \midrule

        \multirow{3}{*}{\rotatebox[origin=c]{45}{SAR}}
        & $10^{-3}$ & 42.0 & 38.3 & 40.0 & 56.6 & 46.5 & 40.5 & 41.8 & 40.7 & 35.3 & 77.6 & 27.8 & 99.6 & 33.5 & 27.5 & 29.0 & 45.1 \\
        & $10^{-4}$ & 42.0 & 38.3 & 40.0 & 56.6 & 46.5 & 40.5 & 41.8 & 40.7 & 35.3 & 77.6 & 27.8 & 99.6 & 33.5 & 27.5 & 29.0 & 45.1 \\
        & $10^{-5}$ & 42.0 & 38.3 & 40.0 & 56.6 & 46.5 & 40.5 & 41.8 & 40.7 & 35.3 & 77.6 & 27.8 & 99.6 & 33.5 & 27.5 & 29.0 & 45.1 \\
        \midrule

        \multirow{3}{*}{\rotatebox[origin=c]{45}{RDUMB}}
        & $10^{-3}$ & 43.3 & 42.6 & 42.1 & 49.2 & 47.3 & 43.3 & 45.7 & 35.0 & 37.3 & 36.5 & 22.2 & 67.1 & 37.1 & 28.3 & 29.8 & 40.5 \\
        & $10^{-4}$ & 46.4 & 43.9 & 45.5 & 59.8 & 63.7 & 48.6 & 53.3 & 40.3 & 46.1 & 49.6 & 22.1 & 90.8 & 49.0 & 33.9 & 32.7 & 48.4 \\
        & $10^{-5}$ & 49.3 & 48.2 & 48.4 & 67.6 & 71.6 & 55.0 & 59.6 & 46.5 & 51.7 & 55.1 & 24.6 & 91.1 & 55.0 & 38.4 & 36.1 & 53.2 \\
        \midrule

        \multirow{3}{*}{\rotatebox[origin=c]{45}{M2A}}
        & $10^{-3}$ & 40.2 & 37.2 & 37.6 & 48.3 & 40.4 & 39.3 & 38.1 & 33.3 & 35.1 & 31.5 & 22.2 & 99.8 & 99.6 & 99.9 & 99.9 & 53.5 \\
        & $10^{-4}$ & 41.4 & 35.8 & 36.6 & 48.9 & 45.0 & 39.0 & 41.5 & 34.5 & 34.2 & 32.6 & 21.1 & 43.3 & 38.6 & 27.0 & 28.6 & 36.6 \\
        & $10^{-5}$ & 47.4 & 41.5 & 41.4 & 60.4 & 57.5 & 45.7 & 50.1 & 39.5 & 41.1 & 85.7 & 23.9 & 98.1 & 52.8 & 42.6 & 38.8 & 51.1 \\
        \midrule

        \multirow{3}{*}{\rotatebox[origin=c]{45}{SEGA}}
        & $10^{-3}$ & 40.4 & 37.0 & 37.6 & 47.9 & 41.4 & 39.9 & 40.3 & 31.9 & 33.3 & 32.9 & 21.2 & 47.1 & 32.3 & 27.8 & 28.5 & 36.0 \\
        & $10^{-4}$ & 41.7 & 35.6 & 36.7 & 49.2 & 44.9 & 39.2 & 41.5 & 34.3 & 36.4 & 32.2 & 21.8 & 85.8 & 39.7 & 30.6 & 31.0 & 40.1 \\
        & $10^{-5}$ & 47.5 & 41.6 & 41.5 & 60.4 & 57.6 & 46.0 & 50.3 & 39.6 & 40.9 & 83.0 & 23.2 & 89.5 & 51.5 & 34.3 & 32.9 & 49.3 \\

        \bottomrule
    \end{tabular}%
    }

\end{table*}

\begin{table*}[h!]
    \centering
    \small
    \caption{ImageNet-C hyperparameter tuning of learning rates with SGD optimizer. Entries report classification error rate (\%) under CTTA (batch size 1) using ViT-Base. Mean is the average across 15 corruption domains, each with 5000 test samples.}
    \label{tab:imagenetc_lrtune_sgd}
    \setlength\tabcolsep{6pt}
    \resizebox{0.9\linewidth}{!}{
    \begin{tabular}{l|l|ccc|cccc|cccc|cccc|c}
        \toprule
        Method & LR &
        \rotatebox[origin=c]{0}{GN} & \rotatebox[origin=c]{0}{SN} & \rotatebox[origin=c]{0}{IN} & \rotatebox[origin=c]{0}{DB} & \rotatebox[origin=c]{0}{GB} & \rotatebox[origin=c]{0}{MB} & \rotatebox[origin=c]{0}{ZB} & \rotatebox[origin=c]{0}{S} & \rotatebox[origin=c]{0}{Fr} & \rotatebox[origin=c]{0}{F}  & \rotatebox[origin=c]{0}{B} & \rotatebox[origin=c]{0}{C} & \rotatebox[origin=c]{0}{ET} & \rotatebox[origin=c]{0}{P} & \rotatebox[origin=c]{0}{JC}
        & Mean$\downarrow$\\\midrule

        \multirow{3}{*}{\rotatebox[origin=c]{45}{ROTTA}}
        & $10^{-3}$ & 49.8 & 48.6 & 48.4 & 68.2 & 71.1 & 55.6 & 61.3 & 46.0 & 49.1 & 57.5 & 24.7 & 90.7 & 54.9 & 38.8 & 35.6 & 53.4 \\
        & $10^{-4}$ & 49.9 & 48.9 & 48.8 & 68.4 & 72.0 & 55.6 & 61.0 & 47.0 & 51.4 & 56.6 & 24.6 & 91.3 & 55.8 & 39.1 & 36.0 & 53.8 \\
        & $10^{-5}$ & 49.9 & 48.9 & 48.9 & 68.5 & 72.4 & 55.9 & 60.5 & 47.6 & 52.2 & 56.0 & 24.8 & 91.2 & 55.8 & 39.2 & 36.7 & 53.9 \\
        \midrule

        \multirow{3}{*}{\rotatebox[origin=c]{45}{RPL}}
        & $10^{-3}$ & 60.5 & 99.8 & 99.9 & 99.8 & 99.9 & 99.9 & 99.9 & 99.9 & 99.9 & 99.9 & 99.9 & 99.9 & 99.9 & 99.9 & 99.9 & 97.3 \\
        & $10^{-4}$ & 46.4 & 40.7 & 41.2 & 55.3 & 56.1 & 44.3 & 50.6 & 40.5 & 42.3 & 41.2 & 23.5 & 56.0 & 47.0 & 32.5 & 31.6 & 43.3 \\
        & $10^{-5}$ & 49.1 & 46.7 & 46.6 & 65.7 & 66.7 & 50.8 & 55.2 & 42.0 & 46.1 & 47.8 & 22.5 & 68.2 & 52.8 & 34.1 & 34.2 & 48.6 \\
        \midrule


        \multirow{3}{*}{\rotatebox[origin=c]{45}{SANTA}}
        & $10^{-3}$ & 45.9 & 43.5 & 45.9 & 61.3 & 65.1 & 49.1 & 54.9 & 40.4 & 47.5 & 42.4 & 23.3 & 93.8 & 48.1 & 34.3 & 33.8 & 48.6 \\
        & $10^{-4}$ & 45.7 & 42.8 & 44.2 & 58.9 & 64.9 & 47.4 & 54.2 & 40.2 & 46.8 & 41.7 & 22.8 & 90.1 & 48.9 & 35.2 & 33.7 & 47.9 \\
        & $10^{-5}$ & 48.0 & 44.6 & 45.0 & 63.1 & 64.7 & 48.6 & 54.1 & 41.2 & 45.8 & 45.0 & 23.3 & 83.1 & 50.1 & 36.0 & 34.3 & 48.4 \\
        \midrule

        \multirow{3}{*}{\rotatebox[origin=c]{45}{TENT}}
        & $10^{-3}$ & 62.3 & 99.9 & 99.9 & 99.9 & 99.9 & 99.9 & 99.9 & 99.9 & 99.9 & 99.9 & 99.9 & 99.9 & 99.9 & 99.9 & 99.9 & 97.4 \\
        & $10^{-4}$ & 43.2 & 38.1 & 38.8 & 51.6 & 52.3 & 44.0 & 47.5 & 45.0 & 71.5 & 99.8 & 98.9 & 99.9 & 99.8 & 99.8 & 99.8 & 68.7 \\
        & $10^{-5}$ & 47.7 & 42.6 & 43.0 & 60.6 & 58.8 & 46.9 & 53.0 & 40.8 & 43.4 & 41.7 & 23.5 & 53.0 & 49.5 & 32.3 & 32.3 & 44.6 \\
        \midrule

        \multirow{3}{*}{\rotatebox[origin=c]{45}{LCOTTA}}
        & $10^{-3}$ & 99.2 & 99.8 & 99.8 & 99.8 & 99.8 & 99.8 & 99.8 & 99.9 & 99.9 & 99.9 & 99.7 & 99.9 & 99.9 & 99.8 & 99.9 & 99.8 \\
        & $10^{-4}$ & 41.6 & 38.0 & 39.2 & 63.6 & 54.0 & 50.3 & 53.8 & 42.4 & 41.7 & 48.3 & 24.2 & 99.9 & 46.1 & 32.0 & 32.7 & 47.2 \\
        & $10^{-5}$ & 45.6 & 39.8 & 41.6 & 62.6 & 62.8 & 47.9 & 53.1 & 39.9 & 42.0 & 41.4 & 23.6 & 59.3 & 48.8 & 31.6 & 31.2 & 44.8 \\
        \midrule

        \multirow{3}{*}{\rotatebox[origin=c]{45}{RES.TTA}}
        & $10^{-3}$ & 42.0 & 39.9 & 39.7 & 48.4 & 46.9 & 44.3 & 45.9 & 46.1 & 42.6 & 37.2 & 22.6 & 83.2 & 34.3 & 29.2 & 29.9 & 42.1 \\
        & $10^{-4}$ & 42.0 & 39.9 & 39.7 & 48.4 & 46.9 & 44.3 & 45.9 & 46.1 & 42.6 & 37.2 & 22.6 & 83.2 & 34.3 & 29.2 & 29.9 & 42.1 \\
        & $10^{-5}$ & 42.0 & 39.9 & 39.7 & 48.4 & 46.9 & 44.3 & 45.9 & 46.1 & 42.6 & 37.2 & 22.6 & 83.2 & 34.3 & 29.2 & 29.9 & 42.1 \\
        \midrule

        \multirow{3}{*}{\rotatebox[origin=c]{45}{SAR}}
        & $10^{-3}$ & 42.0 & 38.3 & 40.0 & 56.6 & 46.5 & 40.5 & 41.8 & 40.7 & 35.3 & 77.6 & 27.8 & 99.6 & 33.5 & 27.5 & 29.0 & 45.1 \\
        & $10^{-4}$ & 42.0 & 38.3 & 40.0 & 56.6 & 46.5 & 40.5 & 41.8 & 40.7 & 35.3 & 77.6 & 27.8 & 99.6 & 33.5 & 27.5 & 29.0 & 45.1 \\
        & $10^{-5}$ & 42.0 & 38.3 & 40.0 & 56.6 & 46.5 & 40.5 & 41.8 & 40.7 & 35.3 & 77.6 & 27.8 & 99.6 & 33.5 & 27.5 & 29.0 & 45.1 \\
        \midrule

        \multirow{3}{*}{\rotatebox[origin=c]{45}{RDUMB}}
        & $10^{-3}$ & 77.6 & 64.2 & 66.9 & 92.9 & 79.4 & 79.8 & 75.5 & 45.5 & 66.4 & 94.6 & 38.6 & 97.4 & 45.7 & 43.9 & 37.5 & 67.1 \\
        & $10^{-4}$ & 44.8 & 42.9 & 43.3 & 54.1 & 55.8 & 45.9 & 49.6 & 38.4 & 42.6 & 43.3 & 22.8 & 90.2 & 43.3 & 31.0 & 31.9 & 45.3 \\
        & $10^{-5}$ & 48.4 & 47.0 & 47.4 & 65.3 & 69.6 & 52.7 & 57.3 & 44.4 & 49.4 & 53.5 & 23.9 & 91.0 & 53.6 & 37.1 & 35.3 & 51.7 \\
        \midrule

        \multirow{3}{*}{\rotatebox[origin=c]{45}{M2A}}
        & $10^{-3}$ & 42.9 & 40.7 & 40.0 & 80.0 & 99.9 & 99.9 & 99.9 & 99.9 & 99.9 & 99.9 & 99.9 & 99.9 & 99.9 & 99.9 & 99.9 & 86.8 \\
        & $10^{-4}$ & 41.0 & 36.0 & 36.6 & 48.8 & 43.6 & 38.7 & 40.2 & 34.0 & 33.5 & 33.0 & 21.1 & 48.6 & 35.6 & 27.4 & 28.6 & 36.5 \\
        & $10^{-5}$ & 47.1 & 40.8 & 40.6 & 58.5 & 54.7 & 44.0 & 48.6 & 38.9 & 40.4 & 47.1 & 22.4 & 48.1 & 44.3 & 31.4 & 31.8 & 42.6 \\
        \midrule

        \multirow{3}{*}{\rotatebox[origin=c]{45}{SEGA}}
        & $10^{-3}$ & 42.3 & 38.3 & 39.8 & 55.6 & 43.5 & 43.9 & 42.1 & 35.7 & 36.9 & 94.4 & 22.5 & 99.6 & 32.7 & 29.2 & 29.2 & 45.7 \\
        & $10^{-4}$ & 41.2 & 36.2 & 36.5 & 50.1 & 43.5 & 39.1 & 44.6 & 34.3 & 35.1 & 63.5 & 22.2 & 87.1 & 37.3 & 30.5 & 31.4 & 42.2 \\
        & $10^{-5}$ & 47.4 & 40.9 & 40.7 & 58.6 & 55.1 & 44.5 & 49.2 & 39.1 & 40.9 & 68.4 & 23.3 & 89.0 & 50.0 & 33.5 & 34.8 & 47.7 \\

        \bottomrule
    \end{tabular}%
    }

\end{table*}

We report complementary learning-rate sweeps for ImageNet-C under CTTA, using both Adam and SGD optimizers with ViT-Base. Table~\ref{tab:imagenetc_lrtune_adam} varies the learning rate over three orders of magnitude for each method with Adam, showing that SEGA and M2A achieve similar best performance at moderate learning rates while very large or very small values either destabilize certain baselines or leave substantial robustness untapped. Table~\ref{tab:imagenetc_lrtune_sgd} provides the analogous sweep for SGD, illustrating that the default learning rate used in the main experiments lies in a broad, stable region rather than at a narrow optimum, and that SEGA's relative gains do not depend on finely tuned optimization hyperparameters.

\section{Extended Analysis}
\label{sec:extended-analysis}

\subsection{Extended Ablation Studies}
\label{sec:appendix-extended-ablation}
\begin{figure*}[t!]
    \centering
    \tiny
    \begin{tabular}{cc}
        \textbf{CIFAR10-C} & \textbf{ImageNet-C} \\
        \subfigure[Deterioration margin $D$]{%
            \includegraphics[width=0.35\textwidth]{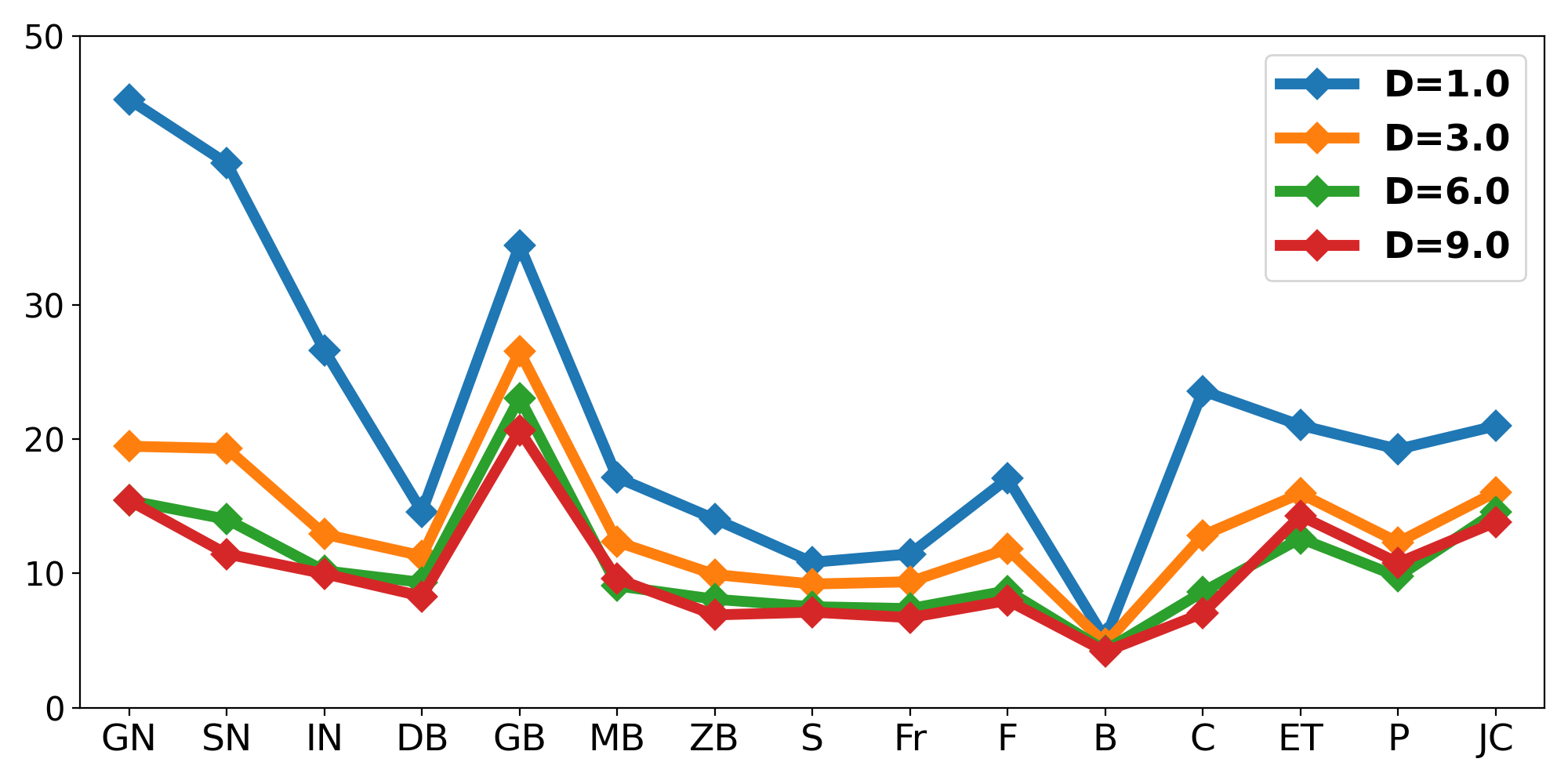}%
        }
        &
        \subfigure[Deterioration margin $D$]{%
            \includegraphics[width=0.35\textwidth]{figures/ablations/imagenet_c/ablation_d.png}%
        }\\[0.5ex]
        \subfigure[Quantile level $Q$]{%
            \includegraphics[width=0.35\textwidth]{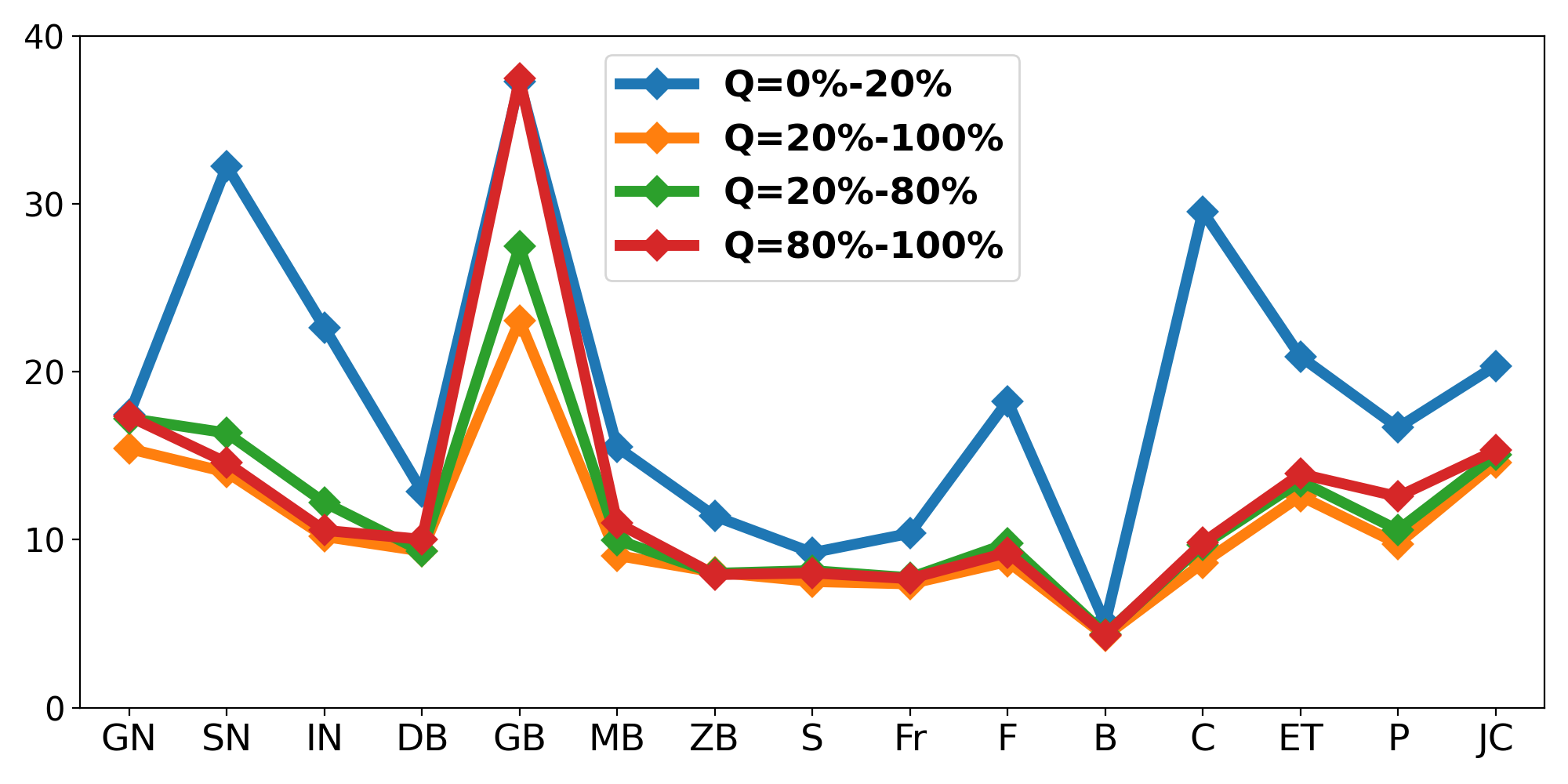}%
        }
        &
        \subfigure[Quantile level $Q$]{%
            \includegraphics[width=0.35\textwidth]{figures/ablations/imagenet_c/ablation_q.png}%
        }\\[0.5ex]
        \subfigure[Minimum activation steps $T$]{%
            \includegraphics[width=0.35\textwidth]{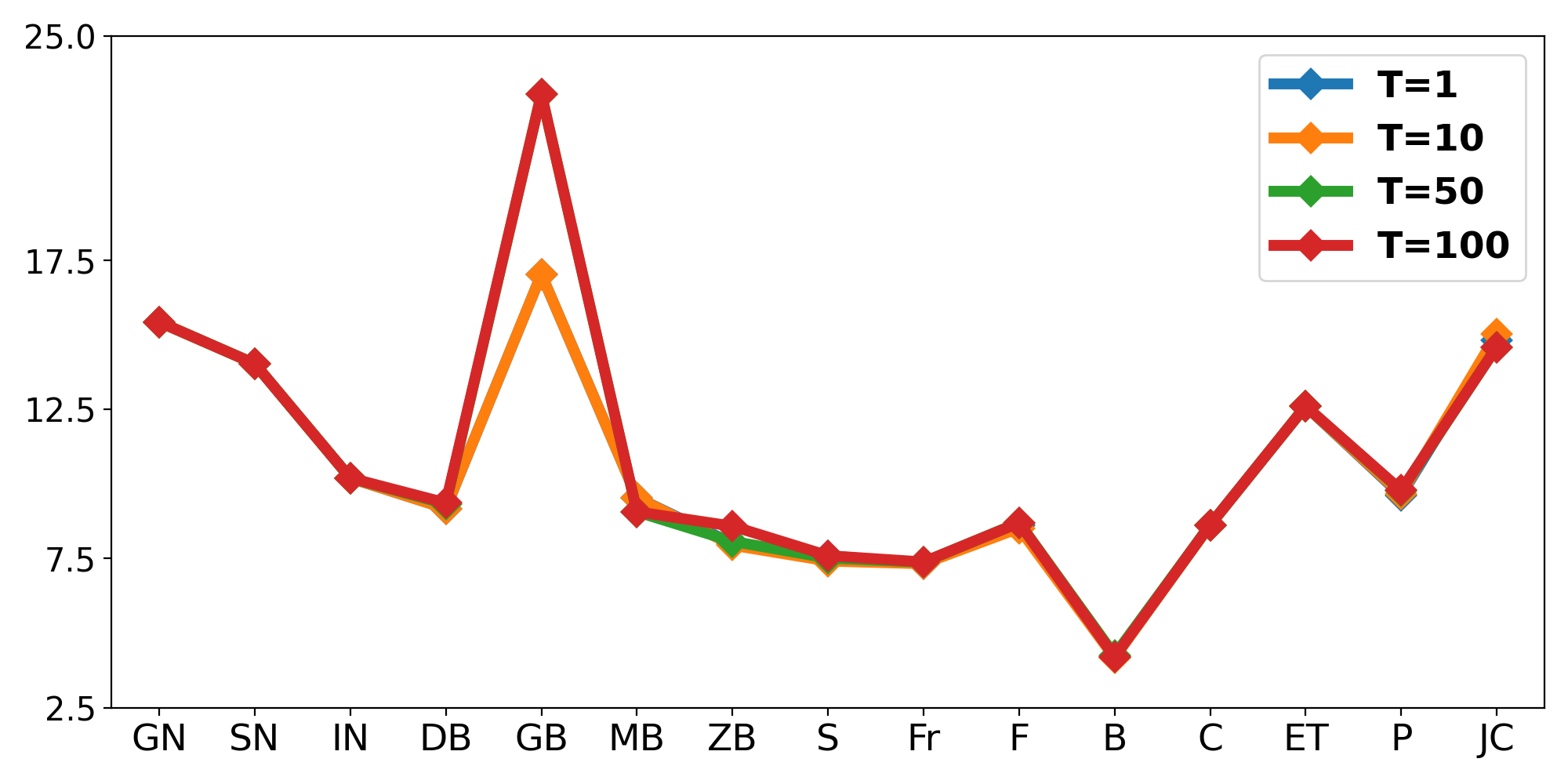}%
        }
        &
        \subfigure[Minimum activation steps $T$]{%
            \includegraphics[width=0.35\textwidth]{figures/ablations/imagenet_c/ablation_t.png}%
        }\\[0.5ex]
        \subfigure[Warm-up steps $U$]{%
            \includegraphics[width=0.35\textwidth]{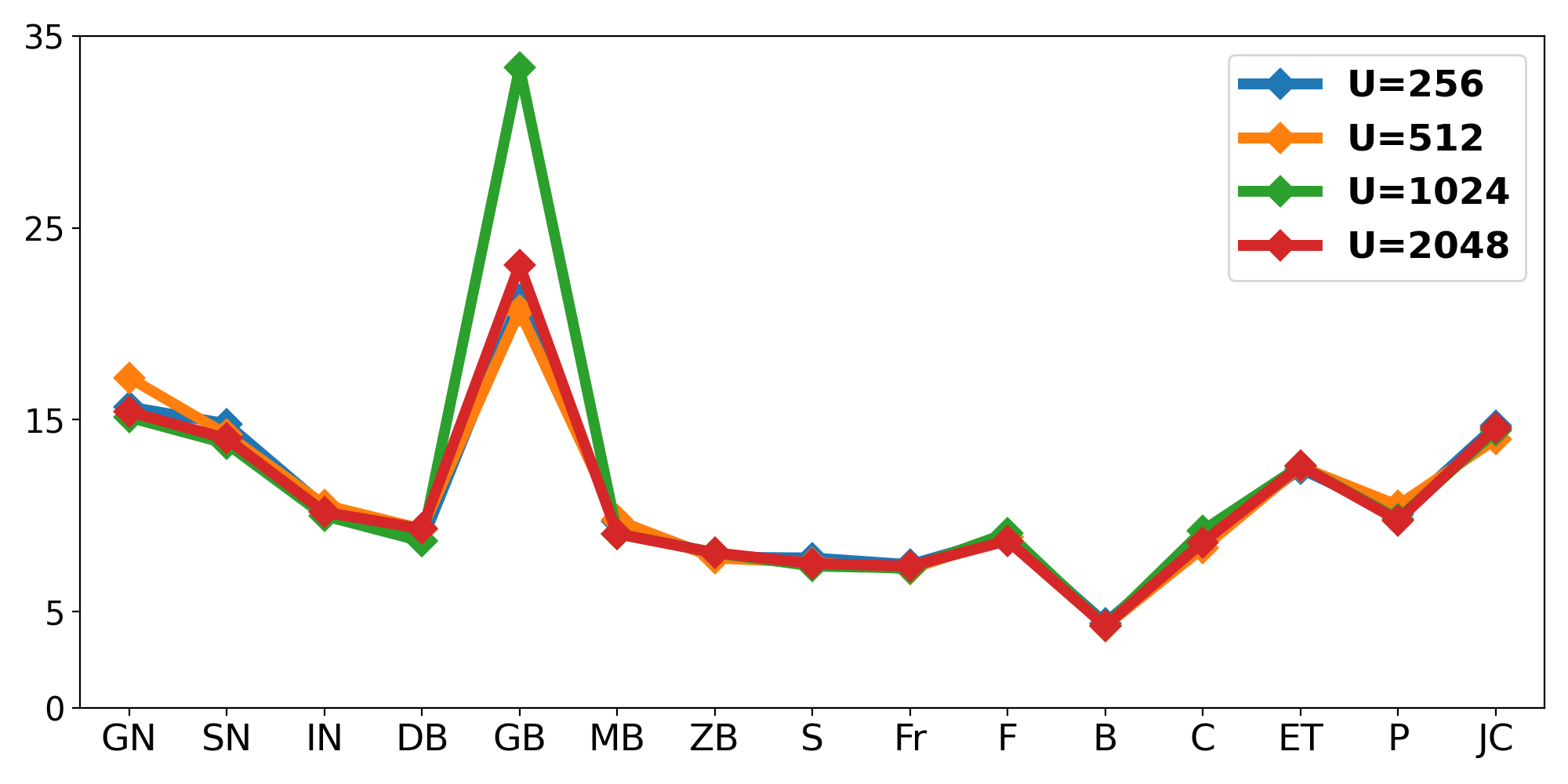}%
        }
        &
        \subfigure[Warm-up steps $U$]{%
            \includegraphics[width=0.35\textwidth]{figures/ablations/imagenet_c/ablation_u.png}%
        }\\[0.5ex]
        \subfigure[EMA weight $\zeta$]{%
            \includegraphics[width=0.35\textwidth]{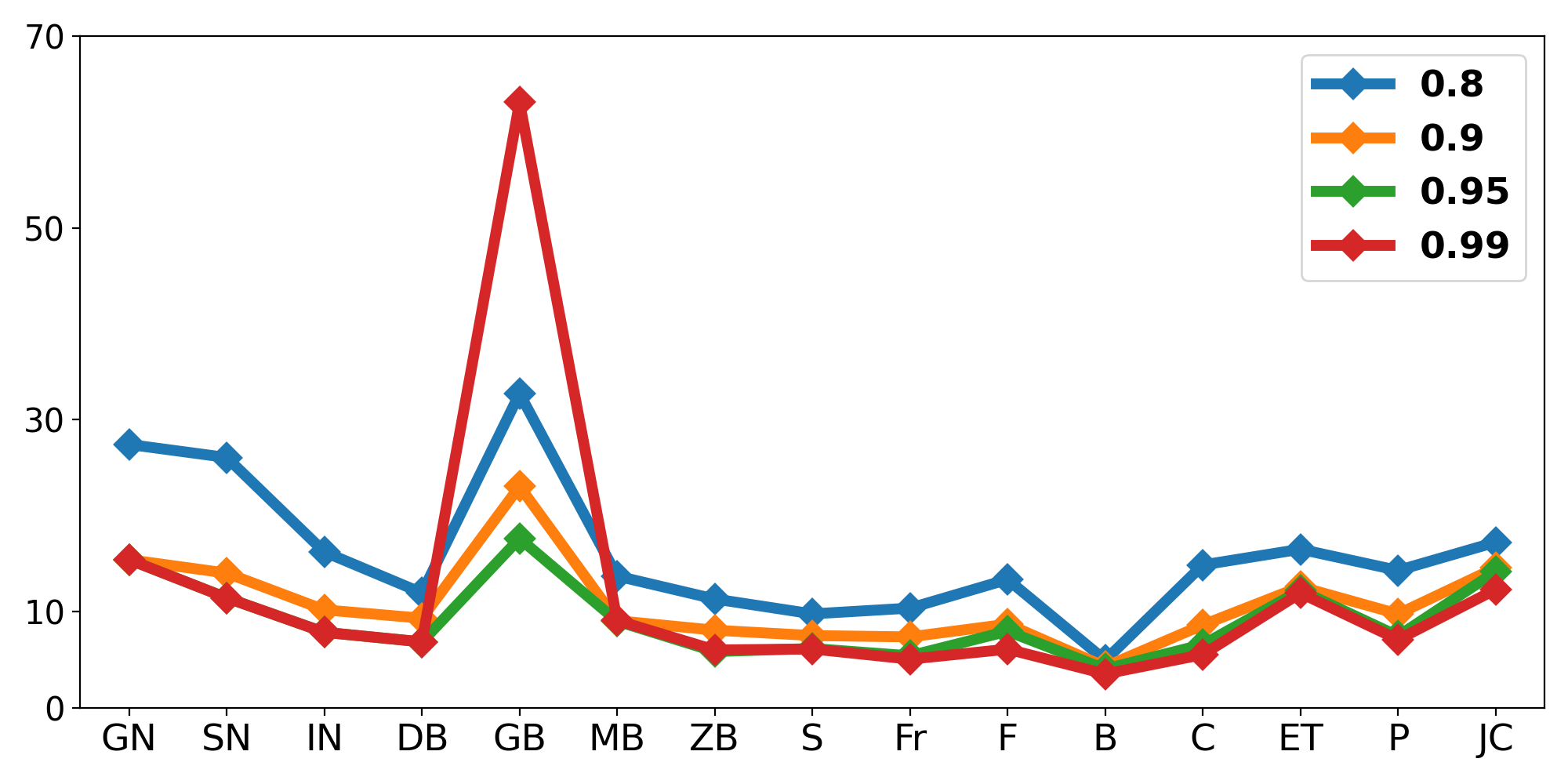}%
        }
        &
        \subfigure[EMA weight $\zeta$]{%
            \includegraphics[width=0.35\textwidth]{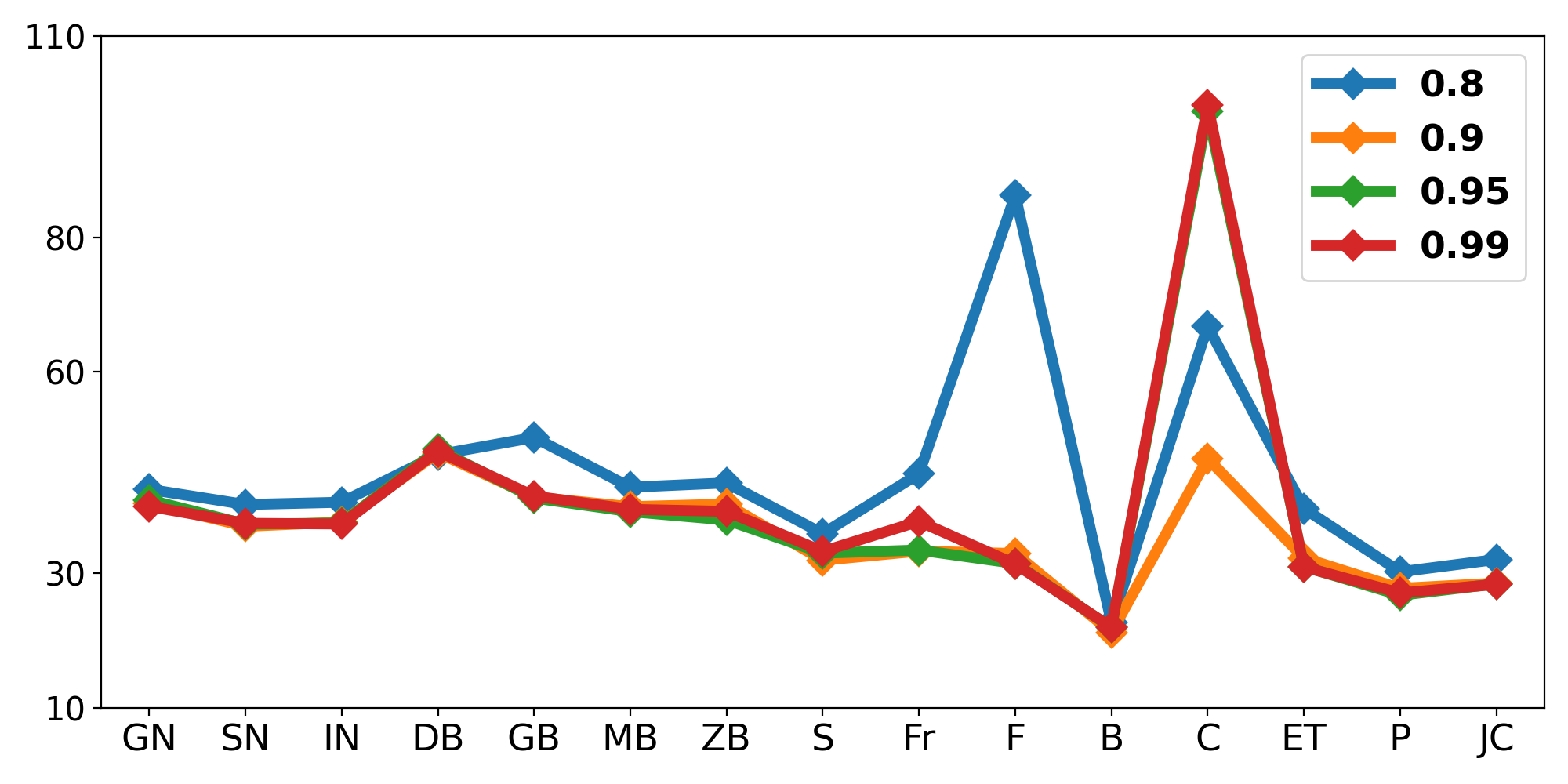}%
        }
    \end{tabular}

    \caption{Extended hyperparameter ablations on SEGA under CTTA (batch size 1), with CIFAR10-C on the left and ImageNet-C on the right. The deterioration-margin plots show that overly small $D$ under-adapts while overly large $D$ delays recovery until collapse; the quantile-level plots show that filtering the lower tail around $Q{=}0.2$ improves adaptation while more aggressive filtering under-adapts; the minimum-activation plots show that recovery should only reactivate after enough post-reset samples, around $T{=}50$; the warm-up plots show that the gate needs enough sensitivity history, around $U{=}2048$, before filtering samples for adaptation; and the EMA-weight plots show that intermediate $\zeta$ in the $0.9$--$0.95$ range yields the most stable errors across corruptions on both datasets.}
    \label{fig:ablation_extended}
\end{figure*}

Fig.~\ref{fig:ablation_extended} exte
nds the main-text ablations by pairing CIFAR10-C and ImageNet-C for the four control hyperparameters that govern recovery and filtering. Across both datasets, the deterioration margin $D$ shows the same trade-off: too small a margin under-adapts the model by triggering recovery too aggressively, whereas too large a margin delays recovery until after collapse. The quantile band $(Q_{\min},Q_{\max})$ likewise compares adaptation ranges $0\%$--$20\%$, $20\%$--$100\%$, $20\%$--$80\%$, and $80\%$--$100\%$ on both CIFAR10-C and ImageNet-C, and shows that using the $20\%$--$100\%$ band (dropping only the $0\%$--$20\%$ tail) performs best, while the $20\%$--$80\%$ and $80\%$--$100\%$ bands under-adapt by excluding too many useful samples. The minimum activation horizon $T$ confirms that the recovery rule should only reactivate after enough post-reset samples have been observed, with about $T{=}50$ steps providing a stable compromise, and the warm-up horizon $U$ shows that the gate needs a sufficiently long sensitivity history, around $U{=}2048$, before it can reliably determine which samples to filter for adaptation. Finally, the EMA-weight $\zeta$ plots indicate that intermediate values around $0.9$--$0.95$ give the most stable error curves across corruptions on both datasets, whereas more extreme weights either overreact to noise or slow the trend signal.

\subsection{Feature Visualization}
\label{sec:appendix-tsne}
\begin{figure*}[t!]
    \centering
    \small
    \subfigure[Source]{
        \includegraphics[width=0.2\textwidth]{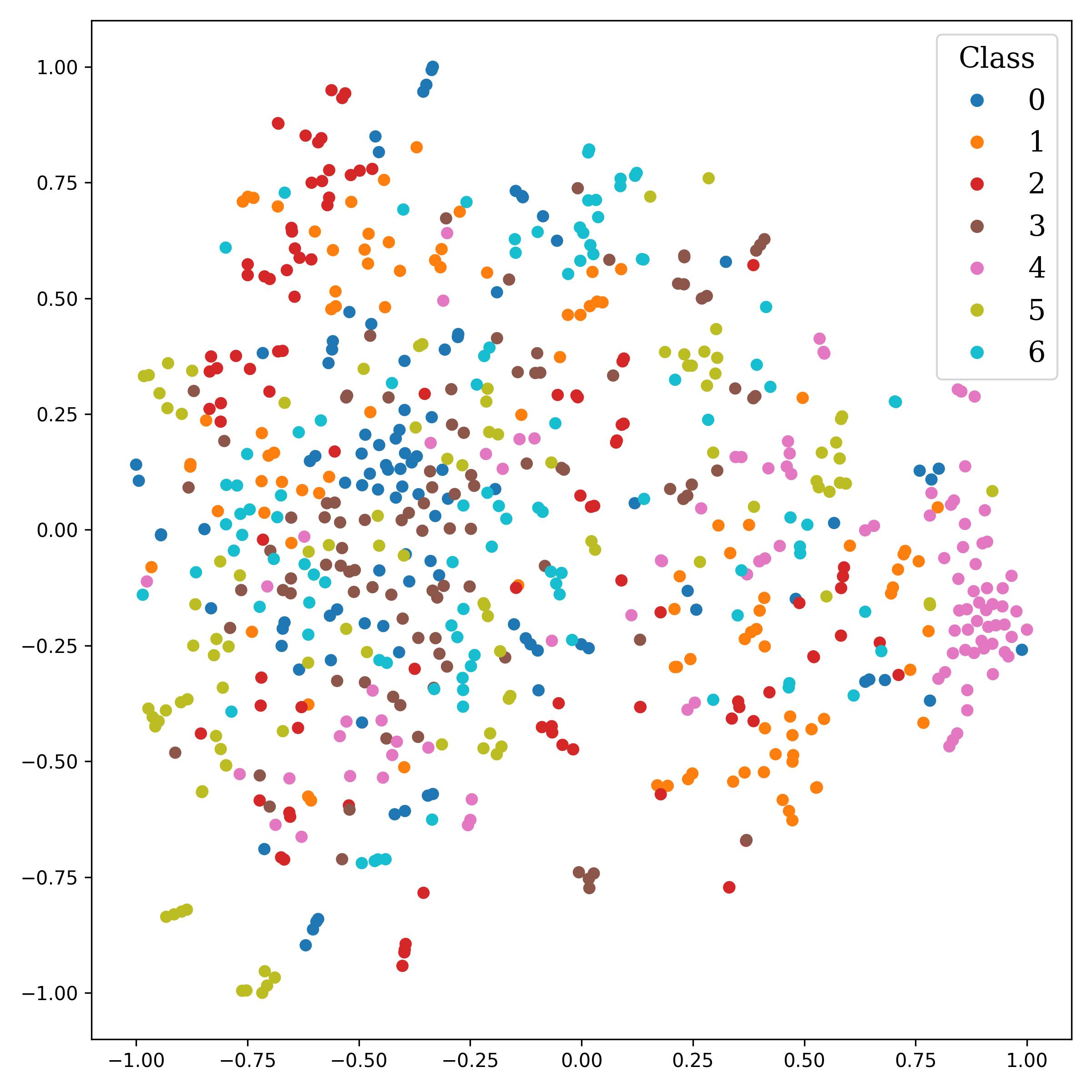}
    }
    \subfigure[$\mathcal{L}_{\mathrm{el}}$]{
        \includegraphics[width=0.2\textwidth]{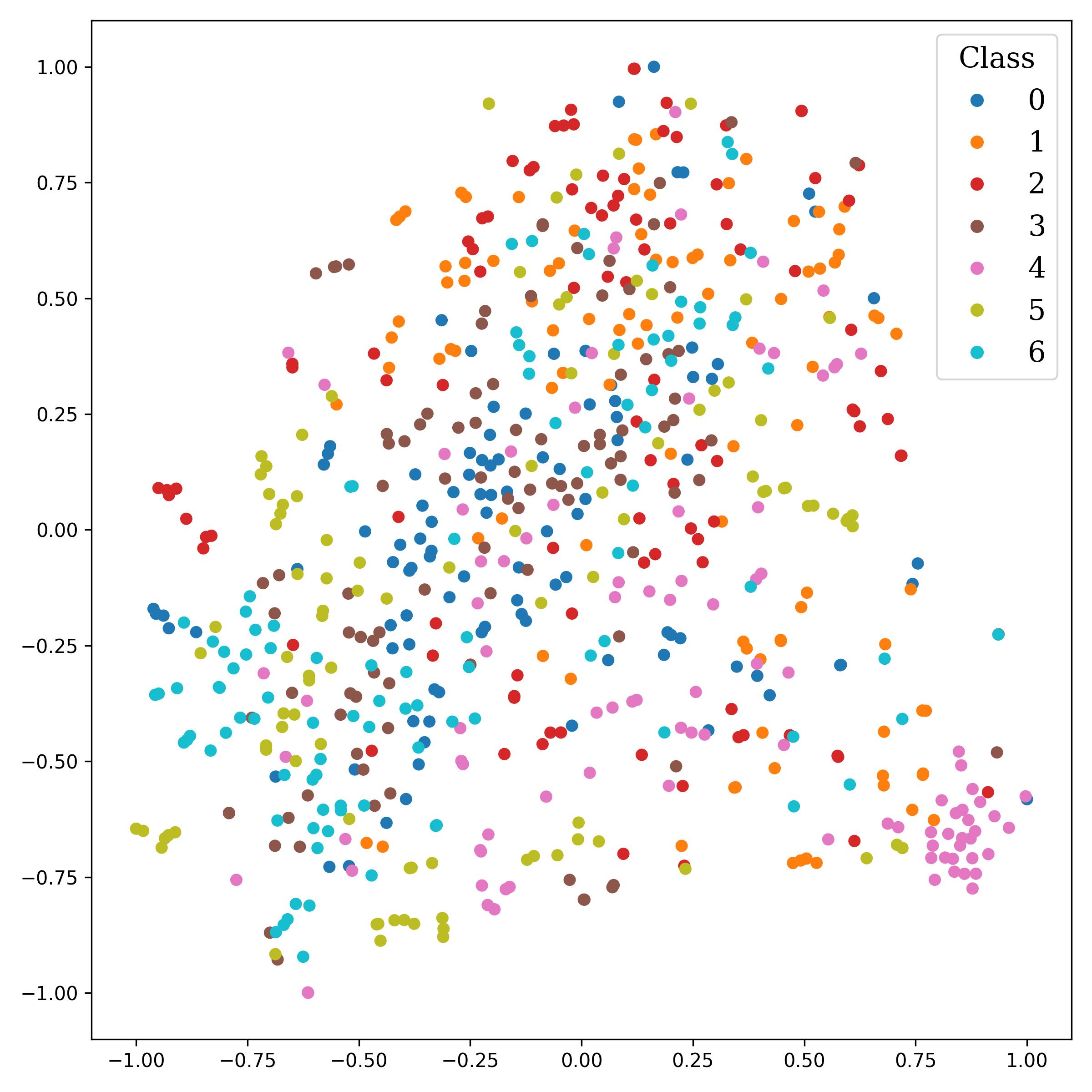}
    }
    \subfigure[$\mathcal{L}_{\mathrm{cl}}$]{
        \includegraphics[width=0.2\textwidth]{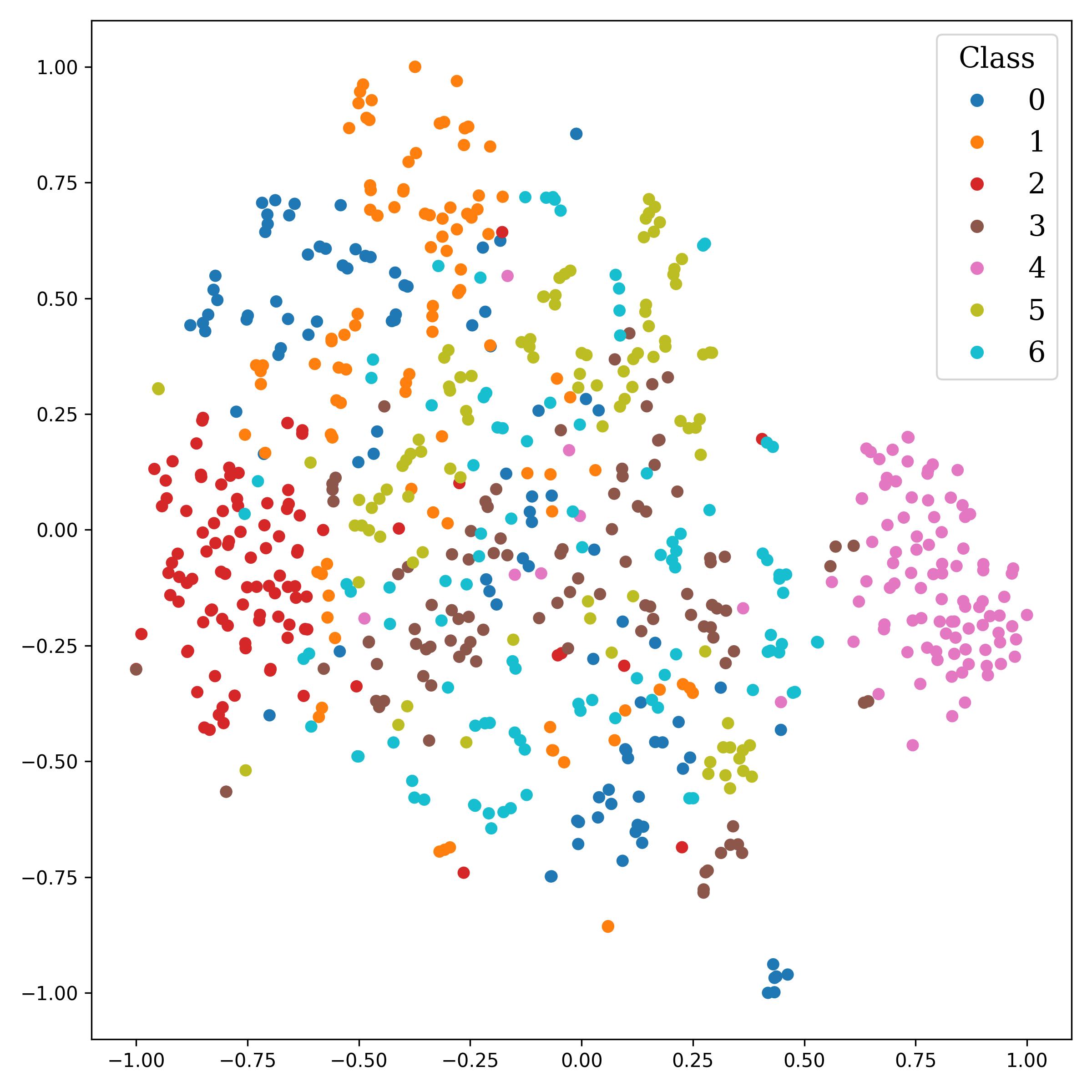}
    }
    \subfigure[$\mathcal{L}_{\mathrm{SEGA}}$]{
        \includegraphics[width=0.2\textwidth]{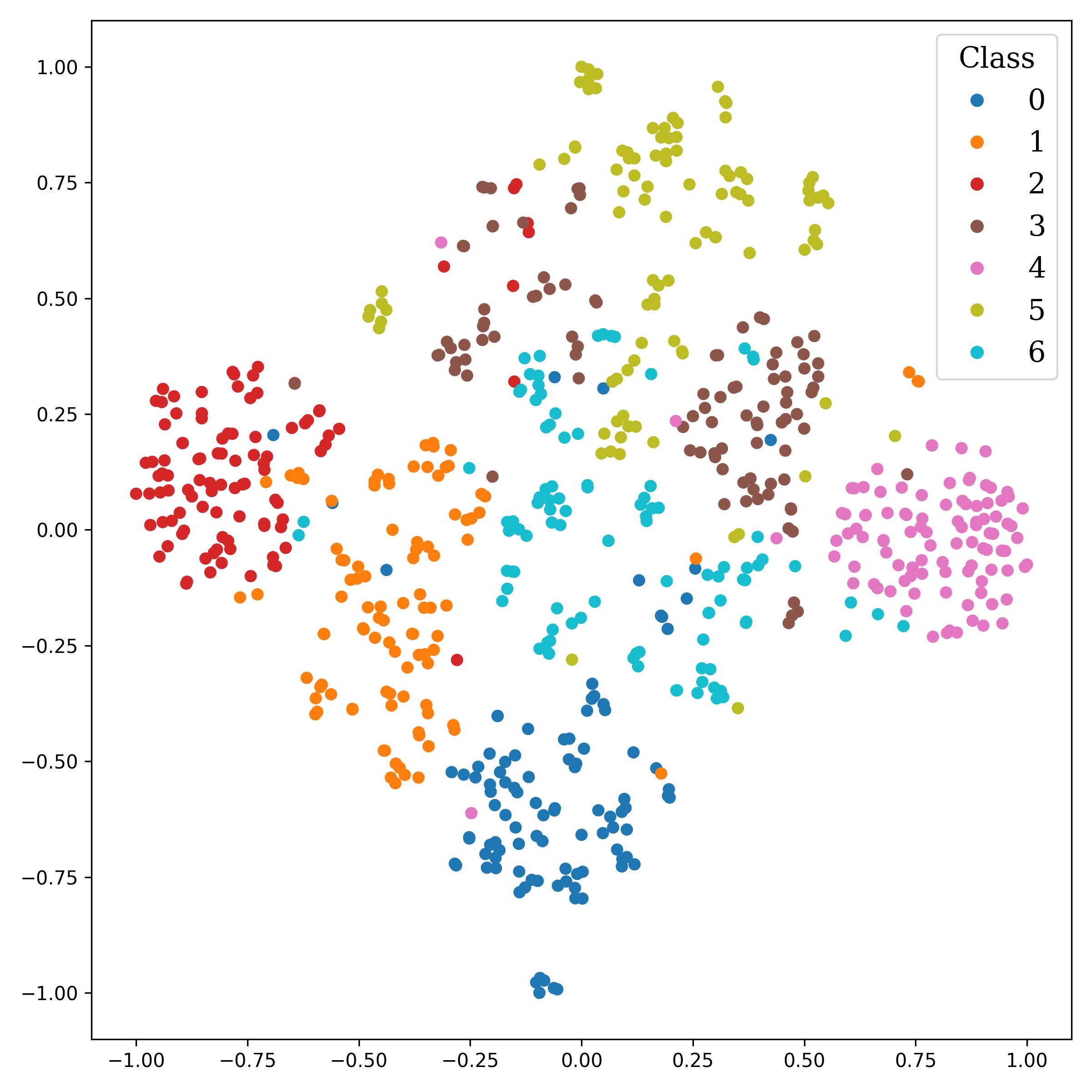}
    }

    \caption{t-SNE visualization of forward features from the adapted ViT-Base on FreshFish-C under the Fog corruption domain at severity level 5, chosen because Fog is one of the most challenging and collapse-prone domains.}
    \label{fig:features_tsne_imagenetc}
\end{figure*}

Fig.~\ref{fig:features_tsne_imagenetc} visualizes forward features under the Fog corruption domain, which we select as a challenging and collapse-prone case. The embeddings reinforce the main-text qualitative results: features from the source model and entropy loss are diffuse, the consistency-loss variant tightens some groups, and the full SEGA objective yields compact, well-separated class clusters. This supports the view that sensitivity-guided adaptation improves representation structure even in hard domains where entropy- or reset-based baselines are more vulnerable to drift.

\subsection{Aquaculture Corruption Samples}
\label{sec:appendix-aqua-samples}
\begin{figure*}[t]
    \centering
    \subfigure[Clean]{
        \includegraphics[width=0.23\linewidth]{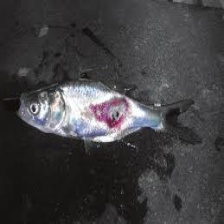}
    }
    \subfigure[Gaussian Noise]{
        \includegraphics[width=0.23\linewidth]{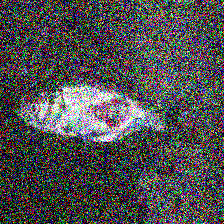}
    }
    \subfigure[Shot Noise]{
        \includegraphics[width=0.23\linewidth]{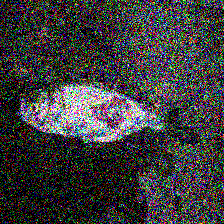}
    }
    \subfigure[Impulse Noise]{
        \includegraphics[width=0.23\linewidth]{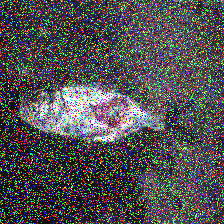}
    }\\[2pt]

    \subfigure[Defocus Blur]{
        \includegraphics[width=0.23\linewidth]{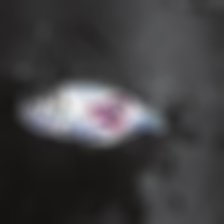}
    }
    \subfigure[Glass Blur]{
        \includegraphics[width=0.23\linewidth]{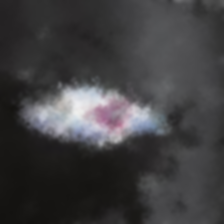}
    }
    \subfigure[Motion Blur]{
        \includegraphics[width=0.23\linewidth]{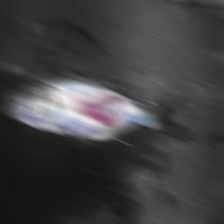}
    }
    \subfigure[Zoom Blur]{
        \includegraphics[width=0.23\linewidth]{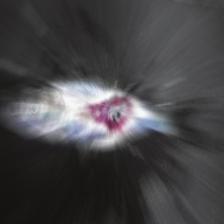}
    }\\[2pt]

    \subfigure[Snow]{
        \includegraphics[width=0.23\linewidth]{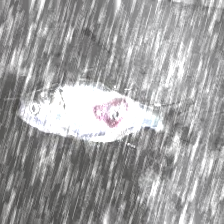}
    }
    \subfigure[Frost]{
        \includegraphics[width=0.23\linewidth]{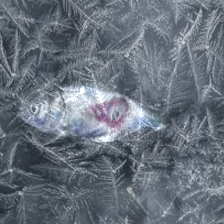}
    }
    \subfigure[Fog]{
        \includegraphics[width=0.23\linewidth]{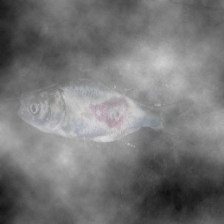}
    }
    \subfigure[Brightness]{
        \includegraphics[width=0.23\linewidth]{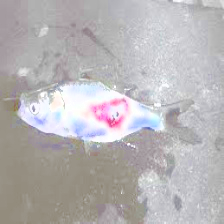}
    }\\[2pt]

    \subfigure[Contrast]{
        \includegraphics[width=0.23\linewidth]{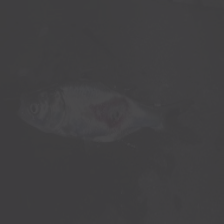}
    }
    \subfigure[Elastic Transform]{
        \includegraphics[width=0.23\linewidth]{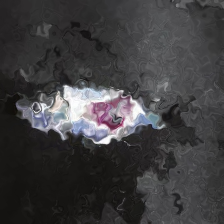}
    }
    \subfigure[Pixelate]{
        \includegraphics[width=0.23\linewidth]{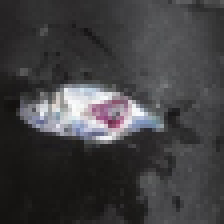}
    }
    \subfigure[JPEG Compression]{
        \includegraphics[width=0.23\linewidth]{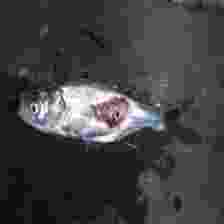}
    }

    \caption{Samples of FreshFish-C dataset at severity level 5 (highest) with different corruption domains applied, following CIFAR-C and ImageNet-C corruptions.}
    \label{fig:samples_freshfish}
\end{figure*}

We visualize representative FreshFish-C images at the highest corruption severity, pairing clean aquaculture frames with the full suite of CIFAR-C/ImageNet-C style corruptions including noise, blur, weather, brightness and contrast shifts, elastic distortions, pixelation, and JPEG artifacts. Together these grids illustrate the range of nuisance factors present in our aquaculture streams and motivate treating them as long-horizon CTTA benchmarks rather than simple static robustness tests.

{
    \small
    \bibliographystyle{ieeenat_fullname}
    \bibliography{main}
}